\input{a2z-font-check}
\PassOptionsToPackage{table}{xcolor}
\documentclass[10pt,a4paper,copyright]{krafton-ai}

\usepackage[authoryear,sort&compress,round]{natbib}
\usepackage{ragged2e}
\usepackage{wrapfig}
\usepackage{afterpage}
\usepackage{fancyvrb}
\usepackage{listings}

\usepackage{amsmath,amsfonts,bm}

\def\eqref#1{equation~\ref{#1}}

\def\1{\bm{1}}

\DeclareMathAlphabet{\mathsfit}{\encodingdefault}{\sfdefault}{m}{sl}
\SetMathAlphabet{\mathsfit}{bold}{\encodingdefault}{\sfdefault}{bx}{n}

\usepackage{xspace}
\usepackage{booktabs}
\usepackage{enumitem}
\usepackage{graphicx}
\usepackage{tabularx}
\usepackage{array}
\usepackage{amssymb}
\usepackage{pifont} 
\usepackage{pgfplots}
\usepackage{pgfplotstable}
\usepgfplotslibrary{statistics}
\usepackage{tikz}
\usepackage{xcolor}
\usepackage{multirow}
\usepackage{fix-cm}
\usepackage{makecell}
\usepackage{tikz}
\usepackage{bbm}
\usetikzlibrary{calc}

\let\AtoZCompanySansDefault\sfdefault
\IfFileExists{sourcesans.sty}{%
  \usepackage[scaled=0.92]{sourcesans}
}{%
  \usepackage[scaled=0.92]{sourcesanspro}
}
\let\AtoZFigureSansDefault\sfdefault
\let\sfdefault\AtoZCompanySansDefault
\newcommand{\AtoZSourceSans}{\fontfamily{\AtoZFigureSansDefault}\selectfont}

\usepackage{subcaption}
\usepackage[normalem]{ulem}

\newcommand{\GDDmetric}{GDD Fidelity\xspace}

\newcommand{\AtoZbench}{A2Z GameSpec-Bench\xspace}

\pgfplotsset{compat=1.18}
\usetikzlibrary{decorations.pathmorphing}

\definecolor{A2ZRed}{HTML}{E65042}
\definecolor{A2ZLight}{HTML}{E8EDEF}
\definecolor{A2ZText}{HTML}{465157}

\definecolor{A2ZFigText}{HTML}{3F454B}
\definecolor{A2ZFigAxis}{HTML}{C9CED3}
\definecolor{A2ZFigRest}{HTML}{F0F2F4}

\colorlet{A2ZGameCraft}{A2ZRed!82!white}
\colorlet{A2ZBGG}{A2ZRed!75!black}

\newcommand{\nomark}{--}

\newcommand{\benchyes}{\ensuremath{\bigcirc}}
\newcommand{\benchno}{\ensuremath{\times}}

\renewcommand{\benchyes}{\textcolor{green!60!black}{\ding{51}}}
\renewcommand{\benchno}{\textcolor{A2ZRed}{\ding{55}}}

\newcommand{\AtoZFigFont}{\AtoZSourceSans}

\newcommand{\Pass}{\textsc{Pass}}

\newcolumntype{C}{>{\centering\arraybackslash}X}
\usepackage{amsmath}

\usepackage{xcolor}
\definecolor{darkergreen}{RGB}{21, 152, 56}
\definecolor{ForestGreen}{RGB}{21, 152, 56}
\definecolor{red2}{RGB}{252, 54, 65}
\definecolor{Gray}{gray}{0.6}
\definecolor{LavenderBlush}{rgb}{1.0, 0.94, 0.96}
\definecolor{lightgray}{gray}{0.93} 

\definecolor{color1}{HTML}{006EB8}
\definecolor{topThreeBlue}{HTML}{08306B} 
\definecolor{topOneBlue}{HTML}{90C5E0}   
\definecolor{Baseline}{HTML}{6EACDA} 
\definecolor{Ours}{HTML}{fa8787}
\definecolor{mygray}{gray}{0.65}
\definecolor{lightblue}{HTML}{cfedfc}
\definecolor{lightred}{HTML}{fae0d2}
\definecolor{lightpink}{HTML}{ffdee5}
\definecolor{lightgreen}{HTML}{0FBD83}
\definecolor{lightyellow}{HTML}{FAA410}
\definecolor{palegreen}{HTML}{9da894}
\definecolor{revisiongray}{RGB}{145,145,145}

\DeclareRobustCommand{\ModelWithIcon}[2]{%
  \makebox[1.15em][c]{%
    \raisebox{-0.15em}[0pt][0pt]{%
      \includegraphics[width=1.0em,height=1.0em,keepaspectratio]{assets/images/complementarity/#1_icon.png}%
    }%
  }%
  \hspace{0.35em}#2%
}

\ifPDFTeX
  \def\AtoZFigureSansDefault{a2zsource}
  \DeclareFontSeriesDefault[rm]{bf}{bx}
\LoadMicrotypeFile{cmr}
\SetProtrusion[name=a2z-lmr-0]{encoding=T1,family=a2zroman}{
  A = {50,50},
  \AE = {50,  },
  F = {  ,50},
  J = {50,  },
  K = {  ,50},
  L = {  ,50},
  T = {50,50},
  V = {50,50},
  W = {50,50},
  X = {50,50},
  Y = {50,50},
  k = {  ,50},
  r = {  ,50},
  t = {  ,70},
  v = {50,50},
  w = {50,50},
  x = {50,50},
  y = {50,70},
  0 = {  ,50},
  1 = {100,200},
  2 = {50,50},
  3 = {50,50},
  4 = {70,70},
  5 = {  ,50},
  6 = {  ,50},
  7 = {50,100},
  8 = {  ,50},
  9 = {  ,50},
  . = { ,700},
  {,} = { ,500},
  : = { ,500},
  ; = { ,500},
  ! = { ,100},
  ? = { ,200},
  @ = {50,50},
  ~ = {200,250},
  \% = {50,50},
  * = {300,300},
  + = {250,250},
  - = {400,500},
  \textendash = {400,300},
  \textemdash = {300,200},
  _ = {200,200},
  \textbackslash = {200,300},
  ' = {300,400},
  \textquoteleft = {300,400},
  \textquoteright = {300,400},
  \textquotedblleft = {300,300},
  \textquotedblright = {300,300},
  \quotesinglbase = {400,400},
  \quotedblbase = {400,400},
  \guilsinglleft = {400,400},
  \guilsinglright = {300,500},
  \guillemotleft = {300,200},
  \guillemotright = {100,400},
  \textexclamdown = {100,   },
  \textquestiondown = {100,   },
  ( = {300,   },
  ) = {   ,300},
  < = {200,100},
  > = {100,200},
  \textbraceleft = {400,200},
  \textbraceright = {200,400}
}
\SetProtrusion[name=a2z-lmr-1]{encoding=T1,family=a2zroman,shape={it,sl}}{
  A = {125,100},
  \AE = {125,-55},
  B = {90,-40},
  C = {145,-75},
  D = {75, -28},
  E = {80,-55},
  F = {85,-80},
  G = {153,-15},
  H = {73,-60},
  I = {140,-120},
  \IJ = {140,-80},
  J = {135,-80},
  K = {70,-30},
  L = {87, 40},
  M = {67,-45},
  N = {75,-55},
  O = {150,-30},
  \OE = {150,-55},
  P = {82,-50},
  Q = {150,-30},
  R = {75, 15},
  S = {90,-65},
  $ = {100,-20},
  T = {220,-85},
  U = {230,-55},
  V = {260,-60},
  W = {185,-55},
  X = {70,-30},
  Y = {250,-60},
  Z = {90,-60},
  a = {150,-10},
  b = {170,   },
  c = {173,-10},
  d = {150,-55},
  e = {180, },
  f = {  ,-250},
  g = {150,-10},
  h = {100, },
  i = {210, },
  \ij = {210,-40},
  j = {  ,-40},
  k = {110,-50},
  l = {240,-110},
  m = {80, },
  n = {115, },
  o = {155, },
  q = {170,-40},
  r = {155,-40},
  s = {130, },
  t = {230,-10},
  u = {120, },
  v = {140,-25},
  w = {98,-20},
  x = {65,-40},
  y = {130,-20},
  z = {110,-80},
  0 = {170,-85},
  1 = {230,110},
  2 = {130,-70},
  3 = {140,-70},
  4 = {130,80},
  5 = {160, },
  6 = {175,-30},
  7 = {250,-150},
  8 = {130,-40},
  9 = {155,-80},
  . = { ,500},
  {,} = { ,450},
  : = { ,300},
  ; = { ,300},
  & = {130,30},
  \% = {180,50},
  * = {380,20},
  + = {180,200},
  @ = {180,10},
  ~ = {200,150},
  ( = {300, },
  ) = {  ,70},
  - = {500,300},
  \textendash = {500,300},
  \textemdash = {400,170},
  _ = {100,200},
  ' = {300,400},
  " = {500,300},
  \textquoteleft = {800,200},
  \textquoteright = {800,-20},
  \textquotedblleft = {540,100},
  \textquotedblright = {500,100},
  \quotesinglbase = {300,700},
  \quotedblbase = {200,600},
  \guilsinglleft = {500,300},
  \guilsinglright = {400,400},
  \guillemotleft = {400,100},
  \guillemotright = {200,300},
  \textexclamdown = {200,   },
  \textquestiondown = {200,   },
  < = {300,100},
  > = {200,100},
  \textbackslash = {300,300},
  \textbraceleft = {400,100},
  \textbraceright = {200,200}
}

\DeclareSymbolFont{legacymaths}{OT1}{cmr}{m}{n}
\SetSymbolFont{legacymaths}{bold}{OT1}{cmr}{bx}{n}
\DeclareMathAccent{\acute}{\mathalpha}{legacymaths}{19}
\DeclareMathAccent{\grave}{\mathalpha}{legacymaths}{18}
\DeclareMathAccent{\ddot}{\mathalpha}{legacymaths}{127}
\DeclareMathAccent{\tilde}{\mathalpha}{legacymaths}{126}
\DeclareMathAccent{\bar}{\mathalpha}{legacymaths}{22}
\DeclareMathAccent{\breve}{\mathalpha}{legacymaths}{21}
\DeclareMathAccent{\check}{\mathalpha}{legacymaths}{20}
\DeclareMathAccent{\hat}{\mathalpha}{legacymaths}{94}
\DeclareMathAccent{\dot}{\mathalpha}{legacymaths}{95}
\DeclareMathAccent{\mathring}{\mathalpha}{legacymaths}{23}
\DeclareMathSymbol{!}{\mathclose}{legacymaths}{33}
\DeclareMathSymbol{:}{\mathrel}{legacymaths}{58}
\DeclareMathSymbol{;}{\mathpunct}{legacymaths}{59}
\DeclareMathSymbol{?}{\mathclose}{legacymaths}{63}
\DeclareMathSymbol{0}{\mathalpha}{legacymaths}{48}
\DeclareMathSymbol{1}{\mathalpha}{legacymaths}{49}
\DeclareMathSymbol{2}{\mathalpha}{legacymaths}{50}
\DeclareMathSymbol{3}{\mathalpha}{legacymaths}{51}
\DeclareMathSymbol{4}{\mathalpha}{legacymaths}{52}
\DeclareMathSymbol{5}{\mathalpha}{legacymaths}{53}
\DeclareMathSymbol{6}{\mathalpha}{legacymaths}{54}
\DeclareMathSymbol{7}{\mathalpha}{legacymaths}{55}
\DeclareMathSymbol{8}{\mathalpha}{legacymaths}{56}
\DeclareMathSymbol{9}{\mathalpha}{legacymaths}{57}
\DeclareMathSymbol{\Gamma}{\mathalpha}{legacymaths}{0}
\DeclareMathSymbol{\Delta}{\mathalpha}{legacymaths}{1}
\DeclareMathSymbol{\Theta}{\mathalpha}{legacymaths}{2}
\DeclareMathSymbol{\Lambda}{\mathalpha}{legacymaths}{3}
\DeclareMathSymbol{\Xi}{\mathalpha}{legacymaths}{4}
\DeclareMathSymbol{\Pi}{\mathalpha}{legacymaths}{5}
\DeclareMathSymbol{\Sigma}{\mathalpha}{legacymaths}{6}
\DeclareMathSymbol{\Upsilon}{\mathalpha}{legacymaths}{7}
\DeclareMathSymbol{\Phi}{\mathalpha}{legacymaths}{8}
\DeclareMathSymbol{\Psi}{\mathalpha}{legacymaths}{9}
\DeclareMathSymbol{\Omega}{\mathalpha}{legacymaths}{10}
\DeclareMathSymbol{+}{\mathbin}{legacymaths}{43}
\DeclareMathSymbol{=}{\mathrel}{legacymaths}{61}
\DeclareMathDelimiter{(}{\mathopen}{legacymaths}{40}{largesymbols}{0}
\DeclareMathDelimiter{)}{\mathclose}{legacymaths}{41}{largesymbols}{1}
\DeclareMathDelimiter{[}{\mathopen}{legacymaths}{91}{largesymbols}{2}
\DeclareMathDelimiter{]}{\mathclose}{legacymaths}{93}{largesymbols}{3}
\DeclareMathDelimiter{/}{\mathord}{legacymaths}{47}{largesymbols}{14}
\DeclareMathSymbol{\mathdollar}{\mathord}{legacymaths}{36}
\DeclareSymbolFont{operators}{T1}{a2zroman}{m}{n}
\SetSymbolFont{operators}{bold}{T1}{a2zroman}{b}{n}
\DeclareSymbolFontAlphabet\mathrm{operators}
\SetMathAlphabet{\mathit}{normal}{T1}{a2zroman}{m}{it}
\SetMathAlphabet{\mathit}{bold}{T1}{a2zroman}{b}{it}
\SetMathAlphabet{\mathbf}{normal}{T1}{a2zroman}{b}{n}
\SetMathAlphabet{\mathsf}{normal}{T1}{a2zurbanist}{m}{n}
\SetMathAlphabet{\mathsf}{bold}{T1}{a2zurbanist}{b}{n}
\SetMathAlphabet{\mathtt}{normal}{T1}{a2zmono}{m}{n}
\SetMathAlphabet{\mathtt}{bold}{T1}{a2zmono}{b}{n}

\fi
\usepackage{cleveref}
\ifPDFTeX\else
\fi
\tcbuselibrary{listings,breakable}

\lstdefinestyle{a2zbot}{
  language={},
  morekeywords={export,default,async,function,await,const,let,if,for,return,true,false,null},
  sensitive=true,
  morecomment=[l]{//},
  morestring=[b]',
  basicstyle=\ttfamily\fontsize{6.5}{7.8}\selectfont,
  keywordstyle=\color[HTML]{2F6FBA}\bfseries,
  commentstyle=\color[HTML]{7E858C}\itshape,
  stringstyle=\color[HTML]{992B32},
  columns=fixed, basewidth=0.5em, keepspaces=true, showstringspaces=false,
  breaklines=true, breakindent=1.5em,
  frame=single, rulecolor=\color[HTML]{CDD2D7}, backgroundcolor=\color[HTML]{F7F7F8},
  xleftmargin=4pt, xrightmargin=4pt, framexleftmargin=4pt, aboveskip=2pt, belowskip=2pt
}
\renewcommand{\floatpagefraction}{.9}

\uselogo{}
\title{A2Z GameSpec-Bench: How Faithfully Can Coding Agents Generate Games from Game Design Specifications?}

\newcommand{\authormark}[1]{%
  \textsuperscript{{\normalfont #1}}%
}

\newcommand{\equalcontrib}{\authormark{*}}
\newcommand{\corecontrib}{\authormark{†}}
\newcommand{\internship}{\authormark{‡}}

\author[1]{Seonho Lee\equalcontrib\corecontrib}
\author[1]{Wonryeol Jeong\equalcontrib\corecontrib}
\author[1]{Alberto Cereser\corecontrib}
\author[1,2]{Inha Kang\internship}
\author[1]{Hyeonjong Kim}
\author[1,3]{Seungmin Kwak\internship}
\author[1]{Dongmin Park}
\affil[1]{KRAFTON}
\affil[2]{KAIST}
\affil[3]{Korea National University of Arts}

\begingroup

\footnotetext[1]{ Equal contribution.}
\footnotetext[2]{ Core contribution.}
\footnotetext[3]{ Work done during an internship at KRAFTON.}
\endgroup

\usepackage[useregional=false]{datetime2}
\DTMsetstyle{iso}
\paperdate{\DTMtoday}
\hypersetup{
  pdftitle={A2Z GameSpec-Bench: How Faithfully Can Coding Agents Generate Games from Game Design Specifications?},
  pdfauthor={Seonho Lee, Wonryeol Jeong, Alberto Cereser, Hyeonjong Kim, Seungmin Kwak},
  pdfsubject={Specification-following evaluation for agentic game development},
  bookmarksnumbered=true
}

\begin{abstract}

Delegating complete application development to coding agents requires preserving the intended design rather than simply producing plausible outputs through na\"ive prompting. Game development provides a demanding testbed, as long-form Game Design Documents (GDDs) describe requirements that must work together across game logic, visual rendering, and player interactions. However, existing game-development benchmarks typically use compact specifications and provide limited support for evaluating interdependent requirements across these aspects in long-form GDDs. We introduce~\textbf{\AtoZbench}, a benchmark of 100 long-form GDDs for evaluating end-to-end game development by agents. We measure \emph{\textbf{faithfulness}} by checking whether the game satisfies the GDD requirements and preserves the relationships among them. Each GDD is turned into a~\textsc{Dependency-Aware Contract} that contains rules, constraints, and prerequisite relations. Following game-development practices, we combine source-code inspection with agent-generated \textsc{Test Policies} for scenario-based replay and adaptive playtesting. The contract remains fixed across agents and revision rounds, while judgments and evidence linked to the same requirements support consistent comparison and failure detection. Our evaluations show that current agents struggle to jointly satisfy interdependent requirements across code implementation and actual play. Requirement-specific feedback improves~\emph{\GDDmetric} by 10.9\% relative to \emph{self-revision} after two rounds. \AtoZbench{} assesses end-to-end specification-following ability beyond implementation judgments and provides targeted feedback to support more faithful game development. Code and datasets are available at~\url{https://a2z-gamespec-bench.github.io}.

\end{abstract}

\begin{document}
\maketitle

\begin{figure}[!htbp]
  \centering
  \includegraphics[width=\textwidth]{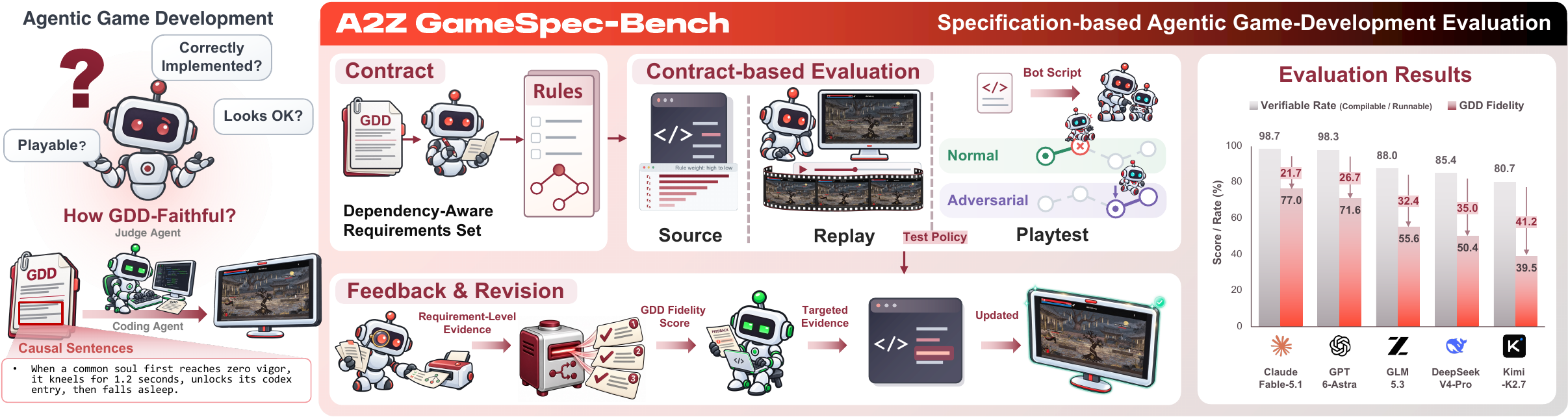}
  \caption{\textbf{Overview of A2Z GameSpec-Bench.} \textsc{Dependency-Aware Contracts} guide evaluation and targeted revision through source-code inspection, replay, and playtesting. Our evaluation results demonstrate that High compilability and runnability do not guarantee high GDD Fidelity.}
  \label{fig:main_figure}
\end{figure}

\section{Introduction}
\label{sec:intro}Coding agents are increasingly moving from isolated programming tasks to complete application development~\citep{qian2024chatdev,yang2026programbench}, including playable games from natural-language prompts~\citep{jiang2026opengame}. However, generating a plausible game from a na\"ive prompt differs significantly from implementing the game that a designer intends. In studio development, delegating implementation requires agents to interpret detailed specifications, connect individual features, and deliver complete games end-to-end without designers repeatedly directing each low-level step. Therefore, specification-driven development is important for retaining design control while scaling the work delegated to the agents. Games provide a demanding testbed for this ability in agents: specified designs must be satisfied across game logic, visual rendering, and player interactions.

\begin{figure*}[t]
\vspace{-5pt}
\centering
\begingroup
\providecolor{A2ZRed}{HTML}{D72D2A}
\providecolor{A2ZFigText}{HTML}{42494F}
\providecolor{A2ZFigAxis}{HTML}{CDD3DA}
\providecolor{A2ZFigRest}{HTML}{F3F4F5}
\providecommand{\AtoZFigFont}{\sffamily}
\colorlet{A2ZStudio}{A2ZRed!76!white}
\colorlet{A2ZGameBench}{A2ZRed!60!white}
\colorlet{A2ZBGG}{A2ZRed!75!black}
\colorlet{A2ZRuleBar}{A2ZRed!55!white}
\colorlet{A2ZPathBar}{A2ZRed!78!black}
\colorlet{A2ZAxisGray}{A2ZFigText}
\colorlet{A2ZGridGray}{A2ZFigAxis!66!white}

\edef\AtoZIntroTotalWidth{\the\linewidth}
\pgfmathsetlengthmacro{\AtoZIntroPanelWidth}{0.24*\AtoZIntroTotalWidth}
\pgfmathsetlengthmacro{\AtoZIntroWidePanelWidth}{2*\AtoZIntroPanelWidth}
\pgfmathsetlengthmacro{\AtoZIntroPanelGap}{(\AtoZIntroTotalWidth-4*\AtoZIntroPanelWidth)/2}
\edef\AtoZIntroGraphicWidth{\the\dimexpr\AtoZIntroPanelWidth-0.05pt\relax}
\edef\AtoZIntroWideGraphicWidth{\the\dimexpr\AtoZIntroGraphicWidth+\AtoZIntroGraphicWidth\relax}
\ifdefined\AtoZIntroRawBox\else\newsavebox{\AtoZIntroRawBox}\fi
\ifdefined\AtoZIntroQualBox\else\newsavebox{\AtoZIntroQualBox}\fi
\ifdefined\AtoZIntroCausalBox\else\newsavebox{\AtoZIntroCausalBox}\fi
\ifdefined\AtoZIntroPathBox\else\newsavebox{\AtoZIntroPathBox}\fi

\def\AtoZIntroFit#1#2#3{%
    \sbox{\AtoZIntroRawBox}{\ignorespaces\input{#3}\unskip}%
    \sbox{#1}{\resizebox{#2}{!}{\usebox{\AtoZIntroRawBox}}}%
}
    \sbox{\AtoZIntroRawBox}{\ignorespaces\input{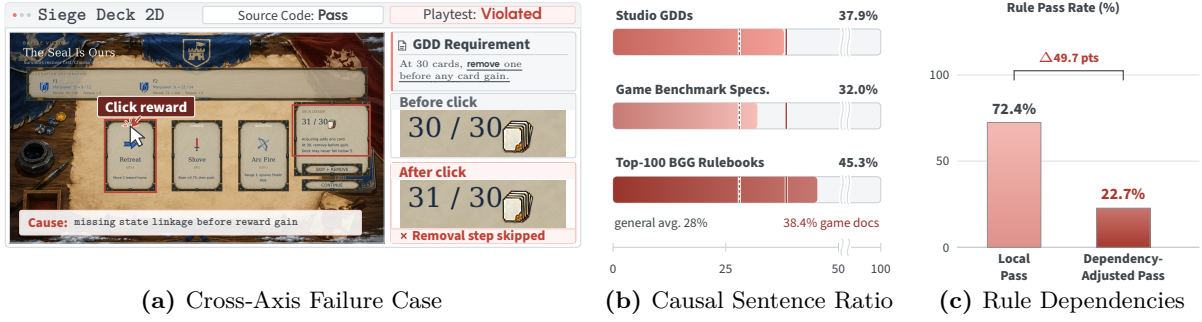}\unskip}%
    \sbox{\AtoZIntroQualBox}{\resizebox{\AtoZIntroWideGraphicWidth}{!}{\usebox{\AtoZIntroRawBox}}}%

    \sbox{\AtoZIntroRawBox}{\ignorespaces

\begingroup%
\providecommand{\AtoZFigFont}{\sffamily}%
\providecolor{A2ZFigText}{HTML}{42494F}%
\providecolor{A2ZFigAxis}{HTML}{CDD3DA}%
\providecolor{A2ZFigRest}{HTML}{F3F4F5}%
\begin{tikzpicture}[
    x=1cm,
    y=1.542035398cm,
    baseline=0.03084070796cm,
    font=\AtoZFigFont\fontsize{5.8}{6.6}\selectfont
]

\def\AtoZXStart{0.10}
\def\AtoZXBreakA{3.30}
\def\AtoZXBreakB{3.48}
\def\AtoZXFinish{3.90}

\newcommand{\AtoZValueX}[1]{%
    \AtoZXStart+(\AtoZXBreakA-\AtoZXStart)*#1/50%
}

\def\AtoZTopY{1.91}
\def\AtoZMidY{1.23}
\def\AtoZLowY{0.55}
\def\AtoZHalfBar{0.12}


\newcommand{\AtoZSoftWaveBreakA}[2]{%
    \pgfmathsetmacro{\AtoZWaveLow}{#1-#2}%
    \pgfmathsetmacro{\AtoZWaveHigh}{#1+#2}%
    \pgfmathsetmacro{\AtoZWaveControlLow}{#1-0.42*#2}%
    \pgfmathsetmacro{\AtoZWaveControlHigh}{#1+0.42*#2}%

    \foreach \AtoZWaveX in {3.35,3.44}{%
        \draw[
            draw=white,
            line width=1.9pt,
            line cap=round
        ]
            (\AtoZWaveX,\AtoZWaveLow)
            .. controls
                ({\AtoZWaveX-0.055},\AtoZWaveControlLow)
                and
                ({\AtoZWaveX+0.055},\AtoZWaveControlHigh)
            ..
            (\AtoZWaveX,\AtoZWaveHigh);

        \draw[
            draw=A2ZFigAxis!72!white,
            line width=0.46pt,
            line cap=round
        ]
            (\AtoZWaveX,\AtoZWaveLow)
            .. controls
                ({\AtoZWaveX-0.055},\AtoZWaveControlLow)
                and
                ({\AtoZWaveX+0.055},\AtoZWaveControlHigh)
            ..
            (\AtoZWaveX,\AtoZWaveHigh);
    }%
}

%
\newcommand{\AtoZCausalRow}[5]{%
    \pgfmathsetmacro{\AtoZYLow}{#1-\AtoZHalfBar}%
    \pgfmathsetmacro{\AtoZYHigh}{#1+\AtoZHalfBar}%
    \pgfmathsetmacro{\AtoZXCausal}{\AtoZValueX{#3}}%

    \pgfmathsetmacro{\AtoZLabelBase}{\AtoZYHigh+0.085}%

    \node[
        anchor=base west,
        inner sep=0pt,outer sep=0pt,
        text=A2ZFigText,
        font=\AtoZFigFont\fontsize{6.2}{7.2}\selectfont\bfseries
    ] at ({\AtoZXStart+0.04},\AtoZLabelBase) {#2};

    \node[
        anchor=base east,
        inner sep=0pt,outer sep=0pt,
        text=A2ZFigText,
        font=\AtoZFigFont\fontsize{6.2}{7.2}\selectfont\bfseries
    ] at ({\AtoZXFinish-0.04},\AtoZLabelBase) {#3\%};

    \path[
        fill=A2ZFigRest!78!white,
        draw=A2ZFigAxis,
        line width=0.32pt,
        rounded corners=1.45pt
    ]
        (\AtoZXStart,\AtoZYLow)
        rectangle
        (\AtoZXFinish,\AtoZYHigh);

    \begin{scope}
        \clip[
            rounded corners=1.45pt
        ]
            (\AtoZXStart,\AtoZYLow)
            rectangle
            (\AtoZXCausal,\AtoZYHigh);

        \shade[
            left color=#4,
            right color=#5,
            shading angle=90
        ]
            (\AtoZXStart,\AtoZYLow)
            rectangle
            (\AtoZXCausal,\AtoZYHigh);
    \end{scope}

    \AtoZSoftWaveBreakA{#1}{0.13}%
}

\AtoZCausalRow
    {\AtoZTopY}
    {Studio GDDs}
    {37.9}
    {A2ZStudio!84!black}
    {A2ZStudio!72!white}

\AtoZCausalRow
    {\AtoZMidY}
    {Game Benchmark Specs.}
    {32.0}
    {A2ZGameBench!82!black}
    {A2ZGameBench!62!white}

\AtoZCausalRow
    {\AtoZLowY}
    {Top-100 BGG Rulebooks}
    {45.3}
    {A2ZBGG!88!black}
    {A2ZBGG!70!white}

%
\newcommand{\AtoZReferenceRule}[2]{%
    \pgfmathsetmacro{\AtoZXRef}{\AtoZValueX{#1}}%

    \foreach \AtoZY in {
        \AtoZTopY,
        \AtoZMidY,
        \AtoZLowY
    }{%
        \draw[
            draw=white,
            opacity=0.90,
            line width=1.25pt
        ]
            (\AtoZXRef,\AtoZY-\AtoZHalfBar)
            --
            (\AtoZXRef,\AtoZY+\AtoZHalfBar);

        \draw[#2]
            (\AtoZXRef,\AtoZY-\AtoZHalfBar)
            --
            (\AtoZXRef,\AtoZY+\AtoZHalfBar);
    }%
}

\AtoZReferenceRule
    {28}
    {
        draw=A2ZFigText!78!white,
        line width=0.46pt,
        dash pattern=on 1.2pt off 0.8pt
    }

\AtoZReferenceRule
    {38.4}
    {
        draw=A2ZBGG!85!black,
        line width=0.52pt
    }

\pgfmathsetmacro{\AtoZXRefPrior}{\AtoZValueX{28}}
\pgfmathsetmacro{\AtoZXRefMean}{\AtoZValueX{38.4}}

\node[anchor=base west,inner sep=0pt,outer sep=0pt,text=A2ZFigText,font=\AtoZFigFont\fontsize{5.8}{6.6}\selectfont] at ({\AtoZXStart+0.02},0.205) {general avg.\ 28\%};
\node[anchor=base east,inner sep=0pt,outer sep=0pt,text=A2ZBGG!86!black,font=\AtoZFigFont\fontsize{5.8}{6.6}\selectfont] at ({\AtoZXFinish-0.02},0.205) {38.4\% game docs};

\draw[
    draw=A2ZFigAxis,
    line width=0.35pt
]
    (\AtoZXStart,0.02) -- (\AtoZXBreakA,0.02)
    (\AtoZXBreakB,0.02) -- (\AtoZXFinish,0.02);

\foreach \AtoZValue/\AtoZLabel in {
    0/0,
    25/25,
    50/50
}{%
    \pgfmathsetmacro{\AtoZXTick}{\AtoZValueX{\AtoZValue}}%

    \draw[
        draw=A2ZFigAxis,
        line width=0.35pt
    ]
        (\AtoZXTick,-0.02) -- (\AtoZXTick,0.06);

    \node[
        anchor=north,
        text=A2ZFigText,
        font=\AtoZFigFont\fontsize{5.8}{6.6}\selectfont
    ] at (\AtoZXTick,-0.055) {\AtoZLabel};
}

\draw[
    draw=A2ZFigAxis,
    line width=0.35pt
]
    (\AtoZXFinish,-0.02) -- (\AtoZXFinish,0.06);

\node[
    anchor=north,
    text=A2ZFigText,
    font=\AtoZFigFont\fontsize{5.8}{6.6}\selectfont
] at (\AtoZXFinish,-0.055) {100};

\AtoZSoftWaveBreakA{0.02}{0.075}

%
\pgfresetboundingbox
\path[use as bounding box]
    (0,-0.40) rectangle (4.10,2.28);

\end{tikzpicture}%
\endgroup%
\unskip}%
    \sbox{\AtoZIntroCausalBox}{\resizebox{\AtoZIntroGraphicWidth}{!}{\usebox{\AtoZIntroRawBox}}}%

    \sbox{\AtoZIntroRawBox}{\ignorespaces
\begingroup%
\providecommand{\AtoZFigFont}{\sffamily}%
\providecolor{A2ZFigText}{HTML}{42494F}%
\providecolor{A2ZFigAxis}{HTML}{CDD3DA}%
\def\AtoZRuleMean{72.403511903}%
\def\AtoZAdjustedMean{22.7}%
\begin{tikzpicture}[
    x=1cm,y=1.528508772cm,baseline=0pt,
    font=\AtoZFigFont\fontsize{6}{7}\selectfont,
    every node/.style={inner sep=0pt,outer sep=0pt,text=A2ZFigText}
]
    \foreach \value in {0,50,100} {
        \pgfmathsetmacro{\AtoZY}{0.016*\value}
        \draw[draw=A2ZFigAxis!66!white,line width=0.35pt]
            (0.54,\AtoZY) -- (3.98,\AtoZY);
        \node[anchor=east,font=\AtoZFigFont\fontsize{5.8}{6.6}\selectfont]
            at (0.43,\AtoZY) {\value};
    }
    \draw[draw=A2ZFigAxis,line width=0.50pt] (0.54,0) -- (3.98,0);
    \node[anchor=north,font=\AtoZFigFont\fontsize{6.1}{7.0}\selectfont\bfseries]
        at (2.05,2.28) {Rule Pass Rate (\%)};

    \begin{scope}
        \clip (0.97,0) rectangle (1.73,{0.016*\AtoZRuleMean});
        \shade[bottom color=A2ZRuleBar!92!black,top color=A2ZRuleBar!72!white]
            (0.97,0) rectangle (1.73,{0.016*\AtoZRuleMean});
    \end{scope}
    \draw[draw=A2ZFigAxis!70!black,line width=0.22pt]
        (0.97,0) rectangle (1.73,{0.016*\AtoZRuleMean});

    \begin{scope}
        \clip (2.52,0) rectangle (3.28,{0.016*\AtoZAdjustedMean});
        \shade[bottom color=A2ZPathBar!92!black,top color=A2ZPathBar!72!white]
            (2.52,0) rectangle (3.28,{0.016*\AtoZAdjustedMean});
    \end{scope}
    \draw[draw=A2ZFigAxis!70!black,line width=0.22pt]
        (2.52,0) rectangle (3.28,{0.016*\AtoZAdjustedMean});

    \node[anchor=south,font=\AtoZFigFont\fontsize{7}{8}\selectfont\bfseries]
        at (1.35,{0.016*\AtoZRuleMean+0.08}) {72.4\%};

    \node[anchor=south,text=A2ZPathBar!92!black,
          font=\AtoZFigFont\fontsize{7}{8}\selectfont\bfseries]
        at (2.90,{0.016*\AtoZAdjustedMean+0.08}) {22.7\%};

    \draw[draw=A2ZFigText!85!white,line width=0.55pt]
        (1.35,1.55) -- (1.35,1.64) -- (2.90,1.64) -- (2.90,1.55);
    \node[
        anchor=south,
        text=A2ZRed!78!black,
        fill=white,
        fill opacity=0.92,
        text opacity=1,
        rounded corners=1pt,
        inner xsep=2pt,
        inner ysep=0.7pt,
        font=\AtoZFigFont\fontsize{5.7}{6.4}\selectfont\bfseries
    ] at (2.125,1.68) {$\Delta$49.7 pts};

    \node[anchor=north,align=center,
          font=\AtoZFigFont\fontsize{6.0}{7.0}\selectfont\bfseries]
        at (1.35,-0.055) {Local\\Pass};
    \node[anchor=north,align=center,
          font=\AtoZFigFont\fontsize{6.0}{7.0}\selectfont\bfseries]
        at (2.90,-0.055) {Dependency-\\Adjusted Pass};
    \pgfresetboundingbox
    \path[use as bounding box] (0,-0.40) rectangle (4.10,2.28);
\end{tikzpicture}%
\endgroup%
\unskip}%
    \sbox{\AtoZIntroPathBox}{\resizebox{\AtoZIntroGraphicWidth}{!}{\usebox{\AtoZIntroRawBox}}}%

\def\AtoZIntroAbove{0pt}%
\def\AtoZIntroBelow{0pt}%
\def\AtoZIntroMeasure#1{%
    \ifdim\ht#1>\AtoZIntroAbove\edef\AtoZIntroAbove{\the\ht#1}\fi
    \ifdim\dp#1>\AtoZIntroBelow\edef\AtoZIntroBelow{\the\dp#1}\fi
}
\AtoZIntroMeasure{\AtoZIntroQualBox}
\AtoZIntroMeasure{\AtoZIntroCausalBox}
\AtoZIntroMeasure{\AtoZIntroPathBox}
\def\AtoZIntroPanel#1{%
    \noindent\makebox[\linewidth][c]{%
        \raisebox{0pt}[\AtoZIntroAbove][\AtoZIntroBelow]{\usebox{#1}}%
    }\par
}
\captionsetup{subrefformat=parens}
\captionsetup[subfigure]{font=small,labelfont=bf,justification=centering,singlelinecheck=false,position=bottom,skip=3pt}
\noindent\makebox[\AtoZIntroTotalWidth][l]{%
\begin{subfigure}[t]{\AtoZIntroWidePanelWidth}
    \centering
    \AtoZIntroPanel{\AtoZIntroQualBox}
    \vspace{-5pt}
    \caption{Cross-Axis Failure Case}
    \label{fig:intro_analysis_contract}
    \label{fig:intro_analysis_qualitative}
\end{subfigure}\hspace{\AtoZIntroPanelGap}%
\begin{subfigure}[t]{\AtoZIntroPanelWidth}
    \centering
    \AtoZIntroPanel{\AtoZIntroCausalBox}
    \vspace{-5pt}
    \caption{Causal Sentence Ratio}
    \label{fig:intro_analysis_causal}
\end{subfigure}\hspace{\AtoZIntroPanelGap}%
\begin{subfigure}[t]{\AtoZIntroPanelWidth}
    \centering
    \AtoZIntroPanel{\AtoZIntroPathBox}
    \vspace{-5pt}
    \caption{Rule Dependencies}
    \label{fig:intro_analysis_dependency}
\end{subfigure}%
}\par
\vspace{-2pt}
\caption{\textbf{Challenges in specification-driven game development.} \protect\subref{fig:intro_analysis_qualitative} In our game \texttt{Siege Deck 2D}, the deck-limit requirement gets full source-code credit, but playtesting reveals a reward gain without the card removal. \protect\subref{fig:intro_analysis_causal} Across three game-design corpora with 353 documents, CiRA~\protect\citep{fischbach2021cira} classifies 38.4\% of sentences as causal on average, compared with 28\% for general documents~\protect\citep{frattini2023causality} (Appendix~\ref{app:causal_game_spec}). \protect\subref{fig:intro_analysis_dependency} Source-code pass rates decrease when all upstream rules must also pass (Appendix~\ref{app:dependency_analysis}).}

\label{fig:intro_analysis}
\endgroup
\vspace{-5pt}
\end{figure*}

In game development, long-form specifications called Game Design Documents (GDDs), including studio documents~\citep{hamilton1995racenchase}, often serve as a shared reference for implementing and verifying the intended design~\citep{callele2005requirements,salazar2012gddproposal}. Although their formats vary, their requirements provide the basis for assessing implementation and commonly involve causal dependencies (Figure~\ref{fig:intro_analysis_causal}). In practice, developers verify the results using code-level checks, visual testing of specific scenarios, and quality assurance (QA) playtesting~\citep{kisel2016running,epicgamesautomation}. These checks must follow the same requirements and account for relations in which one rule's effects satisfy another's conditions. They should also identify requirement-specific mismatches that guide revisions toward the intended design. Thus, faithfulness requires preserving the specified rules and their relationships across implementation, rendered behavior, and actual play. This raises the main question: \emph{How \textbf{faithfully} can coding agents implement design specifications as a complete game?}

Existing game-development benchmarks assess specified behavior through replay rubrics~\citep{luo2026gamecraft}, state-initialized tests~\citep{jia2026gamegen}, or source-code inspection~\citep{chi2026gamedevbench}. However, their input specifications are substantially compact, and they provide limited support for assessing long-form GDDs through a common representation of requirement dependencies across implementation, rendering, and continuous play (Table~\ref{tab:intro_bench_comparison}). As shown in Figure~\ref{fig:intro_analysis_qualitative}, a single evaluation axis used in existing benchmarks such as source-code inspection can miss design violations that our playtests detect during runtime. This implies that evaluating fidelity to long-form GDDs needs tracking dependencies and checking the same requirements through code, visual, and execution evidence.

We introduce~\textbf{\AtoZbench}, a benchmark of 100 long-form GDDs for evaluating end-to-end specification-driven game development. Each task asks a coding agent to deliver a source project and playable build from a GDD to measure its faithfulness defined as~\emph{\GDDmetric}. The corpus contains 50 \emph{Small} designs with compact scope and 50 \emph{Big} designs with extensive content. We construct these GDDs from game briefs through an agentic authoring pipeline that checks for omissions and inconsistencies (Appendix~\ref{app:gdd_dataset}). For evaluation, each GDD is turned into a~\textsc{Dependency-Aware Contract} that records individual rules, constraints, and prerequisite relations between rules. Constructed from the GDD and kept fixed across agents and revision rounds, the contract makes these relationships explicit rather than treating requirements as an independent checklist.

Then, we organize evaluation from code-level checks and visual assessment to QA playtesting. Alongside source-code inspection, agents construct requirement-specific \textsc{Test Policies} that determine how to exercise the game and collect evidence against the contract. Scenario-based replays reproduce specified situations, while trace-guided frame selection retrieves visual evidence of the expected responses. For adaptive playtesting, \textsc{Code-as-Policy}~\citep{liang2023codeaspolicies} bots select player inputs based on the current state. Normal play checks how behaviors connect from the initial state, while targeted adversarial tests establish preconditions for unverified requirements and check their subsequent effects. Consequently, agents construct and execute tests to verify each game based on our contract as summarized in Figure~\ref{fig:main_figure}. The resulting judgments and evidence are linked to the same requirements for~\GDDmetric scoring and further revision feedback.


\begin{table}[t]
    \centering
    \caption{\textbf{Comparison of agentic game development benchmarks.} \AtoZbench{} fixes a dependency-aware evaluation contract from long-form GDDs independently of generated outputs and links judgments from complementary evidence channels to the same requirements.}
    \label{tab:intro_bench_comparison}
    \vspace{-8pt}

    \begingroup
    \scriptsize
    \setlength{\tabcolsep}{2.2pt}
    \renewcommand{\arraystretch}{1.22}

    \resizebox{\linewidth}{!}{%
    \begin{tabular}{@{}lclr@{\hspace{7pt}}lcccccc@{}}
        \toprule
        & \multicolumn{3}{c}{\textbf{Benchmark Setup}}
        & \multicolumn{4}{c}{\textbf{Requirement Evaluation}}
        & \multicolumn{3}{c}{\textbf{Evaluation Channels}} \\
        \cmidrule(lr){2-4}
        \cmidrule(lr){5-8}
        \cmidrule(lr){9-11}

        \multirow[c]{1}{*}[+6pt]{\textbf{Benchmark}}
        & \makecell[c]{\textbf{Full-Game}\\\textbf{Generation}}
        & \makecell[c]{\textbf{Input}\\\textbf{Type}}
        & \multicolumn{1}{c}{\makecell[c]{\textbf{\# of Spec.}\\\textbf{Tokens}}}
        & \multicolumn{1}{c}{\makecell[c]{\textbf{Evaluation}\\\textbf{Target}}}
        & \textbf{Relations}
        & \textbf{Predefined}
        & \textbf{Cross-Axis}
        & \makecell[c]{\textbf{Source}\\\textbf{Code}}
        & \makecell[c]{\textbf{Rendered}\\\textbf{Behavior}}
        & \makecell[c]{\textbf{Adaptive}\\\textbf{Play}} \\
        \midrule

        GameDevBench
        & \benchno & Task + project & 185
        & Task-specific tests
        & \benchno & \benchyes & \benchno
        & \benchyes & \benchno & \benchno \\

        GameEngineBench
        & \benchno & Spec. + project & \nomark
        & Tests + LLM judge
        & \benchno & \benchyes & \benchno
        & \benchyes & \benchno & \benchno \\

        OpenGame-Bench
        & \benchyes & Game spec. & 1,830
        & Build / visual / intent
        & \benchno & \benchyes & \benchno
        & \benchno & \benchyes & \benchno \\

        WebGameBench
        & \benchyes & Structured spec. & \nomark
        & Runtime quality
        & \benchno & \benchyes & \benchno
        & \benchno & \benchyes & \benchyes \\

        GameCraft-Bench
        & \benchyes & Game spec. & 1,547
        & Predefined rubric
        & \benchno & \benchyes & \benchno
        & \benchno & \benchyes & \benchno \\

        PlaytestArena
        & \benchyes & Short prompt & 131
        & Behavior rubric
        & \benchno & \benchyes & \benchno
        & \benchno & \benchyes & \benchyes \\

        GameGen-Verifier
        & \benchno & Spec. + project & 7,998
        & Sparse keypoints
        & \benchyes & \benchyes & \benchyes
        & \benchyes & \benchyes & \benchno \\

        GameXpert-Bench$^{\dagger}$
        & \benchyes & Design brief & 84
        & Post-hoc event rubric
        & \benchno & \benchno & \benchyes
        & \benchyes & \benchno & \benchyes \\

        \midrule
        \textbf{\AtoZbench{}}
        & \benchyes & \textbf{Long-form GDD} & \textbf{20,191}
        & \textbf{GDD contract set}
        & \benchyes & \benchyes & \benchyes
        & \benchyes & \benchyes & \benchyes \\
        \bottomrule
    \end{tabular}%
    }

    \vspace{2pt}
    \begin{minipage}{\linewidth}
        \scriptsize
        $^\dagger$ We report the GameGen track; human ratings are omitted from this table.
        Spec. tokens are rounded mean lengths of accessible specifications using
        \texttt{o200k\_base} without project code; PlaytestArena is author-reported. Dash (\nomark{}) means that specs. cannot be publicly obtained. For \textit{Adaptive Play},~\benchyes denotes \underline{closed-loop} evaluator interaction in which the next action is selected online based on the current observation or runtime state.
    \end{minipage}
    \endgroup
    \vspace{-5pt}
\end{table}

Our evaluation shows that high compilability and runnability do not imply high~\GDDmetric{} (Figure~\ref{fig:main_figure}). For example, Claude-Fable-5.1 achieves a verifiable rate of 98.7\% under compile and runtime checks, while its overall~\GDDmetric{} is 77.0. In addition, as shown in Figure~\ref{fig:intro_analysis_dependency}, our code-level evaluation shows that locally passing rules can still depend on failed prerequisites. Across 100 GPT-5.6-Sol games, the mean pass rate for dependency-linked rules is 72.4\% from local source-code judgments but 22.7\% when all upstream rules must also pass, showing a graph-based reachability proxy (Appendix~\ref{app:dependency_analysis}). By checking the same requirements through replay and playtests, \AtoZbench{} further exposes design mismatches in visual output and execution. With enumerated source judgments held fixed, adding dependency context increases coverage of recorded playtest violations from 71.1\% to 80.2\% across the 100-game analysis (Section~\ref{sec:integration_diagnosis}). On 50 \emph{Big} GDDs, requirement-specific feedback improves overall~\GDDmetric{} by 10.9\% relative to \emph{self-revision} after two rounds from the same initial builds. \AtoZbench{} supports systematic comparison of coding agents' specification-following ability and targeted revision toward the intended design.

\section{Related Work}
\label{sec:related_work}
\paragraph{Specification-Based Evaluation.} Evaluating generative outputs against detailed specifications often begins by decomposing the specifications into concrete criteria. In text-to-image evaluation, TIFA~\citep{hu2023tifa} and VPEval~\citep{cho2023visual} decompose text prompts into fine-grained visual checks and evaluate them using visual question answering and specialized modules, respectively. In text-to-CAD generation, MUSE~\citep{dong2026muse} pairs design instances with structured specifications and evaluates generated CAD models through code execution, geometric validation, and VLM-based design-intent alignment. For interactive web artifacts, LiveEvalBench~\citep{wang2026liveevalbench} combines build, source-code, and browser-interaction evidence, while WebVR~\citep{dai2026webvr} evaluates generated HTML and execution videos against visual and interaction rubrics. For games, these criteria may not be independent: an unmet prerequisite can prevent downstream behavior from being exercised or observed at runtime.

\paragraph{Structured Evaluation Contracts.}
Davidsonian Scene Graph~\citep{cho2024davidsonian} links prompt-derived questions through prerequisite entities, while ComplexBench~\citep{wen2407benchmarking} aggregates instruction-following judgments according to constraint composition. DafnyCOMP~\citep{xu2026local} studies whether specifications generated for individual functions remain valid when functions interact. This provides a formal-verification analogue, not direct evidence of failures in generated games. For web artifacts, WebRISE~\citep{meng2026webrise} represents requirements as Interaction Contract Graphs over observable states, transitions, and predicates. WebGrader~\citep{chen2026webgrader} first plans required flows from the request, then uses source code and the live DOM to ground executable Flow Contracts and evidence-based verdicts. It develops these graders for reinforcement learning. Our contracts apply related principles to GDDs and preserve the distinction between local tests under supplied preconditions and evidence of the specified gameplay paths.

\paragraph{Agentic Game Development and Evaluation.}
Benchmarks for agentic game development range from scoped game-engine tasks to complete game generation. GameDevBench~\citep{chi2026gamedevbench} and GameEngineBench~\citep{la2026gameenginebench} evaluate agents on scoped tasks in existing game projects. At the full-game level, OpenGame-Bench~\citep{jiang2026opengame} scores browser-native games on build health, visual usability, and intent alignment, while WebGameBench~\citep{zhang2026webgamebench} evaluates games from specifications through browser interaction. GameCraft-Bench~\citep{luo2026gamecraft} replays traces and assesses fixed-rate sampled frames against a hidden multi-modal rubric. PlaytestArena uses a GUI agent to play browser games against behavior rubrics, and Play2Code uses its feedback for iterative game revision~\citep{huang2026playtest}. Orak~\citep{park2026orak}, GameWorld~\citep{ouyang2026gameworld}, and OmniGameArena~\citep{lin2026omnigamearena} instead evaluate game-playing agents in predefined games, addressing a different setting from testing newly generated games. GameGen-Verifier~\citep{jia2026gamegen} extracts verifiable keypoints from a game specification, initializes their required runtime states, and tests them through bounded interaction. GameXpert-Bench~\citep{chen2026gamexpertbench} combines code inspection and live validation in its GameGen track using an event rubric derived from cross-model outputs. \AtoZbench{} instead fixes a~\textsc{Dependency-Aware Contract} from each long-form GDD before inspecting generated builds. The contract guides source-code inspection and agent-generated~\textsc{Test Policies} for scenario-based replay and adaptive playtesting. Judgments and evidence remain linked to the same requirements across agents and revision rounds for consistent assessment of specification faithfulness and targeted revision.

\section{A2Z GameSpec-Bench}
\label{sec:method}
\begin{figure*}[t]
    \centering
    \includegraphics[width=\textwidth]
    {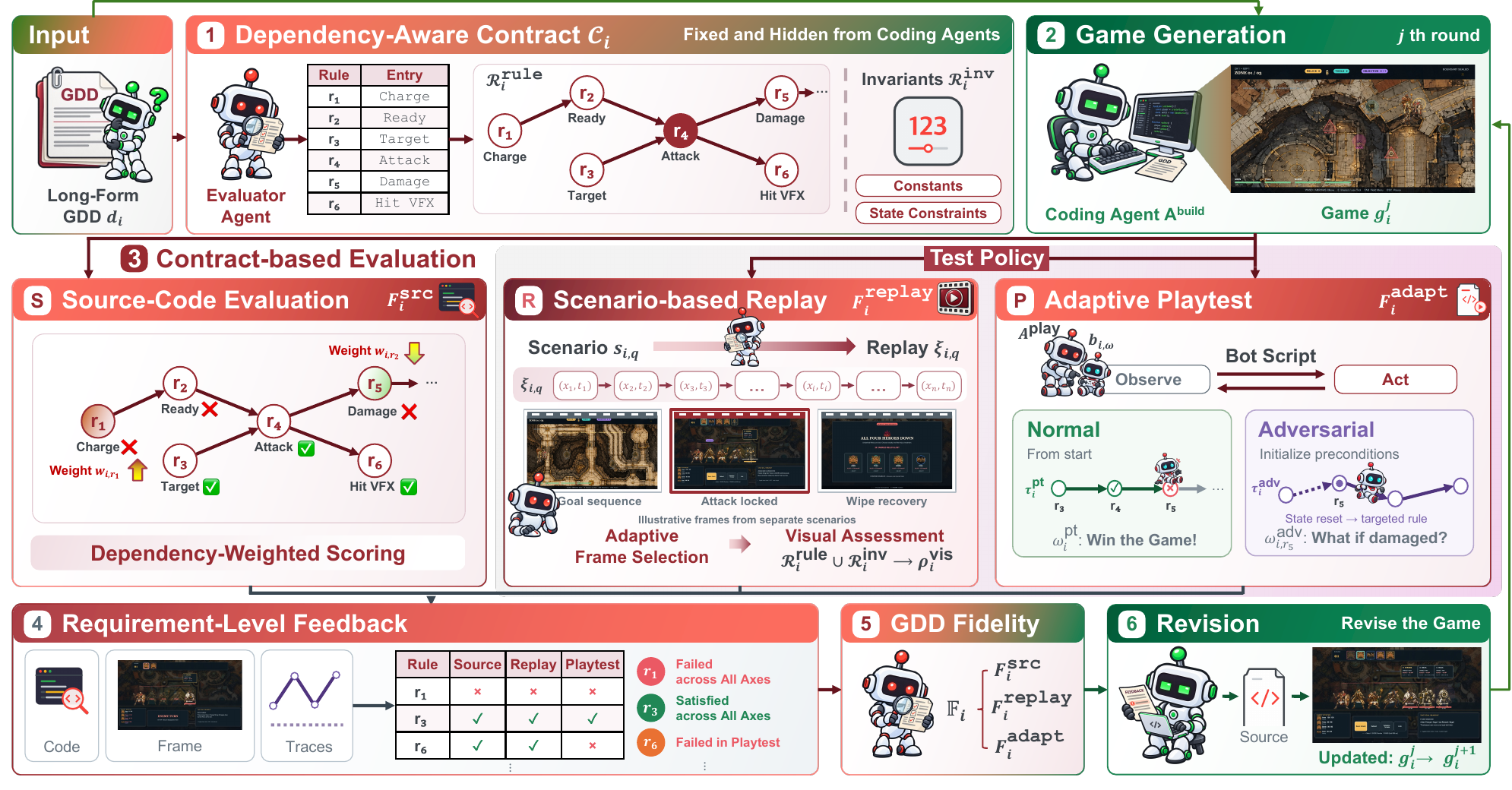}
    \vspace{-18pt}
    \caption{
        \textbf{Overall evaluation and revision pipeline of \AtoZbench{}.}
    }
    \label{fig:method_overview}
    \vspace{-5pt}
\end{figure*}

Our benchmark measures how faithfully coding agents implement a long-form GDD as a complete game. Rather than considering the GDD as a single free-form rubric, we first convert the GDD into a \textsc{Dependency-Aware Contract} that defines which requirements and dependencies must be verified. The contract is withheld from the building agent during initial generation. Then, our \textsc{Test Policies} specify how to exercise the game and collect evidence for them. Together, source-code inspection, scenario-based replay, and adaptive playtesting provide complementary judgments linked to the same requirements for~\emph{\GDDmetric} scoring and revision feedback, as illustrated in Figure~\ref{fig:method_overview}. We use a 2D single-player Phaser setting for controlled comparison; Section~\ref{sec:3d_extension} demonstrates the framework's extension to 3D games in Three.js.

\subsection{Benchmark Setting}
\label{sec:benchmark_setting}

Let $\mathcal{D}=\{d_i\}_{i=1}^{N}$ be a corpus of $N$ GDDs, with $\mathcal{U}_i$ denoting the requirements stated in $d_i$. These include conditional game rules, constraints on states and values, and expected visual responses. Given $d_i$ and a development environment $\Omega$, a game-building agent $A^{\mathrm{build}}$ produces
\begin{equation}
    g_i^0 = A^{\mathrm{build}}(d_i;\Omega),
    \label{eq:game_generation}
\end{equation}
where $g_i^0$ contains the source project and executable build. The agent receives the GDD $d_i$, while its implementation is assessed against an evaluation contract $\mathcal{C}_i$ constructed from that document and withheld during initial generation. We write $g_i$ when evaluating a single build and introduce revision rounds in Section~\ref{sec:evidence_feedback}.

\paragraph{Game Specifications and GDD Generation Protocol.}

We construct 100 GDDs in two stages: each game brief is first expanded into a creative vision (CV) defining the intended design, and then into a detailed GDD. Then, an agentic authoring pipeline checks each GDD for completeness and consistency. We split the GDDs by implementation scope: 50 \emph{Small} designs with compact scope and 50 \emph{Big} designs with broader content and interacting systems. \emph{Small} GDDs average 14,085 tokens, while \emph{Big} GDDs average 26,297 tokens. Further details of the generation procedure, validation, dataset statistics, and representative examples are in Appendix~\ref{app:gdd_dataset}.

\subsection{Dependency-Aware Contract}
\label{sec:dependency_contract}

We construct an evaluation contract from the requirements $\mathcal{U}_i$ in the GDD $d_i$:
\begin{equation}
    \mathcal{C}_i=(\mathcal{R}_i,\mathcal{E}_i),
    \qquad
    \mathcal{R}_i=\mathcal{R}_i^{\mathrm{rule}}\cup\mathcal{R}_i^{\mathrm{inv}},
    \label{eq:contract}
\end{equation}
where rules specify conditions, triggering events, and expected effects, while invariants specify constraints within a stated scope. The directed edges $\mathcal{E}_i$ connect rules when one updates a state referenced by another rule's preconditions or emits an event that triggers it. For example, in our game \texttt{traces\_left}, placing sensors changes the sensor count required by the `Run' rule, linking placement to subsequent execution. For this example, source inspection checks the placement and execution handlers, replay checks the visible response to the prescribed inputs, and adaptive playtesting checks whether the required state transition occurs during play. The dependency identifies which prerequisite to inspect when execution cannot start; each downstream rule still requires its own evidence for a verdict.

Contract construction first fixes a shared state vocabulary and rule entries bound to their supporting GDD rows. Extracted read/write facts determine the dependency skeleton, and rule generation fills in conditions, triggers, and effects within this structure. The accepted contract is frozen before comparing builds, so agents and revision rounds are evaluated against the same requirement definitions and dependencies. Appendix~\ref{app:gdd-contract-construction} gives the construction and acceptance procedures, and Section~\ref{sec:evaluation_reliability} reports consistency across repeated generations.

\subsection{Contract-Guided Evaluation}
\label{sec:contract_evaluation}

On average, each GDD specifies 69 outcome requirements and 32 invariants, and evaluation uses 20 replay scenarios. Generated games differ in their internal project structures, and many contract requirements concern rendered or interactive outcomes. We therefore use three complementary evaluation axes: source-code inspection, scenario-based replay, and adaptive playtesting. Source-code inspection checks how requirements are implemented, while replay and adaptive playtesting collect runtime evidence of their visual and interactive outcomes through \textsc{Test Policies}. A scenario-based replay policy specifies a fixed input sequence and its timing, whereas an adaptive playtest policy selects inputs in response to runtime observations. The contract defines \emph{what} to verify, and each test policy specifies \emph{how} to exercise the game to obtain the required evidence. Judgments from all three axes remain linked to the contract requirements for scoring and feedback.

\subsubsection{Source-Code Evaluation}
\label{sec:source_eval}

First, an evaluator agent inspects the source of $g_i$ against $\mathcal{C}_i$ without executing the game. For each rule, it checks the implementation of the specified conditions, triggers, and effects, assigning a completeness score $f_{i,r}^{\mathrm{src}}\in[0,1]$ with source-code references. Invariants use $f_{i,r}^{\mathrm{src}}\in\{0,1\}$ to record whether their constraints are preserved within the stated scope.

\paragraph{Dependency-Weighted Scoring.}
Because a failure in one rule can affect requirements that depend on its outputs, we account for each rule's downstream reach when aggregating source-code judgments. For scoring, we use the state-dependency links $\mathcal{E}_i^{\mathrm{state}}\subseteq\mathcal{E}_i$: one rule writes an attribute inspected by another rule's condition. These links represent potential prerequisite relations. Let $D_{i,r}$ count the downstream rules reachable from $r$ via these links, and let $D_i^{\max}=\max_{r\in\mathcal{R}_i^{\mathrm{rule}}}D_{i,r}$. For $D_i^{\max}>0$, the weight $w_{i,r}$ is defined as:
\begin{equation}
    w_{i,r}=\alpha+(1-\alpha)
    \frac{\log(1+D_{i,r})}{\log(1+D_i^{\max})}.
    \label{eq:dependency_weights}
\end{equation}
The parameter $\alpha\in[0,1]$ sets the weight floor: $\alpha=1$ yields uniform weights, while our default $\alpha=0.5$ gives a rule at most twice the weight of a rule with no downstream dependencies. We set all weights to $1$ when $D_i^{\max}=0$. Then, the source-code score combines the weighted rule score with the invariant pass rate:
\begin{equation}
    F_i^{\mathrm{src}}=F_i^{\mathrm{rule}}F_i^{\mathrm{inv}},
    \quad
    F_i^{\mathrm{rule}}=
    \frac{\sum_{r\in\mathcal{R}_i^{\mathrm{rule}}}w_{i,r}f_{i,r}^{\mathrm{src}}}
         {\sum_{r\in\mathcal{R}_i^{\mathrm{rule}}}w_{i,r}},
    \quad
    F_i^{\mathrm{inv}}=
    \frac{\sum_{r\in\mathcal{R}_i^{\mathrm{inv}}}f_{i,r}^{\mathrm{src}}}
         {|\mathcal{R}_i^{\mathrm{inv}}|}.
    \label{eq:source_score}
\end{equation}
Thus, rules with greater downstream reach receive greater influence on the source-code score, while invariant violations reduce it. Appendix~\ref{app:ablation-dependency-weight} analyzes this weighting.

\subsubsection{Scenario-Based Replay Assessment}
\label{sec:replay_eval}

Source inspection cannot determine whether the specified visual responses actually appear during execution. Therefore, we use controlled scenario replays to collect rendered evidence for the contract requirements. We derive a canonical scenario set
$\mathcal{S}_i=\{s_{i,q}\}_{q=1}^{Q_i}$ from $\mathcal{C}_i$, where each scenario groups requirements with a shared entry condition. 
After finalizing the build, $A^{\mathrm{build}}$ provides a fixed replay~\textsc{Test Policy} with a sequence of player inputs and their timing for each scenario $s_{i,q}$:
\begin{equation}
    \xi_{i,q}=[(x_n,t_n)]_{n=1}^{N_{i,q}},
    \label{eq:replay_policy}
\end{equation}
where $x_n$ is the input issued at time $t_n$. The policy is fixed before execution and does not change with runtime observations. We execute $\xi_{i,q}$ under $s_{i,q}$ and record the full replay with input events.

Following the rubric format and evaluation protocol of GameCraft-Bench~\citep{luo2026gamecraft}, we express replay-observable rules and invariants as a game-level visual rubric $\rho_i^{\mathrm{vis}}$, preserving their conditions and expected visible outcomes. Each item remains linked to its corresponding requirements. Every canonical scenario is replayed and scored against the fixed rubric using a build-specific input policy, and each judgment must cite observed frame evidence.

\paragraph{Adaptive Frame Selection.}
Fixed-rate sampling can miss brief responses, while a short temporal window can omit delayed outcomes. Therefore, we keep the full scenario replay and let the judge retrieve evidence relevant to $\rho_i^{\mathrm{vis}}$. Given $s_{i,q}$, $\rho_i^{\mathrm{vis}}$, and $\xi_{i,q}$, the judge identifies relevant trace events and requests their temporal intervals. Then, a deterministic tool maps these requests to recorded timestamps and returns up to $M$ frames. Using these frames, the multimodal judge scores each rubric item in $[0,1]$ with supporting visual evidence. We combine the item rewards across scenarios using the category weights and aggregation rules of GameCraft-Bench to obtain $F_i^{\mathrm{replay}}$.

\begin{table}[t]
    \centering
    \caption{
        \textbf{Main results on \AtoZbench.} Overall \GDDmetric{}, per-axis scores, and verifiable rates on 100 GDDs. \textbf{Bold} represents the best result, and \underline{underline} indicates the second-best.
    }
    \label{tab:main_results}
    \vspace{-5pt}

    \scriptsize
    \setlength{\tabcolsep}{1.6pt}
    \renewcommand{\arraystretch}{1.03}
    \setlength{\aboverulesep}{1.2pt}
    \setlength{\belowrulesep}{1.5pt}

    \providecommand{\MainSplitLabel}[1]{%
        {\fontsize{6.3}{7.2}\selectfont #1}%
    }

    \begin{tabularx}{\textwidth}{
        @{\hspace{2pt}}
        l
        |
        *{9}{>{\centering\arraybackslash}X}
        |
        *{2}{>{\centering\arraybackslash}X}
        @{\hspace{2pt}}
    }
        \toprule
        \multicolumn{1}{c}{
            \multirow[c]{1}{*}[-12pt]{\textbf{Model}}
        }
        & \multicolumn{3}{c}{
            \multirow[c]{2}{*}[-1.5pt]{\textbf{Overall GDD Fidelity}}
        }
        & \multicolumn{6}{c}{\textbf{Evaluation Axis}}
        & \multicolumn{2}{c}{
            \multirow[c]{2}{*}[-1.5pt]{\textbf{Verifiable Rate}}
        }
        \\
        \cmidrule(lr){5-10}
        \multicolumn{1}{c}{}
        & \multicolumn{3}{c}{}
        & \multicolumn{2}{c}{\textbf{Source}}
        & \multicolumn{2}{c}{\textbf{Replay}}
        & \multicolumn{2}{c}{\textbf{Playtest}}
        & \multicolumn{2}{c}{}
        \\
        \cmidrule(lr){2-4}
        \cmidrule(lr){5-6}
        \cmidrule(lr){7-8}
        \cmidrule(lr){9-10}
        \cmidrule(lr){11-12}
        \multicolumn{1}{c}{}
        & \MainSplitLabel{All}
        & \MainSplitLabel{\emph{Small}}
        & \MainSplitLabel{\emph{Big}}
        & \MainSplitLabel{\emph{Small}}
        & \MainSplitLabel{\emph{Big}}
        & \MainSplitLabel{\emph{Small}}
        & \MainSplitLabel{\emph{Big}}
        & \MainSplitLabel{\emph{Small}}
        & \multicolumn{1}{>{\centering\arraybackslash}X}{\MainSplitLabel{\emph{Big}}}
        & \MainSplitLabel{\emph{Small}}
        & \MainSplitLabel{\emph{Big}}
        \\
        \midrule

        \ModelWithIcon{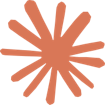}{Claude-Fable-5.1}
        & \textbf{77.0} & \textbf{82.8} & \textbf{71.1} & \textbf{89.7} & \textbf{83.5} & \textbf{73.5} & \textbf{55.1} & \textbf{85.2} & \textbf{74.7} & 98.7\% & 98.7\% \\

        \ModelWithIcon{claude}{Claude-Opus-5}
        & \underline{73.9} & \underline{80.9} & \underline{66.8} & \underline{87.9} & \underline{78.4} & \underline{71.2} & \underline{54.4} & \underline{83.6} & \underline{67.7} & \textbf{100.0\%} & \underline{99.3}\% \\

        \ModelWithIcon{claude}{Claude-Opus-4.8}
        & 56.8 & 67.9 & 45.6 & 72.5 & 49.6 & 62.5 & 41.4 & 68.7 & 46.0 & \underline{99.3\%} & \textbf{100.0\%} \\

        \ModelWithIcon{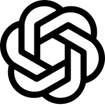}{GPT-6-Astra}
        & 71.6 & 78.5 & 64.6 & 84.2 & 74.4 & 69.9 & 52.1 & 81.6 & 67.4 & \underline{99.3\%} & 97.3\% \\

        \ModelWithIcon{gpt}{GPT-5.6-Sol}
        & 61.8 & 72.8 & 50.8 & 78.5 & 46.3 & 63.8 & 48.3 & 76.0 & 57.9 & 98.7\% & 98.0\% \\

        \ModelWithIcon{gpt}{GPT-5.5}
        & 63.2 & 73.1 & 53.4 & 81.0 & 55.9 & 60.8 & 42.5 & 77.5 & 61.7 & 97.3\% & 98.7\% \\

        \ModelWithIcon{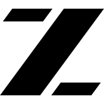}{GLM-5.3}
        & 55.6 & 66.9 & 44.3 & 78.6 & 58.1 & 57.3 & 34.5 & 64.9 & 40.4 & 88.7\% & 87.3\% \\

        \ModelWithIcon{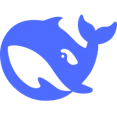}{DeepSeek-V4-Pro}
        & 50.4 & 65.3 & 35.6 & 75.7 & 53.8 & 52.2 & 25.9 & 68.0 & 27.0 & 88.0\% & 82.7\% \\

        \ModelWithIcon{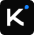}{Kimi-K2.7}
        & 39.5 & 51.4 & 27.5 & 60.8 & 37.3 & 47.7 & 19.0 & 45.7 & 26.2 & 75.3\% & 86.0\% \\

        \bottomrule
    \end{tabularx}
    \vspace{-10pt}
\end{table}

\subsubsection{Adaptive Playtest}
\label{sec:adaptive_playtest}

A fixed replay provides controlled visual evidence, but it cannot adjust its inputs to the runtime state of the game. Unlike the fixed replay policy, adaptive playtesting requires the~\textsc{Test Policy} to respond to the runtime observations. Therefore, we implement it as an executable program following \textsc{code-as-policy}~\citep{liang2023codeaspolicies}. An evaluator-side coding agent $A^{\mathrm{play}}$ reads a test objective $\omega$ and a shared interface $\mathrm{API}$ for observing game states and returning valid player inputs:
\begin{equation}
    b_{i,\omega}=A^{\mathrm{play}}(g_i,\omega,\mathrm{API}).
    \label{eq:playtest_bot}
\end{equation}
The bot uses conditions and loops to select its next input from runtime observations. Executing it from an initial state $z$ produces a trace $\tau_i(\omega,z)$ as below:
\begin{equation}
    \tau_i(\omega,z)=[(o_t,x_t)]_{t=1}^{T_i},
    \label{eq:playtest_trace}
\end{equation}
where $o_t$ is the observation at step $t$ and $x_t$ is the selected player input. The trace records the states, inputs, events, and timestamps used for requirement-level judgment.

\paragraph{Normal Playtest.}
First, the objective $\omega_i^{\mathrm{pt}}$ asks the bot to complete $g_i$ from its default state $z_i^0$:
\begin{equation}
    \tau_i^{\mathrm{pt}}
    =\tau_i(\omega_i^{\mathrm{pt}},z_i^0).
    \label{eq:normal_playtest}
\end{equation}
Only player inputs are allowed, with no intermediate state initialization. This setting tests which contract requirements can be reached and verified through ordinary gameplay. Let $\mathcal{R}_i^{\mathrm{pt}}\subseteq\mathcal{R}_i^{\mathrm{rule}}$ denote the verified rules with conclusive evidence from this trace.

\paragraph{Adversarial Playtest.}
Normal play may leave requirements unverified because their preconditions are not reached. Therefore, we apply targeted adversarial tests to
$\mathcal{R}_i^{\mathrm{adv}}
\subseteq
\mathcal{R}_i^{\mathrm{rule}}\setminus\mathcal{R}_i^{\mathrm{pt}}$ for checking edge cases that were not reached during normal play.
For each rule $r$, we initialize the game to a state $z_{i,r}$ satisfying its preconditions and give the bot an objective $\omega_{i,r}^{\mathrm{adv}}$ for testing its expected effects:
\begin{equation}
    z_{i,r}\models\phi_r,
    \qquad
    \tau_{i,r}^{\mathrm{adv}}
    =\tau_i(\omega_{i,r}^{\mathrm{adv}},z_{i,r}).
    \label{eq:adversarial_playtest}
\end{equation}
Initialization supplies the target rule's preconditions rather than its expected effects; assisted attempts are marked in the trace. After initialization, the protocol requires player inputs through the same interface as normal play, without further state injection. Before-trigger snapshots establish the situation and do not by themselves establish the outcome; the judge checks the subsequent trigger and effects against the recorded execution. Appendix~\ref{app:playtest_protocol} details the initialization and trace protocol.

\paragraph{Judgment And Scoring.}
A separate evaluator compares each tested rule's conditions, trigger, and expected effects with the recorded trace. A rule is \emph{satisfied} when the evidence supports its specified effects and \emph{violated} when the observed outcome contradicts them. Rules whose required situation is not established remain \emph{unverified}. Each rule receives three judgments, and the strict majority determines its verdict. Conclusive normal-play verdicts are retained, while adversarial tests extend runtime verification to rules left unverified during normal play. The adaptive-play score $F_i^{\mathrm{adapt}}$ is
\begin{equation}
    F_i^{\mathrm{adapt}}=
    \frac{\sum_{r\in\mathcal{R}_i^{\mathrm{rule}}}f_{i,r}^{\mathrm{adapt}}}
         {|\mathcal{R}_i^{\mathrm{rule}}|},
    \label{eq:playtest_score}
\end{equation}
where $f_{i,r}^{\mathrm{adapt}}=1$ only when $r$ is confirmed as \emph{satisfied} and $0$ otherwise. We separately report \emph{judgment coverage}, the fraction of all rules judged either satisfied or violated, to measure the extent of conclusive execution evidence. 
Detailed interface, judgment, and aggregation procedures are provided in Appendix~\ref{app:playtest_protocol}.

%
\begingroup

\usetikzlibrary{shapes.callouts, arrows.meta}

\providecommand{\AtoZFigFont}{\sffamily}
\providecolor{A2ZRed}{HTML}{E65042}
\providecolor{A2ZFigText}{HTML}{3F454B}
\providecolor{A2ZFigAxis}{HTML}{C9CED3}
\providecolor{A2ZFigRest}{HTML}{F0F2F4}

\colorlet{AtoZCaseAccent}{A2ZRed!78!black}
\definecolor{AtoZCardFill}{HTML}{F4F5F7}

\colorlet{AtoZQualMuted}{A2ZFigText!83!white}%

\begin{figure*}[!t]
\centering

\renewcommand{\thesubfigure}{(\alph{subfigure})}
\captionsetup{subrefformat=simple}
\captionsetup[subfigure]{font=scriptsize,labelfont=bf,labelformat=simple,labelsep=space,justification=centering,singlelinecheck=false,position=bottom,skip=0pt,hypcap=true}

\begin{subfigure}[t]{0.55\textwidth}
\vspace{0pt}\centering
\begingroup

\resizebox{0.98\linewidth}{!}{%
\begin{tikzpicture}[
    x=1cm,y=1cm,
    font=\AtoZFigFont\tiny,
    every node/.style={inner sep=0pt},
    pretty arrow/.style={
        -{Stealth[length=1.5mm, width=1.4mm]},
        draw=AtoZCaseAccent,
        line width=0.85pt
    }
]

\newcommand{\AtoZCaseHeader}[3]{%
\path[fill=white,draw=A2ZFigAxis,line width=0.28pt,rounded corners=1.5pt] (-0.56,#1-1.50) rectangle (7.88,#1-0.30);
\path[fill=A2ZFigRest!75!white,rounded corners=1.1pt] (-0.53,#1-1.47) rectangle (0.04,#1-0.33);
\node[rotate=90,align=center,text width=1.08cm,font=\AtoZFigFont\fontsize{5.6}{6.1}\selectfont] at (-0.245,#1-0.90)
 {\textcolor{AtoZCaseAccent}{\textbf{#2}}\\[-0.2pt]{\fontsize{3.5}{4.0}\selectfont\ttfamily\bfseries\textcolor{A2ZFigText}{#3}}};
}

\newcommand{\AtoZTextCard}[5]{%
    \path[fill=AtoZCardFill,draw=A2ZFigAxis,line width=0.28pt,rounded corners=1.1pt]
        (#1,#2) rectangle (#1+#3,#2+#4);

    \begin{scope}[shift={(#1+0.06, #2+#4-0.19)}]
        \draw[AtoZQualMuted,line width=0.35pt,line join=round]
            (0,0.13) -- (0,0) -- (0.10,0) -- (0.10,0.095) -- (0.065,0.13) -- cycle;
        \draw[AtoZQualMuted,line width=0.30pt]
            (0.065,0.13) -- (0.065,0.095) -- (0.10,0.095)
            (0.02,0.05) -- (0.08,0.05)
            (0.02,0.025) -- (0.07,0.025);
    \end{scope}

    \node[anchor=north west,text width=#3cm-0.26cm,align=left,text=A2ZFigText,
        font=\AtoZFigFont\fontsize{3.5}{4.0}\rmfamily\selectfont]
        at (#1+0.20,#2+#4-0.06) {#5};
}

\newsavebox{\AtoZImageMeasure}
\newcommand{\AtoZImageCard}[5]{%
    \sbox{\AtoZImageMeasure}{\includegraphics[width=#3cm]{assets/images/complementarity/#5}}%
    \begin{scope}
        \clip[rounded corners=0pt] (#1,#2) rectangle (#1+#3,#2+#4);
        \node[anchor=center] at (#1+#3/2+0.36,#2+#4/2)
            {\ifdim\dimexpr\ht\AtoZImageMeasure+\dp\AtoZImageMeasure\relax<#4cm
                \includegraphics[height=#4cm]{assets/images/complementarity/#5}%
             \else
                \usebox{\AtoZImageMeasure}%
             \fi};
    \end{scope}
    \draw[draw=A2ZFigAxis,line width=0.28pt,rounded corners=0pt]
        (#1,#2) rectangle (#1+#3,#2+#4);
}

\newcommand{\AtoZSmallLabel}[4]{%
    \node[text=#4,font=\AtoZFigFont\fontsize{5.5}{6.0}\selectfont\bfseries]
        at (#1,#2) {#3};
}

\tikzset{AtoZCalloutPaint/.style={fill=white,fill opacity=0.92,text opacity=1,draw=AtoZCaseAccent,line width=0.35pt,rounded corners=1.2pt}}
\newcommand{\AtoZBubble}[5]{%
    \node[rectangle callout, AtoZCalloutPaint, 
          inner xsep=2.0pt, inner ysep=1.5pt, 
          callout absolute pointer={#3}, callout pointer width=0.16cm, 
          anchor=#5, align=center, text=AtoZCaseAccent, 
          font=\AtoZFigFont\fontsize{4.4}{5.0}\selectfont\bfseries]
        at (#1, #2) {#4};
}

\def\cardW{2.42}
\def\cardH{0.88}
\def\xA{0.13}
\def\xB{2.74}
\def\xC{5.35}

\def\yA{2.20} 
\def\yB{0.67} 
\def\yC{-0.86} 

\begin{scope}[xshift=0.58cm,yshift=-0.70cm]
\AtoZCaseHeader{3.45}{Source}{\texttt{Chalk Escape}}

\AtoZTextCard{\xA}{\yA}{\cardW}{\cardH}{
\textbf{GDD Requirement}\\[0.5pt]
An invalid start press makes the start-zone \uline{outline pulse \textbf{twice}}.}

\path[fill=black!96,draw=A2ZFigAxis,line width=0.28pt] (\xB,\yA) rectangle (\xB+\cardW,\yA+\cardH);
\node[anchor=north west,text=green!22!white,align=left,font=\ttfamily\fontsize{4.1}{4.6}\selectfont] at (\xB+0.07,\yA+\cardH-0.06) {\textbf{Code}\\[0.5pt]> 4 * Math.abs(\\\hspace*{0.1cm}Math.sin(...\\\hspace*{0.1cm}* Math.PI * 4))};
\AtoZBubble{4.99}{2.22}{(3.89,2.68)}{This makes \textbf{four} peaks.}{south east}

\AtoZImageCard{\xC}{\yA}{\cardW}{\cardH}{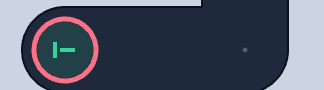}

\draw[pretty arrow]
    (\xB+\cardW+0.02,\yA+\cardH/2) -- (\xC-0.02,\yA+\cardH/2);

\AtoZBubble{7.62}{2.30}{(6.12, 2.66)}{Rendered pulse\\[-0.5pt]expands more\\[-0.5pt]than twice}{south east}

\AtoZSmallLabel{\xA+\cardW/2}{\yA-0.13}{Specification}{A2ZFigText}
\AtoZSmallLabel{\xB+\cardW/2}{\yA-0.13}{Implementation}{A2ZFigText}
\AtoZSmallLabel{\xC+\cardW/2}{\yA-0.13}{Rendered Output}{AtoZCaseAccent}

\end{scope}
\begin{scope}[xshift=0.58cm,yshift=-0.35cm]
\AtoZCaseHeader{1.92}{Replay}{\texttt{Neon Spray}}

\AtoZTextCard{\xA}{\yB}{\cardW}{\cardH}{
\textbf{GDD Requirement}\\[0.5pt]
The pause menu should show clear pause controls, including a \uline{readable \textbf{settings}} entry.}

\begin{scope}
    \clip[rounded corners=0.0pt] (\xB,\yB) rectangle (\xB+\cardW,\yB+\cardH);
    \node[anchor=north] at (\xB+\cardW/2,\yB+\cardH)
        {\includegraphics[width=\cardW cm]{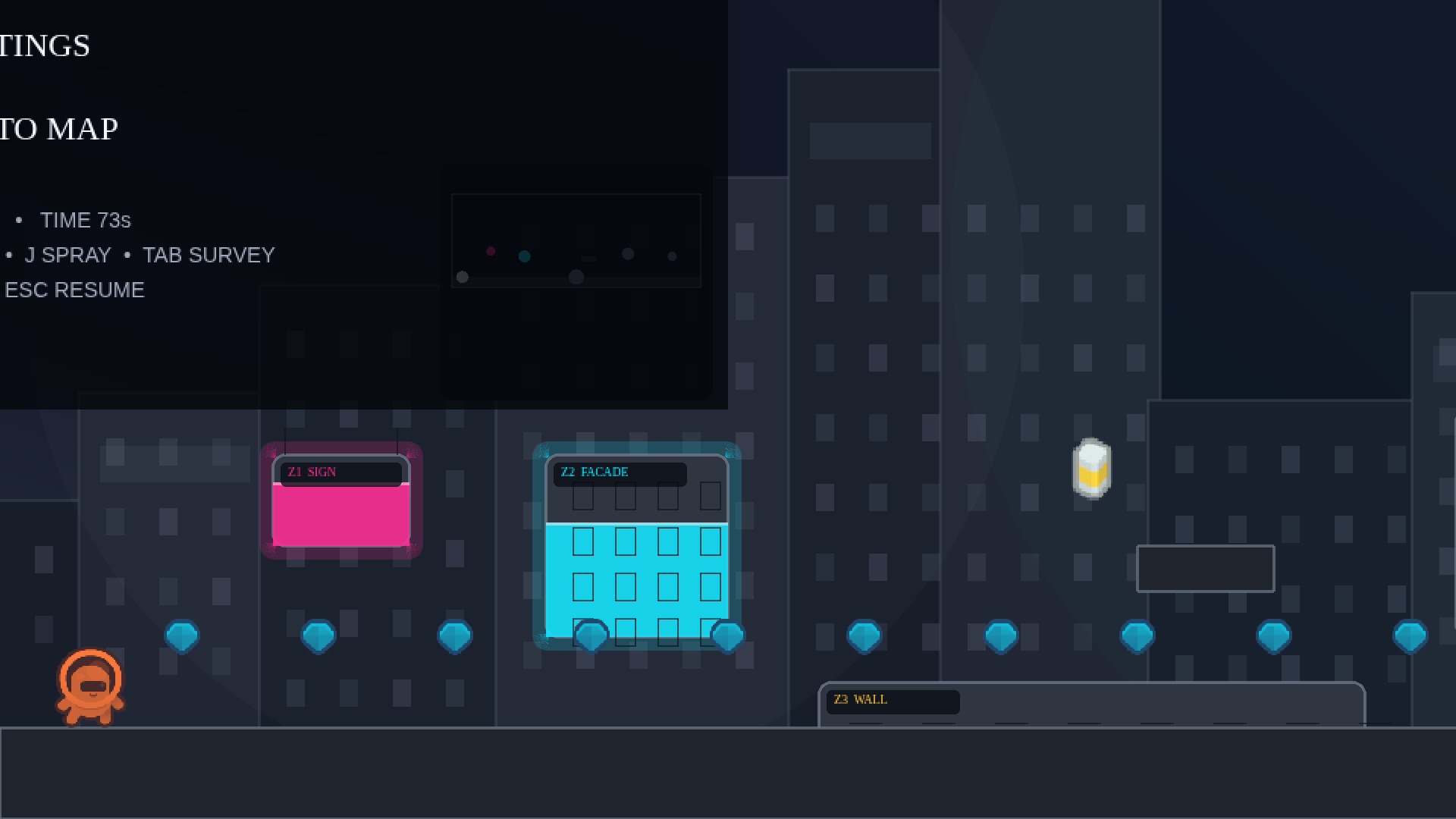}};
\end{scope}
\draw[draw=A2ZFigAxis,line width=0.28pt,rounded corners=0pt]
    (\xB,\yB) rectangle (\xB+\cardW,\yB+\cardH);

\begin{scope}
    \clip[rounded corners=0pt] (\xC,\yB) rectangle (\xC+\cardW,\yB+\cardH);
    \node[anchor=north west,overlay] at (\xC,\yB+\cardH+0.16)
        {\includegraphics[width=9.50cm]{assets/images/complementarity/neon_pause.png}};
\end{scope}
\draw[draw=A2ZFigAxis,line width=0.28pt,rounded corners=0pt]
    (\xC,\yB) rectangle (\xC+\cardW,\yB+\cardH);

\draw[draw=AtoZCaseAccent,line width=0.42pt]
    (\xB+0.01,\yB+\cardH-0.03) rectangle (\xB+0.40,\yB+\cardH-0.27);

\draw[pretty arrow]
    (\xB+\cardW+0.02,\yB+\cardH/2) -- (\xC-0.02,\yB+\cardH/2);

\node[rectangle,draw=none,fill=none,text opacity=0,inner xsep=2pt,inner ysep=1.5pt,anchor=south east,align=center,font=\AtoZFigFont\fontsize{4.4}{5.0}\selectfont\bfseries] (AtoZSettingsBubble) at (\xC+\cardW-0.05,\yB+0.38) {"SETTINGS" text clipped}; \path[AtoZCalloutPaint]  (AtoZSettingsBubble.north west) -- ([xshift=0.36cm]AtoZSettingsBubble.north west) [sharp corners]  -- (\xC+0.66,\yB+\cardH-0.14) -- ([xshift=0.52cm]AtoZSettingsBubble.north west) [rounded corners=1.2pt]  -- (AtoZSettingsBubble.north east) -- (AtoZSettingsBubble.south east) -- (AtoZSettingsBubble.south west) -- cycle; \node[anchor=center,inner sep=0pt,text=AtoZCaseAccent,font=\AtoZFigFont\fontsize{4.4}{5.0}\selectfont\bfseries] at (AtoZSettingsBubble.center) {"SETTINGS" text clipped};

\AtoZSmallLabel{\xA+\cardW/2}{\yB-0.13}{Specification}{A2ZFigText}
\AtoZSmallLabel{\xB+\cardW/2}{\yB-0.13}{Selected Replay Frame}{A2ZFigText}
\AtoZSmallLabel{\xC+\cardW/2}{\yB-0.13}{Visual Evidence}{AtoZCaseAccent}

\end{scope}
\begin{scope}[xshift=0.58cm]
\AtoZCaseHeader{0.39}{Playtest}{\texttt{Mobile Balance}}

\AtoZTextCard{\xA}{\yC}{\cardW}{\cardH}{
\textbf{GDD Requirement}\\[0.5pt]
After the drag button is released, the hanging weight should \uline{\textbf{stop} following the pointer}.}

\AtoZImageCard{\xB}{\yC}{\cardW}{\cardH}{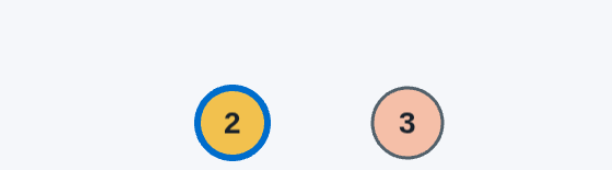}
\AtoZImageCard{\xC}{\yC}{\cardW}{\cardH}{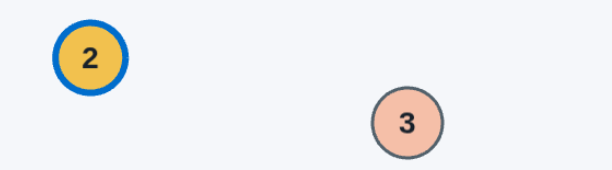}

\draw[pretty arrow]
    (\xB+\cardW+0.02,\yC+\cardH/2) -- (\xC-0.02,\yC+\cardH/2);

\draw[draw=AtoZCaseAccent,densely dashed,line width=0.40pt]
    (\xC+1.17,\yC+0.27) circle (0.21);
\draw[->,draw=AtoZCaseAccent,line width=0.50pt]
    (\xC+0.98,\yC+0.38) -- (\xC+0.49,\yC+0.57);

\node[rectangle,draw=none,fill=none,text opacity=0,inner xsep=2pt,inner ysep=1.5pt,anchor=north east,align=center,font=\AtoZFigFont\fontsize{4.4}{5.0}\selectfont\bfseries] (AtoZMovingBubble) at (7.72,\yC+\cardH-0.05) {Still moving};
\path[AtoZCalloutPaint]
 (AtoZMovingBubble.north west) -- (AtoZMovingBubble.north east) -- (AtoZMovingBubble.south east)
 -- ([xshift=0.22cm]AtoZMovingBubble.south west) [sharp corners]
 -- (\xC+1.20,\yC+0.44) -- ([xshift=0.06cm]AtoZMovingBubble.south west) [rounded corners=1.2pt]
 -- (AtoZMovingBubble.south west) -- cycle;
\node[anchor=center,inner sep=0pt,text=AtoZCaseAccent,font=\AtoZFigFont\fontsize{4.4}{5.0}\selectfont\bfseries] at (AtoZMovingBubble.center) {Still moving};

\AtoZSmallLabel{\xA+\cardW/2}{\yC-0.13}{Specification}{A2ZFigText}
\AtoZSmallLabel{\xB+\cardW/2}{\yC-0.13}{After Release}{A2ZFigText}
\AtoZSmallLabel{\xC+\cardW/2}{\yC-0.13}{Trace Evidence}{AtoZCaseAccent}

\end{scope}

\pgfresetboundingbox
\path[use as bounding box] (0,-1.451) rectangle (8.48,2.47);
\end{tikzpicture}%
}
\endgroup
\par
\vspace{-5pt}
\caption{Evidence across Axes}
\label{fig:evidence_axes}
\end{subfigure}%
\hfill
\begin{subfigure}[t]{0.245\textwidth}
\vspace{-5pt}\centering

\colorlet{A2ZOutcomeSat}{A2ZRed!55!white}
\colorlet{A2ZOutcomeFail}{A2ZRed!78!black}
\definecolor{A2ZOutcomeTrack}{HTML}{E2E5E8}

\resizebox{\linewidth}{!}{%
\begin{tikzpicture}[
    x=1cm,y=1.1087cm, 
    font=\AtoZFigFont\tiny,
    every node/.style={inner sep=0pt}
]
\path[use as bounding box] (0,-0.60) rectangle (5.35,4.305);

\def\AtoZBarStart{1.08}
\def\AtoZBarEnd{5.20}
\pgfmathsetmacro{\AtoZBarWidth}{\AtoZBarEnd-\AtoZBarStart}
\def\AtoZPctX#1{\AtoZBarStart+\AtoZBarWidth*(#1)/100}

\def\AtoZFigThreeBRow#1#2#3#4{%
    \pgfmathsetmacro{\AtoZSatEnd}{\AtoZPctX{#3}}%
    \pgfmathsetmacro{\AtoZFailEnd}{\AtoZPctX{100}}%
    \pgfmathsetmacro{\AtoZSatMid}{(\AtoZBarStart+\AtoZSatEnd)/2}%
    \pgfmathsetmacro{\AtoZFailMid}{(\AtoZSatEnd+\AtoZFailEnd)/2}%

    \node[
        anchor=east,align=right,text=A2ZFigText,
        font=\AtoZFigFont\fontsize{7.5}{8.0}\selectfont\bfseries
    ] at (0.94,#1) {#2};

    \path[fill=A2ZOutcomeTrack,rounded corners=1.4pt]
        (\AtoZBarStart,#1-0.15) rectangle (\AtoZBarEnd,#1+0.15);

    \begin{scope}
        \clip[rounded corners=1.4pt]
            (\AtoZBarStart,#1-0.15) rectangle (\AtoZBarEnd,#1+0.15);
        \shade[left color=A2ZOutcomeSat!92!black,right color=A2ZOutcomeSat!72!white]
            (\AtoZBarStart,#1-0.15) rectangle (\AtoZSatEnd,#1+0.15);
        \shade[left color=A2ZOutcomeFail!92!black,right color=A2ZOutcomeFail!72!white]
            (\AtoZSatEnd,#1-0.15) rectangle (\AtoZFailEnd,#1+0.15);
    \end{scope}

    \draw[draw=A2ZFigAxis,line width=0.25pt,rounded corners=1.4pt]
        (\AtoZBarStart,#1-0.15) rectangle (\AtoZBarEnd,#1+0.15);

    \node[text=A2ZFigText,font=\AtoZFigFont\fontsize{6.3}{7}\selectfont\bfseries]
        at (\AtoZSatMid,#1+0.29)
        {\pgfmathprintnumber[fixed,fixed zerofill,precision=1,assume math mode=true]{#3}};
    \node[text=A2ZFigText,font=\AtoZFigFont\fontsize{6.3}{7}\selectfont\bfseries]
        at (\AtoZFailMid,#1+0.29)
        {\pgfmathprintnumber[fixed,fixed zerofill,precision=1,assume math mode=true]{#4}};
}

\node[text=A2ZFigText,font=\AtoZFigFont\fontsize{8.2}{9}\selectfont\bfseries\rmfamily]
    at (3.14,3.93) {Playtest Outcomes (\%)};
\node[text=A2ZFigText,align=center,
    font=\AtoZFigFont\fontsize{7.4}{8}\selectfont\bfseries]
    at (0.52,3.50) {Source-Code\\Item Score};

\path[fill=A2ZOutcomeSat,rounded corners=0.6pt]
    (1.78,3.40) rectangle (2.04,3.52);
\node[anchor=west,text=A2ZFigText,
    font=\AtoZFigFont\fontsize{8.4}{9}\selectfont]
    at (2.22,3.46) {Score = 1};

\path[fill=A2ZOutcomeFail,rounded corners=0.6pt]
    (3.58,3.40) rectangle (3.84,3.52);
\node[anchor=west,text=A2ZFigText,
    font=\AtoZFigFont\fontsize{8.4}{9}\selectfont]
    at (4.02,3.46) {Score = 0};

\AtoZFigThreeBRow{2.40}{Full\\$(1)$}{76.768388}{23.231612}

\AtoZFigThreeBRow{1.30}{Partial\\$(0,1)$}{54.314416}{45.685584}

\AtoZFigThreeBRow{0.20}{Zero\\$(0)$}{25.651631}{74.348368}

\draw[draw=A2ZFigAxis,line width=0.34pt]
    (\AtoZBarStart,-0.26)--(\AtoZBarEnd,-0.26); 
\foreach \value in {0,50,100}{
    \pgfmathsetmacro{\AtoZXTick}{\AtoZPctX{\value}}%
    \draw[draw=A2ZFigAxis,line width=0.34pt]
        (\AtoZXTick,-0.30)--(\AtoZXTick,-0.22);
    \node[anchor=north,text=A2ZFigText,
        font=\AtoZFigFont\fontsize{6.0}{7}\selectfont]
        at (\AtoZXTick,-0.34) {\value};
}
\end{tikzpicture}%
}
\par
\vspace{0pt}
\caption{Playtest Confirmation Gap}
\label{fig:runtime_confirmation}
\end{subfigure}%
\hfill
\begin{subfigure}[t]{0.185\textwidth}
\vspace{-5pt}\centering
\begingroup
\colorlet{A2ZOrderingPaletteRed}{A2ZRed}
\colorlet{A2ZOrderingSource}{A2ZOrderingPaletteRed!76!white}
\colorlet{A2ZOrderingReplay}{A2ZOrderingPaletteRed!60!white}
\colorlet{A2ZOrderingPlaytest}{A2ZOrderingPaletteRed!75!black}
\resizebox{\linewidth}{!}{%
\begin{tikzpicture}[x=1cm,y=1.0539cm,font=\AtoZFigFont\tiny,every node/.style={inner sep=0pt}]
\path[use as bounding box] (0,-0.60) rectangle (3.84,4.305);

\node[text=A2ZFigText,font=\AtoZFigFont\fontsize{8.2}{9.0}\selectfont\bfseries\rmfamily]
    at (1.99,3.93) {Reversed Orderings (\%)};

\node[text=A2ZFigText,font=\AtoZFigFont\fontsize{8.4}{9.0}\selectfont\bfseries]
    at (2.07,3.52) {Excluded Axis};

\def\AtoZOrderingBase{-0.26}
\def\AtoZOrderingScale{0.158}

\foreach \v in {0,10,20}{%
    \pgfmathsetmacro{\yy}{\AtoZOrderingBase+\AtoZOrderingScale*\v}%
    \draw[draw=A2ZFigAxis!80!white,line width=0.30pt]
        (0.48,\yy)--(3.66,\yy);
    \node[anchor=east,text=A2ZFigText,font=\AtoZFigFont\fontsize{5.8}{6.4}\selectfont]
        at (0.38,\yy) {\v};
}

\newcommand{\AtoZOrderingBar}[5]{%
    \pgfmathsetmacro{\yy}{\AtoZOrderingBase+\AtoZOrderingScale*(#2)}%
    \shade[bottom color=#4,top color=#5]
        (#1-0.39,\AtoZOrderingBase) rectangle (#1+0.39,\yy);
    \draw[draw=A2ZFigAxis,line width=0.28pt]
        (#1-0.39,\AtoZOrderingBase) rectangle (#1+0.39,\yy);
    \node[anchor=south,text=A2ZFigText,
        font=\AtoZFigFont\fontsize{7.0}{7.6}\selectfont\bfseries\boldmath]
        at (#1,\yy+0.09) {\,#2};
    \node[text=A2ZFigText,
        font=\AtoZFigFont\fontsize{6.3}{6.9}\selectfont]
        at (#1,-0.43) {#3};
}

\AtoZOrderingBar{0.98}{20.5}{Source}
    {A2ZOrderingSource!84!black}{A2ZOrderingSource!72!white}
\AtoZOrderingBar{2.07}{8.9}{Replay}
    {A2ZOrderingReplay!82!black}{A2ZOrderingReplay!62!white}
\AtoZOrderingBar{3.16}{11.2}{Playtest}
    {A2ZOrderingPlaytest!88!black}{A2ZOrderingPlaytest!70!white}

\draw[draw=A2ZFigAxis,line width=0.34pt]
    (0.48,\AtoZOrderingBase)--(3.66,\AtoZOrderingBase);

\end{tikzpicture}%
}
\endgroup
\par
\vspace{0pt}
\caption{Axis Omission}
\label{fig:dependency_context}
\end{subfigure}%

\vspace{-5pt}
\caption{
    \textbf{Complementarity of source-code evaluation, scenario-based replay, and adaptive playtesting.}
    \subref{fig:evidence_axes} Examples of specification violations identified through different evidence traces.
    \subref{fig:runtime_confirmation} Playtest judgment outcomes grouped by source-code score.
    \subref{fig:dependency_context} Ratio of reversed GDD fidelity rank orderings across all 100 evaluated games built by GPT-5.6-Sol upon omission of individual evaluation axes.
}
\label{fig:complementary_axis}
\vspace{-5pt}
\end{figure*}
\endgroup

\subsection{Requirement-Level Evidence and Feedback}
\label{sec:evidence_feedback}

For revision round $j$, including the initial build at $j=0$, we calculate overall~\emph{\GDDmetric{}} as
\begin{equation}
    \mathbf{F_i^j}=
    \frac{F_i^{\mathrm{src},j}+F_i^{\mathrm{replay},j}+F_i^{\mathrm{adapt},j}}{3}.
    \label{eq:overall_fidelity}
\end{equation}
We report the three axis scores separately and give each evidence channel equal weight in overall~\GDDmetric{}, without fitting weights to the evaluated agents. To analyze agent performance during execution, we also define \emph{runtime fidelity} as the equal-weight mean $F_i^{\mathrm{run},j}=(F_i^{\mathrm{replay},j}+F_i^{\mathrm{adapt},j})/2$ of the scenario-based replay and adaptive-playtest scores. Equivalently, $\mathbf{F_i^j}=(F_i^{\mathrm{src},j}+2F_i^{\mathrm{run},j})/3$. Appendix~\ref{app:axis_weight_sensitivity} examines alternative axis weights while retaining the default dependency weighting within source-code evaluation.

The score $\mathbf{F_i^j}$ supports agent comparison, while the requirement-level judgments and supporting evidence such as code, frames, and traces identify where the implementation differs from the design. These results form feedback $h_i^j$ for the next revision based on $g_i^j$:
\begin{equation}
    g_i^{j+1}=A^{\mathrm{build}}(d_i,g_i^j,h_i^j;\Omega).
    \label{eq:game_revision}
\end{equation}
We assess the revised build against the same contract using all three axes, keeping the evaluation target fixed while measuring changes in specification faithfulness.

\section{Experiments}
\label{sec:experiment}

\subsection{Experimental Setup}
\label{sec:experimental_setup}

We evaluate coding-agent configurations based on Claude-Fable-5.1~\citep{anthropic2026claudecode}, Claude-Opus-5~\citep{anthropic2026claudecode}, Claude-Opus-4.8~\citep{anthropic2026claudecode}, GPT-6-Astra~\citep{openai2026gpt6astra}, GPT-5.6-Sol~\citep{openai2026gpt56}, GPT-5.5~\citep{openai2026gpt55}, Kimi-K2.7~\citep{moonshot2025kimik2}, GLM-5.3~\citep{glm5team2026glm5}, and DeepSeek-V4-Pro~\citep{deepseekai2026deepseekv4}. Each configuration is evaluated on all 100 GDDs. For each GDD, the same contract and replay scenarios are used across all outputs.

Unless otherwise stated, evaluator-side agents use GPT-5.6-Luna at high reasoning effort. Evaluator-model comparisons and cost analyses are provided in Appendix~\ref{app:cost_evaluation}, with further analysis of scenario-based replay assessment in Appendix~\ref{app:ablation-mllm}. 
Each configuration generates three builds per GDD, and each build is independently judged three times, with axis-specific aggregation described in Appendix~\ref{app:eval_details}.
We report overall~\emph{\GDDmetric}, the source-code, replay, and adaptive-play scores, and the \emph{verifiable rate} defined as the average compile and runtime pass rate across the three axes. 
Benchmark fidelity scores are reported on a 0--100 scale. Further details on agent configurations, computational resources, and evaluation settings are provided in Appendix~\ref{app:experimental_details}.

\subsection{Main Benchmark Results}
\label{sec:main_results}

As shown in Table~\ref{tab:main_results}, Claude-Fable-5.1 achieves the highest overall~\GDDmetric{} of 77.0 across all 100 GDDs, followed by Claude-Opus-5 at 73.9. Across the evaluated agents, mean overall~\GDDmetric{} is 71.1 on \emph{Small} and 51.1 on \emph{Big}, while the corresponding verifiable rates are 93.9\% and 94.2\%. Although most games can be compiled and executed under our evaluation, high verifiability does not imply faithful implementation of the intended design. In addition, implementing games from \emph{Big} GDDs is consistently more difficult across all three evaluation axes than implementing games from \emph{Small} GDDs. Overall score decreases by 20.0 points from \emph{Small} to \emph{Big}. The average source-code, scenario-based replay, and adaptive playtest scores similarly decrease by 19.1, 20.6, and 20.3 points, which shows that difficulty of broader specifications is not confined to source implementation, visual rendering, or runtime behavior alone.

Figure~\ref{fig:source_execution} compares the nine agents by source-code score and runtime fidelity, using the average of scenario-based replay and adaptive-playtest scores defined in Section~\ref{sec:evidence_feedback}. Both coordinates follow the aggregation protocol of Table~\ref{tab:main_results}, and dashed lines indicate overall~\GDDmetric{}. Claude-Fable-5.1 leads on both coordinates, while other agents exhibit different performance profiles. Specifically, GPT-5.5 and GLM-5.3 have nearly equal source-code scores (68.5 vs.\ 68.4), but their runtime fidelity differs by 11.4 points (60.6 vs.\ 49.3). Figure~\ref{fig:source_execution}\subref{fig:source_execution_big} shows a sharper contrast on \emph{Big}: GLM-5.3 has a higher source-code score (58.1 vs.\ 55.9), while GPT-5.5 has higher runtime fidelity (52.1 vs.\ 37.4). These differences show how evaluation through Test Policies distinguishes agents with comparable source-code scores by assessing their rendered outcomes and interactive behavior. Further pairwise comparisons are provided in Appendix~\ref{app:pairwise_elo}.

\input{figures/exp_source_execution}

\subsection{Complementarity of the Evaluation Axes}
\label{sec:axis_complementarity}

Figure~\ref{fig:complementary_axis} demonstrates that the three axes capture different failures for the same specification. Figure~\ref{fig:evidence_axes} presents violations identified through different evidence channels: an incorrect pulse count in the implementation, clipped menu text in replay frames, and continued object movement after input release during playtesting.
As shown in Figure~\ref{fig:runtime_confirmation}, even among rules that pass with a full source-code score, only 76.8\% are confirmed as satisfied during playtesting, while 23.2\% do not receive a full score. In contrast, 25.7\% of rules receiving zero source-code score are judged as satisfied during playtesting. These disagreements show that implementation-level judgments and runtime observations provide different evidence on requirement satisfaction. Additionally, Figure~\ref{fig:dependency_context} reports ordering reversals across the 100 games generated by GPT-5.6-Sol when each axis is omitted. Relative to the full three-axis evaluation, omitting source-code evaluation, scenario-based replay assessment, or adaptive playtesting yields reversed-ordering rates of 20.5\%, 8.9\%, and 11.2\%, respectively. Appendix~\ref{app:complementarity} details the axis-omission analysis and presents additional source-code and runtime discrepancies.

\subsection{Diagnosing Integration Failures}
\label{sec:integration_diagnosis}

We test whether dependency context identifies useful inspection targets beyond an \emph{enumerated} evaluation that scores GDD-derived requirements individually without explicit dependency edges. Using one initial GPT-5.6-Sol build for each of the 100 GDDs, we hold the builds, enumerated source judgments, and requirement mappings fixed. A rule with full enumerated credit is selected for further inspection when a direct predecessor receives less than full credit. This identifies 408 of 908 eligible full-credit targets (44.9\%) across 53 games. The share increases from 25.7\% on \emph{Small} to 58.6\% on \emph{Big}, concentrating additional inspection targets in the broader \emph{Big} designs.

The recorded playtests provide a separate check of these inspection targets. Among 2,185 targets with matched source-code scores and conclusive playtest verdicts, 560 are violated. The enumerated inspection set, consisting of targets with source-code scores below 1.0, includes 398 of these violations. Adding dependency context increases this count to 449, improving violation coverage from 71.1\% to 80.2\% (Table~\ref{tab:source_playtest_diagnostic_coverage}). The 9.1-point gain has a 95\% paired game bootstrap interval of 6.1--13.0 points, and the additional 51 violations span 26 games. Thus, dependencies direct inspection to runtime failures even when the target itself received full enumerated source credit. Appendix~\ref{app:dependency_source_playtest} gives the matching procedure, denominators, and source-level comparisons.

Requirement-linked evidence also explains what must change in the implementation. In \texttt{Grand Atelier 2D}, a collection with insufficient grades or inconsistent composition must fail with a reputation penalty of 12. The enumerated source judgments credit the grade and coherence checks, but the submission handler permits only passing collections, leaving the required failure transition and penalty unimplemented. Tracing the stated condition through its required effects reveals the missing branch. This illustrates why diagnosis must follow conditions through the transitions and state updates required by the design. Additional code-level and runtime examples appear in Appendix~\ref{app:dependency_source_playtest} and Appendix~\ref{app:complementarity}.

\begin{figure*}[!t]
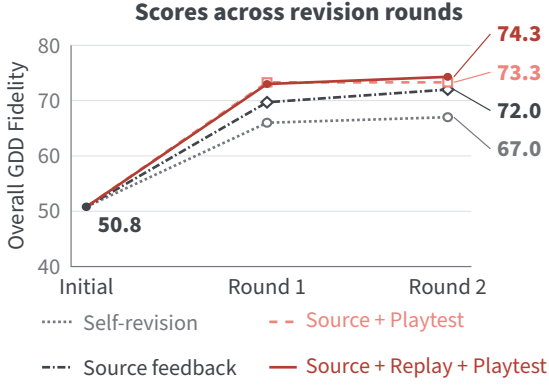

\centering
\begin{minipage}[b]{.475\textwidth}
\vspace{0pt}
\centering
\begingroup
\input{figures/revision_cases/case_helpers}
\colorlet{IRPlay}{A2ZRed!70!white}
\resizebox{\linewidth}{!}{%
\begin{tikzpicture}[x=1cm,y=-.93cm,text=RCInk,font=\AtoZFigFont,every node/.style={inner sep=0pt,outer sep=0pt},
  ir text/.style={font=\AtoZFigFont\fontsize{9.2}{11.2}\selectfont},
  ir legend/.style={font=\AtoZFigFont\fontsize{9}{11}\selectfont},
  ir label/.style={font=\AtoZFigFont\fontsize{9.4}{11.4}\selectfont\bfseries}]
\path[use as bounding box] (0,0) rectangle (7.50,5.50);
\node[font=\AtoZFigFont\fontsize{10.2}{12.2}\selectfont\bfseries] at (3.79,.20) {Scores across revision rounds};
\foreach \score in {40,50,60,70,80}{%
  \pgfmathsetmacro{\IRY}{3.80-(\score-40)*.078}
  \draw[RCBorder,line width=.35pt] (.80,\IRY)--(6.05,\IRY);
  \node[anchor=east,ir text,text=RCMuted] at (.66,\IRY) {\score};
}
\draw[RCMuted,line width=.45pt] (.80,.68)--(.80,3.80)--(6.05,3.80);
\node[ir text,rotate=90] at (.10,2.24) {Overall \GDDmetric{}};
\foreach \x/\label in {1.00/Initial,3.37/Round 1,5.74/Round 2}{%
  \node[ir text] at (\x,4.08) {\label};
}
\draw[RCMuted,line width=1pt,densely dotted] (1.00,2.9576)--(3.37,1.7720)--(5.74,1.6940);
\foreach \x/\y in {3.37/1.7720,5.74/1.6940}{\draw[fill=white,draw=RCMuted,line width=.8pt] (\x,\y) circle (.056);}
\draw[RCInk,line width=1pt,dash pattern=on 3pt off 1pt on .7pt off 1pt] (1.00,2.9576)--(3.37,1.4834)--(5.74,1.3040);
\foreach \x/\y in {3.37/1.4834,5.74/1.3040}{\draw[fill=white,draw=RCInk,line width=.8pt] (\x,{\y-.075})--({\x+.075},\y)--(\x,{\y+.075})--({\x-.075},\y)--cycle;}
\draw[IRPlay,line width=1pt,dashed] (1.00,2.9576)--(3.37,1.2026)--(5.74,1.2026);
\foreach \x/\y in {3.37/1.2026,5.74/1.2026}{\draw[fill=white,draw=IRPlay,line width=.8pt] ({\x-.058},{\y-.058}) rectangle ({\x+.058},{\y+.058});}
\draw[RCAccent,line width=1pt] (1.00,2.9576)--(3.37,1.2260)--(5.74,1.1246);
\foreach \x/\y in {3.37/1.2260,5.74/1.1246}{\fill[RCAccent] (\x,\y) circle (.056);}
\fill[RCInk] (1.00,2.9576) circle (.06);
\node[anchor=west,ir label] at (1.15,3.21) {50.8};
\foreach \y/\labely/\score/\col in {1.6940/2.14/67.0/RCMuted,1.3040/1.60/72.0/RCInk,1.2026/1.06/73.3/IRPlay,1.1246/.52/74.3/RCAccent}{%
  \draw[\col,line width=.45pt] (5.82,\y)--(6.30,\labely);
  \node[anchor=west,ir label,text=\col] at (6.39,\labely) {\score};
}
\draw[RCMuted,line width=1pt,densely dotted] (.42,4.58)--(.82,4.58);
\node[anchor=west,ir legend,text=RCMuted] at (.96,4.58) {Self-revision};
\draw[RCInk,line width=1pt,dash pattern=on 3pt off 1pt on .7pt off 1pt] (.42,5.22)--(.82,5.22);
\node[anchor=west,ir legend] at (.96,5.22) {Source feedback};
\draw[IRPlay,line width=1pt,dashed] (3.40,4.58)--(3.76,4.58);
\node[anchor=west,ir legend,text=IRPlay] at (3.88,4.58) {Source + Playtest};
\draw[RCAccent,line width=1pt] (3.40,5.22)--(3.76,5.22);
\node[anchor=west,ir legend,text=RCAccent] at (3.88,5.22) {Source + Replay + Playtest};
\end{tikzpicture}%
}
\endgroup
\caption{\textbf{Iterative revision.} Mean overall \GDDmetric{} on 50 \emph{Big} GDDs under matched revision budgets. All conditions share the initial builds and revision agent; Figure~\ref{fig:revision_cases} shows qualitative comparisons.}
\label{fig:iterative_revision}\label{fig:revision_trajectory}
\par\vspace{0pt}
\end{minipage}\hfill%
\begin{minipage}[b]{.485\textwidth}
\vspace{0pt}
\input{tables/exp_dependency_diagnosis}
\par\vspace{0pt}
\end{minipage}
\end{figure*}

\subsection{Iterative Revision with Three-Axis Feedback}
\label{sec:iterative_revision}

We study whether requirement-level feedback based on our contract improves overall \GDDmetric{} beyond revision guided by direct inspection of the GDD and source code, using 50 \emph{Big} GDDs with GPT-5.6-Sol. All conditions start from the same initial builds, use the same revision agent for two rounds, and are evaluated on all three axes against the same contracts. The \emph{self-revision} baseline gets feedback from an evaluator agent that directly reviews the GDD and the source code without using our contract or contract-based feedback. We compare it with three benchmark-feedback conditions: \emph{Source feedback} provides contract-based source-code evaluation, \emph{Source + Playtest} adds playtest feedback, and \emph{Source + Replay + Playtest} corresponds to the three-axis feedback. As shown in Figure~\ref{fig:iterative_revision}, \emph{Source + Replay + Playtest} reaches a mean overall \GDDmetric{} of 74.3 after two rounds, exceeding the 67.0 of \emph{self-revision} by 7.3 points (10.9\% relative). After two revision rounds, the larger gains from our feedback support using requirement-level judgment evidence to guide revisions toward the specified design, rather than relying solely on the GDD and source code.

\begin{figure*}[!p]
\centering
\begingroup
\input{figures/revision_cases/case_helpers}
\begin{subcaptiongroup}
\resizebox{\linewidth}{!}{%
\begin{tikzpicture}[x=1cm,y=-1cm,text=RCInk,font=\AtoZFigFont,every node/.style={inner sep=0pt,outer sep=0pt},
  rc title/.style={anchor=west,font=\AtoZFigFont\fontsize{11.2}{13.2}\selectfont\bfseries},
  rc heading/.style={font=\AtoZFigFont\fontsize{9.8}{11.8}\selectfont\bfseries},
  rc label/.style={anchor=north west,text=RCMuted,font=\AtoZFigFont\fontsize{9.6}{11.6}\selectfont\bfseries},
  rc body/.style={anchor=north west,text width=3.22cm,align=left,font=\AtoZFigFont\fontsize{9.8}{12.2}\selectfont},
  rc result/.style={font=\AtoZFigFont\fontsize{10}{12}\selectfont\bfseries}]
\path[use as bounding box] (0,0) rectangle (15.44,19.31);
\newcommand{\RCNativeCrop}[9]{%
  \begin{scope}[shift={(#1,#2)}]
    \pgfmathsetmacro{\RCScale}{#3/#7}
    \pgfmathsetmacro{\RCImageWidth}{#9*\RCScale}
    \clip (0,0) rectangle (#3,#4);
    \node[anchor=north west] at ({-#5*\RCScale},{-#6*\RCScale}) {\includegraphics[width=\RCImageWidth cm]{#8}};
  \end{scope}
  \draw[RCBorder,line width=.4pt] (#1,#2) rectangle ({#1+#3},{#2+#4});
}
\newcommand{\RCRequirement}[4]{%
  \node[rc label] at (.40,1.21) {Condition};
  \node[rc body] at (.40,1.54) {#1};
  \node[rc label] at (.40,#2) {Required effect};
  \node[rc body] at (.40,#3) {#4};
}
\newcommand{\RCInitialPanel}[4]{%
  \node[rc title] at (.20,.23) {\phantomsubcaption\label{#1}(\thesubfigure) #2};
  \draw[RCBorder,line width=.5pt] (.20,.47)--(15.18,.47);
  \node[rc heading,anchor=west] at (.40,.72) {GDD requirement};
  \node[rc heading] at (5.75,.72) {Initial build};
  \node[rc heading] at (9.59,.72) {Self-revision (baseline)};
  \node[rc heading,text=RCAccent] at (13.43,.72) {Source + Replay + Playtest};
  \path[fill=RCFill!75!white,draw=RCBorder,line width=.4pt] (.20,.97) rectangle (3.71,#3);
  \node[anchor=south west,text=RCMuted,font=\AtoZFigFont\fontsize{9.6}{11.6}\selectfont] at (.40,{#3-.15}) {#4};
}
\newcommand{\RCInitialView}[7]{%
  \RCNativeCrop{#1}{.97}{3.50}{1.96875}{0}{0}{#3}{#2}{#3}
  \pgfmathsetmacro{\RCFullScale}{3.50/#3}
  \draw[white,line width=1.6pt] ({#1+#4*\RCFullScale},{.97+#5*\RCFullScale}) rectangle ({#1+(#4+280)*\RCFullScale},{.97+(#5+88)*\RCFullScale});
  \draw[RCAccent,line width=.6pt] ({#1+#4*\RCFullScale},{.97+#5*\RCFullScale}) rectangle ({#1+(#4+280)*\RCFullScale},{.97+(#5+88)*\RCFullScale});
  \RCNativeCrop{#1}{3.10}{3.50}{1.10}{#4}{#5}{280}{#2}{#3}
  \draw[RCAccent,line width=.6pt] (#1,3.10) rectangle ({#1+3.50},4.20);
  \path[fill=#7!6!white,draw=RCBorder,line width=.4pt] (#1,4.20) rectangle ({#1+3.50},4.62);
  \node[font=\AtoZFigFont\fontsize{9.8}{11.8}\selectfont\bfseries,text=#7] at ({#1+1.75},4.41) {#6};
}
\newcommand{\RCDragView}[4]{%
  \RCNativeCrop{#1}{.97}{3.50}{1.96875}{0}{0}{1920}{#2}{1920}
  \draw[white,line width=1.6pt] ({#1+370*3.5/1920},{.97+535*3.5/1920}) rectangle ({#1+1130*3.5/1920},{.97+(535+760*1.1/3.5)*3.5/1920});
  \draw[RCAccent,line width=.6pt] ({#1+370*3.5/1920},{.97+535*3.5/1920}) rectangle ({#1+1130*3.5/1920},{.97+(535+760*1.1/3.5)*3.5/1920});
  \RCNativeCrop{#1}{3.10}{3.50}{1.10}{370}{535}{760}{#2}{1920}
  \draw[RCAccent,line width=.6pt] (#1,3.10) rectangle ({#1+3.50},4.20);
  \pgfmathsetmacro{\RCPointerX}{#1+480*3.5/760}
  \pgfmathsetmacro{\RCPointerY}{3.1+75*3.5/760}
  \RCPointer{\RCPointerX}{\RCPointerY}
  \node[text=white,font=\AtoZFigFont\fontsize{9.1}{11}\selectfont\bfseries] at ({#1+2.66},3.76) {Pointer};
  \draw[white,line width=.55pt] ({#1+2.46},3.65)--({#1+2.25},3.51);
  \path[fill=#4!6!white,draw=RCBorder,line width=.4pt] (#1,4.20) rectangle ({#1+3.50},4.62);
  \node[rc result,text=#4] at ({#1+1.75},4.41) {#3};
}
\newcommand{\RCOxygenView}[5]{%
  \RCNativeCrop{#1}{.97}{3.50}{1.18}{520}{0}{880}{#2}{1920}
  \node[text=RCMuted,font=\AtoZFigFont\fontsize{9.5}{11.5}\selectfont\bfseries] at ({#1+1.75},2.45) {Recorded runtime state};
  \path[fill=RCFill!75!white,draw=RCBorder,line width=.4pt] (#1,2.70) rectangle ({#1+3.50},4.20);
  \node[anchor=west,font=\AtoZFigFont\fontsize{11.4}{13.4}\selectfont] at ({#1+.22},3.12) {Oxygen};
  \node[anchor=east,font=\AtoZFigFont\fontsize{12.2}{14.2}\selectfont\bfseries] at ({#1+3.28},3.12) {0};
  \draw[RCBorder,line width=.4pt] ({#1+.22},3.43)--({#1+3.28},3.43);
  \node[anchor=west,font=\AtoZFigFont\fontsize{11.4}{13.4}\selectfont] at ({#1+.22},3.81) {Hull};
  \node[anchor=east,text=#5,font=\AtoZFigFont\fontsize{13.2}{15.2}\selectfont\bfseries] at ({#1+3.28},3.81) {#3};
  \path[fill=#5!6!white,draw=RCBorder,line width=.4pt] (#1,4.20) rectangle ({#1+3.50},4.62);
  \node[rc result,text=#5] at ({#1+1.75},4.41) {#4};
}
\begin{scope}
\RCInitialPanel{fig:revision_case_chameleon}{\texttt{Chameleon Slide}}{4.62}{RT-COLOR K1}
\RCRequirement{Key 1--4 pressed\\while at rest}{2.47}{2.80}{Set the body color to\\the selected color\\instantly.}
\RCInitialView{4.00}{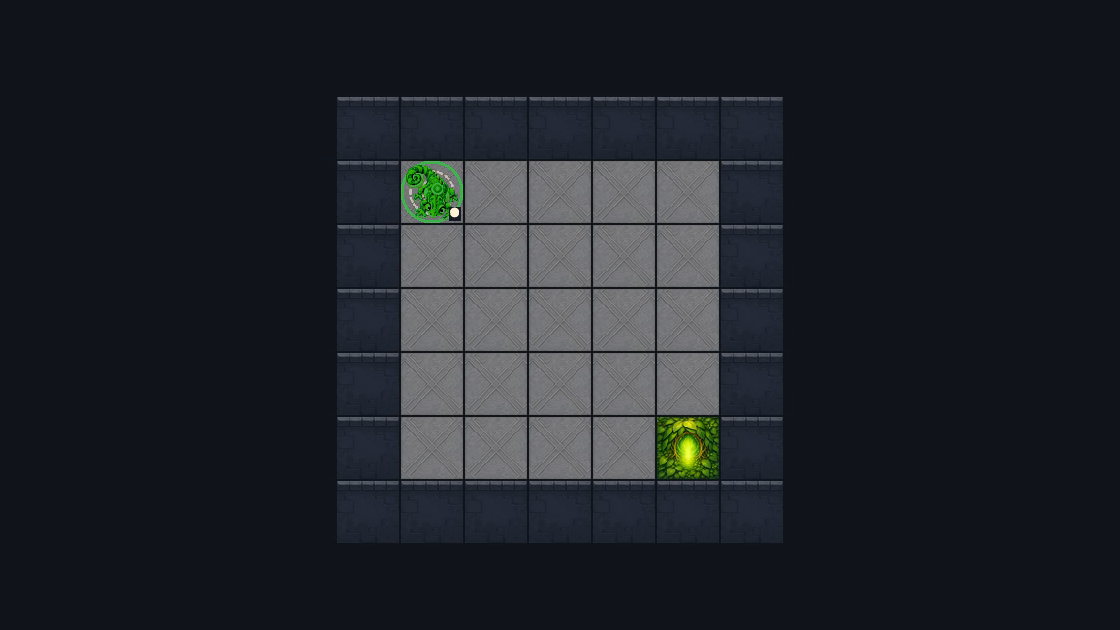}{1120}{330}{150}{Body remains green}{RCInk}
\RCInitialView{7.84}{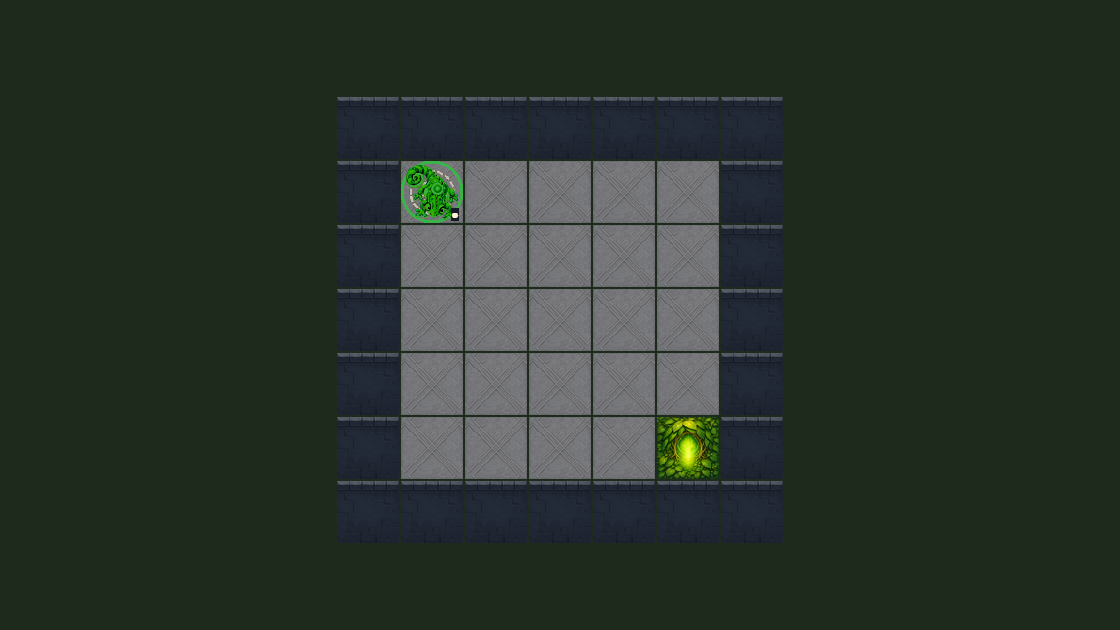}{1120}{330}{150}{Body remains green}{RCInk}
\RCInitialView{11.68}{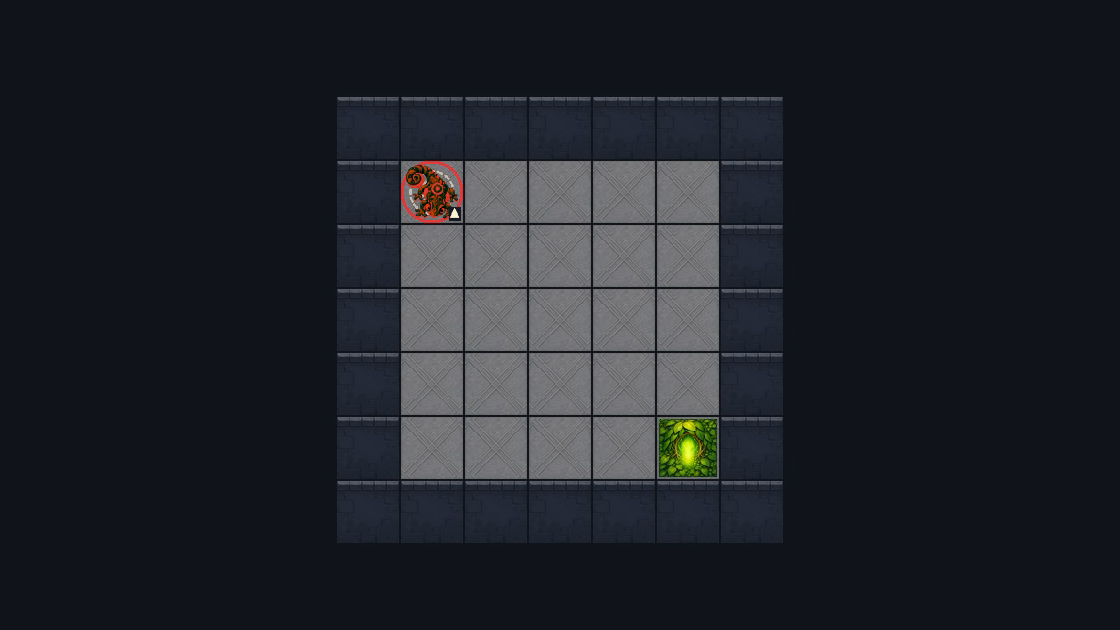}{1120}{330}{150}{Body recolors}{RCAccent}
\end{scope}
\begin{scope}[shift={(0,4.87)}]
\RCInitialPanel{fig:revision_case_fogfall}{\texttt{Fogfall Delivery}}{4.62}{RT-ESTIMATE R3}
\RCRequirement{$R < d \leq R+4$,\\no lighthouse coverage}{2.47}{2.80}{Show estimate class\\icons with confidence;\\badge color by band.}
\RCInitialView{4.00}{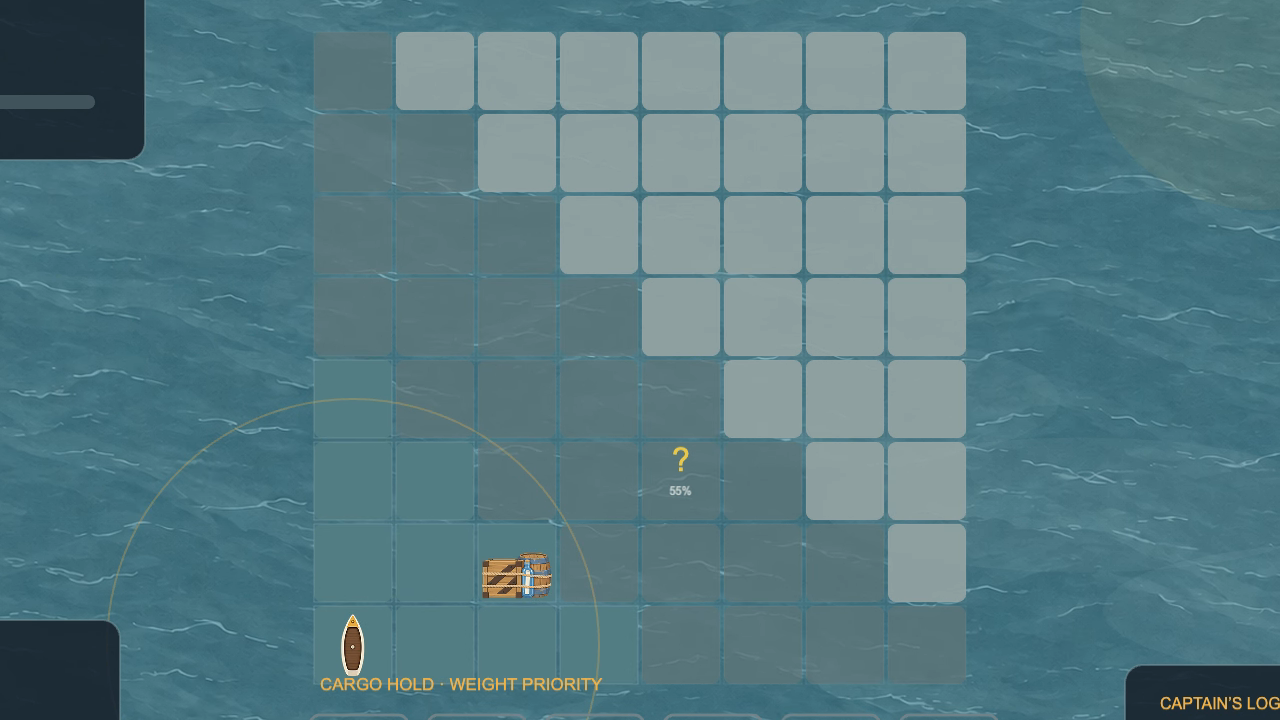}{1280}{395}{190}{Blank estimate cells}{RCInk}
\RCInitialView{7.84}{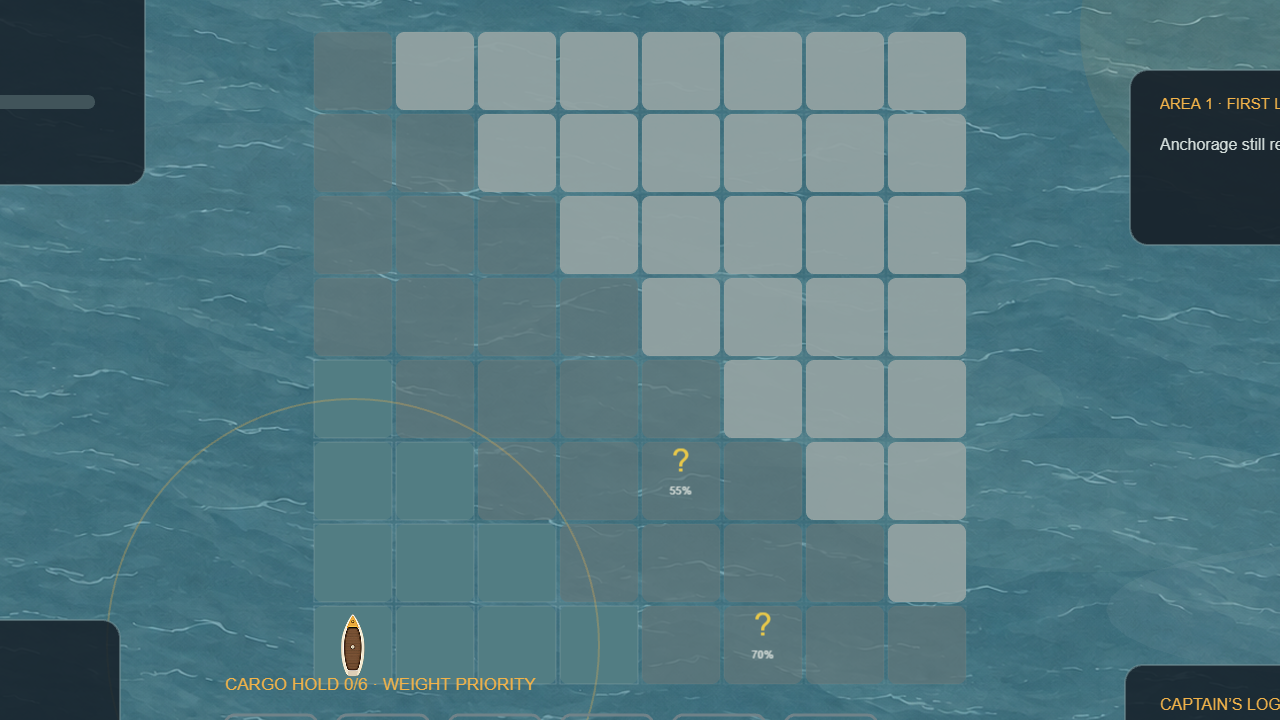}{1280}{395}{190}{Blank estimate cells}{RCInk}
\RCInitialView{11.68}{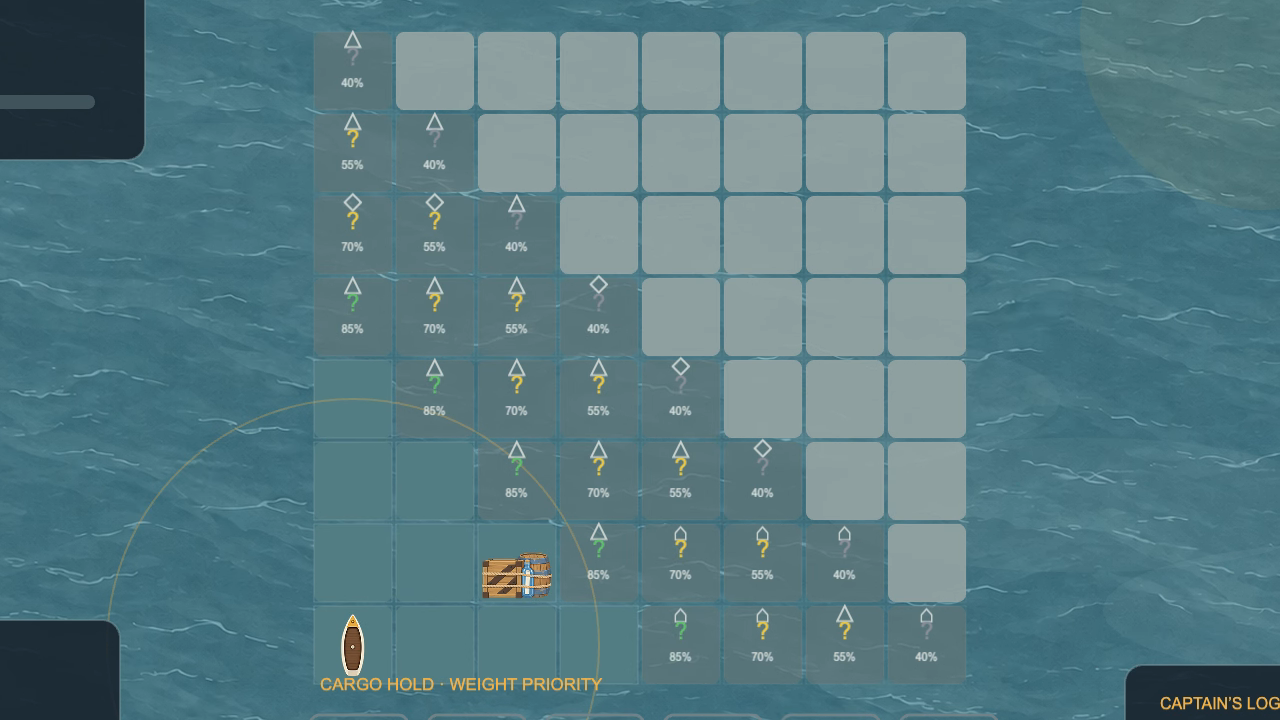}{1280}{395}{190}{Icons and confidence}{RCAccent}
\end{scope}
\begin{scope}[shift={(0,9.74)}]
\RCInitialPanel{fig:revision_case_beat}{\texttt{Beat Reroute}}{4.62}{SUR-V1}
\RCRequirement{Pointer moves while\\a fragment is held.}{2.47}{2.80}{Fragment follows\\the pointer with a\\20\,px vertical offset.}
\RCDragView{4.00}{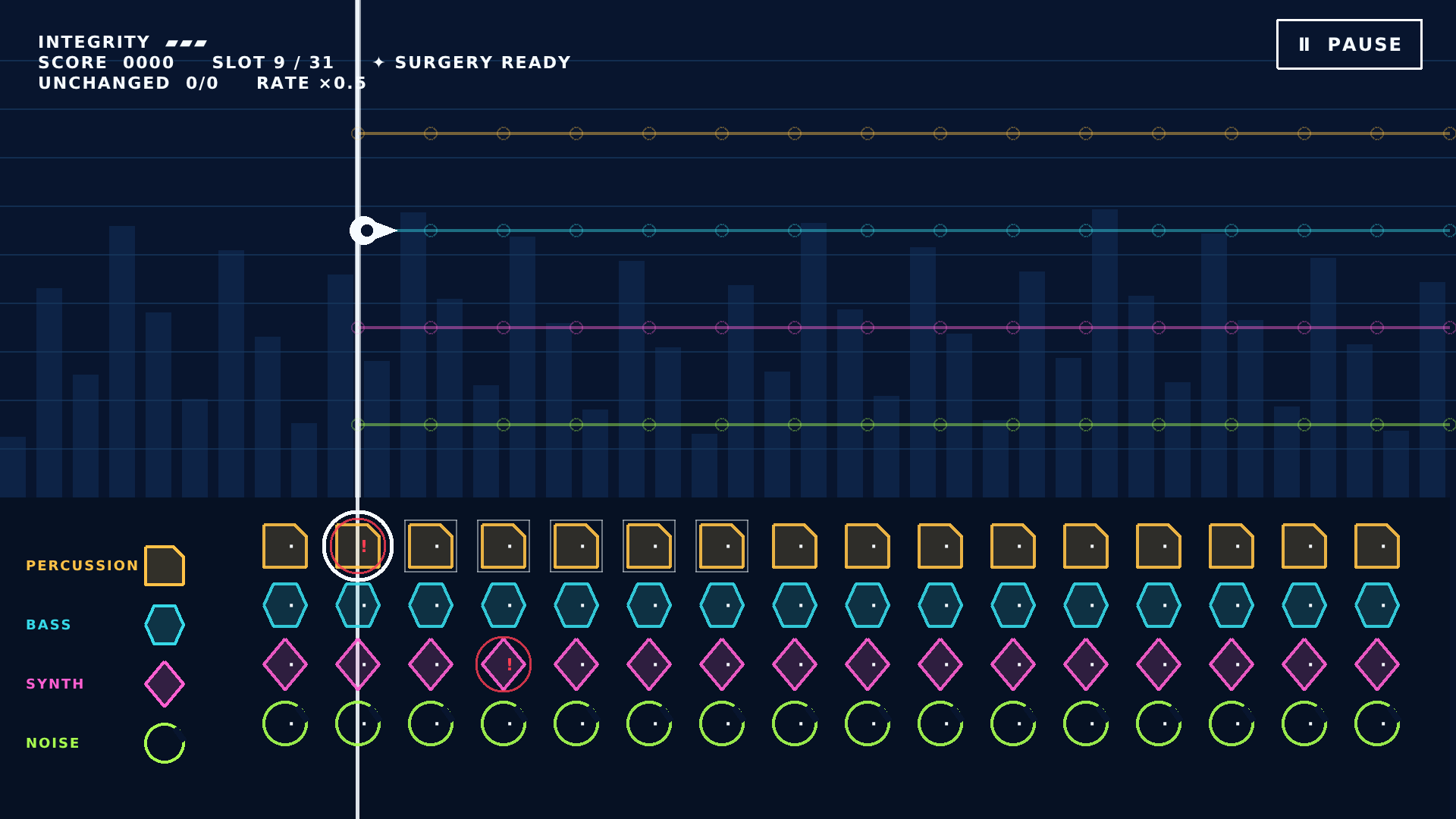}{Stays in slot}{RCInk}
\RCDragView{7.84}{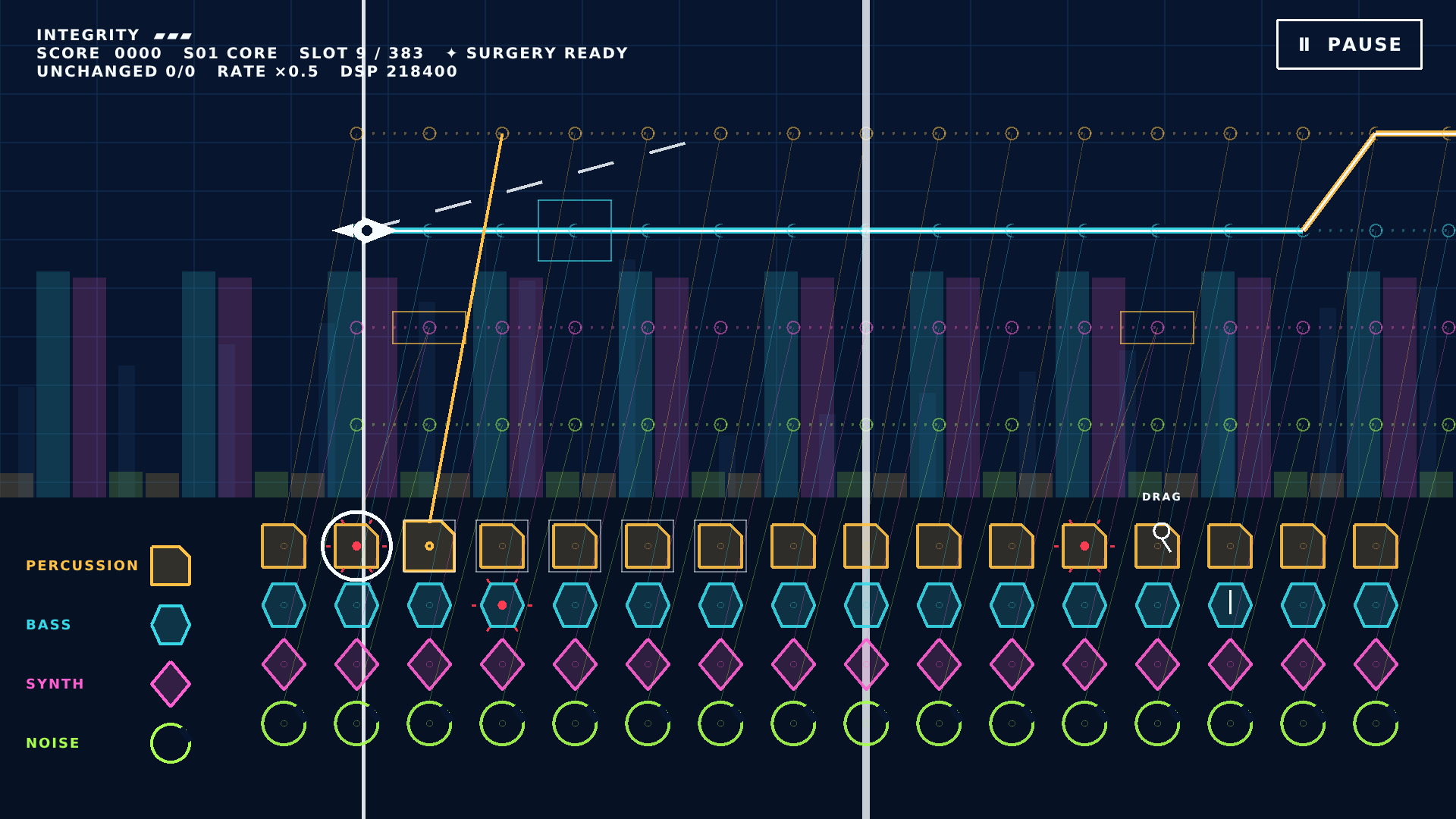}{Stays in slot}{RCInk}
\RCDragView{11.68}{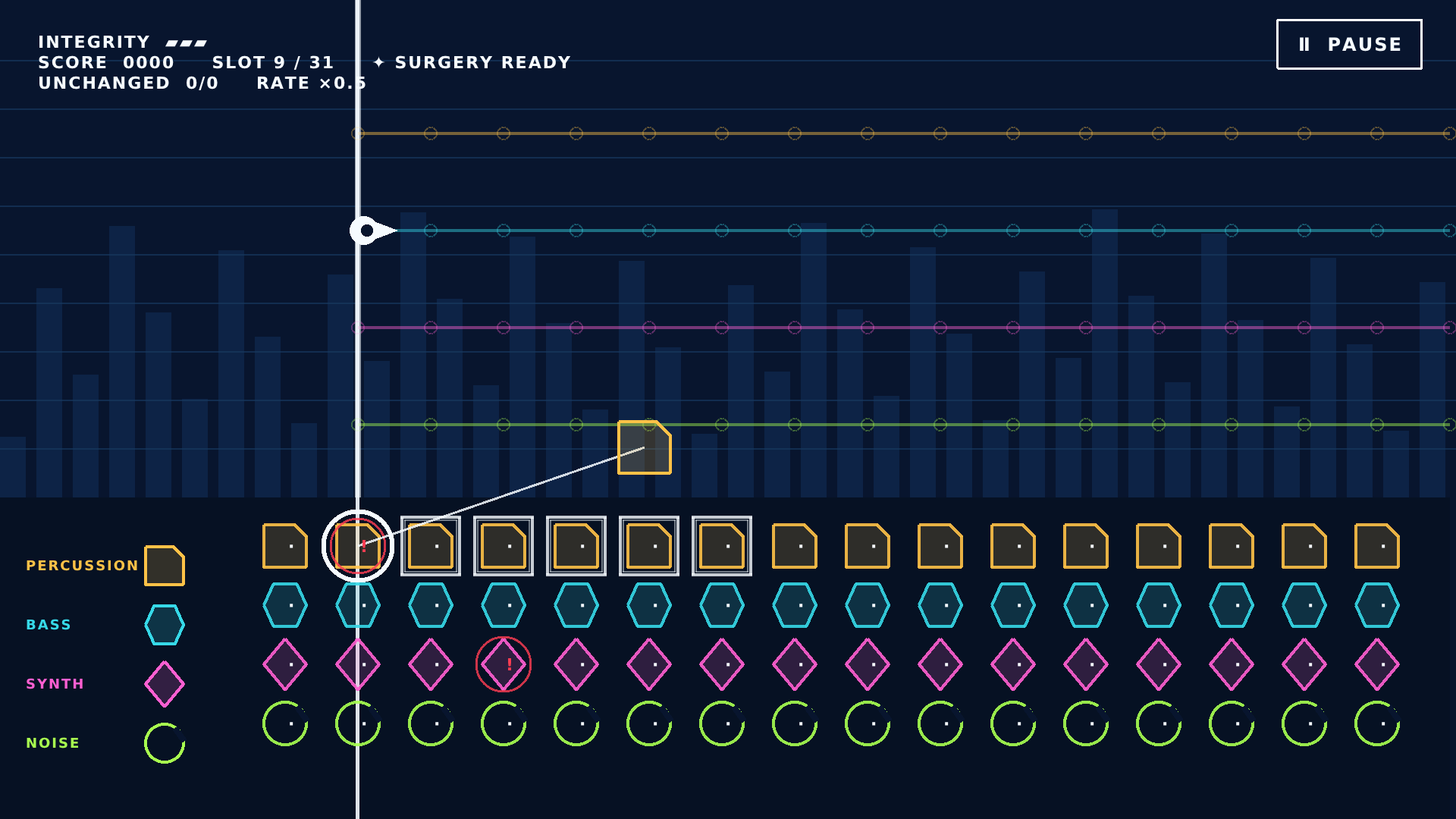}{Follows pointer}{RCAccent}
\end{scope}
\begin{scope}[shift={(0,14.61)}]
\RCInitialPanel{fig:revision_case_abyssal}{\texttt{Abyssal Chain}}{4.62}{RT-DAMAGE D5}
\RCRequirement{Oxygen reaches 0.}{2.47}{2.80}{Set Hull to 0\\immediately.}
\RCOxygenView{4.00}{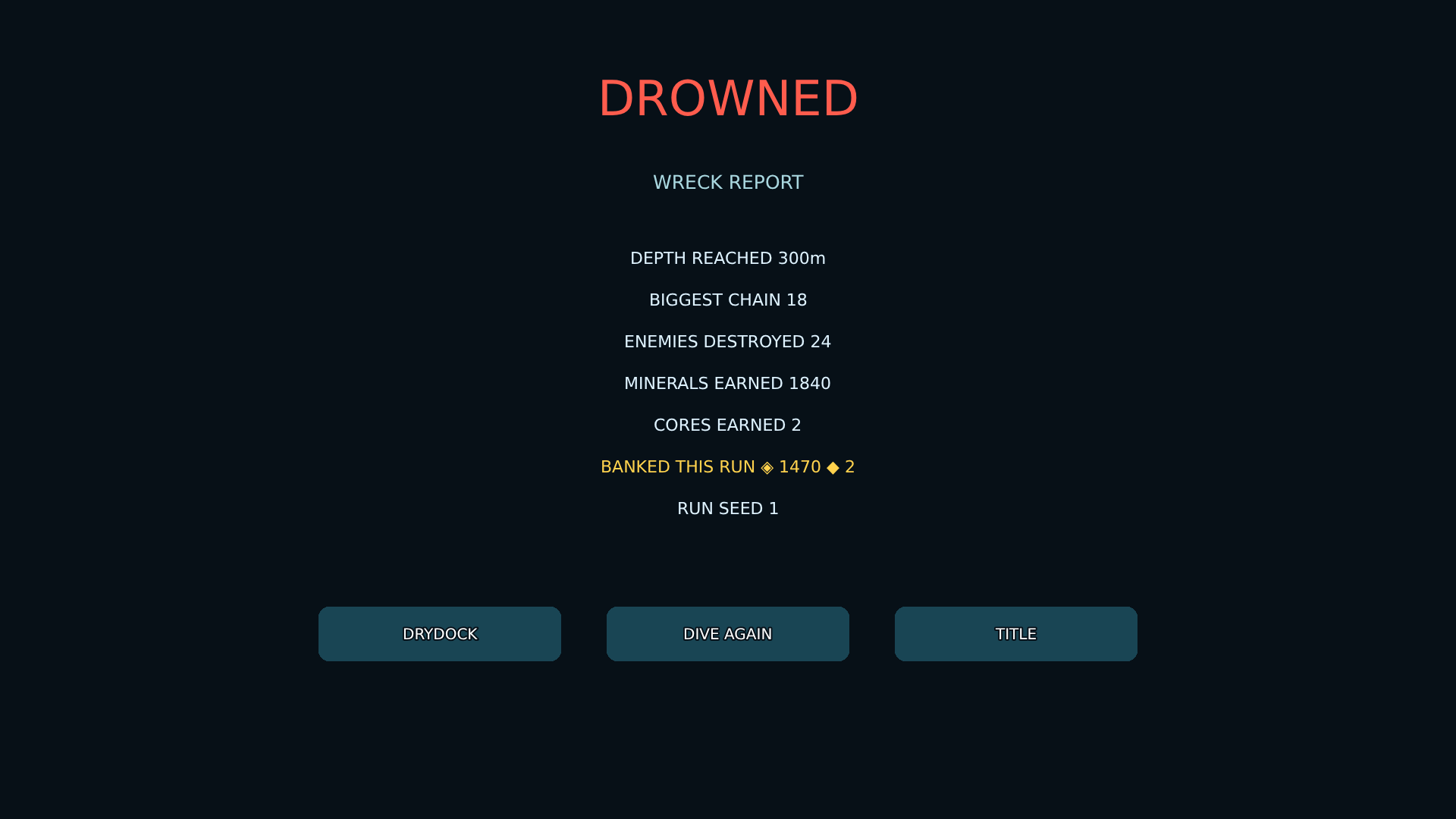}{100}{Hull update missing}{RCInk}
\RCOxygenView{7.84}{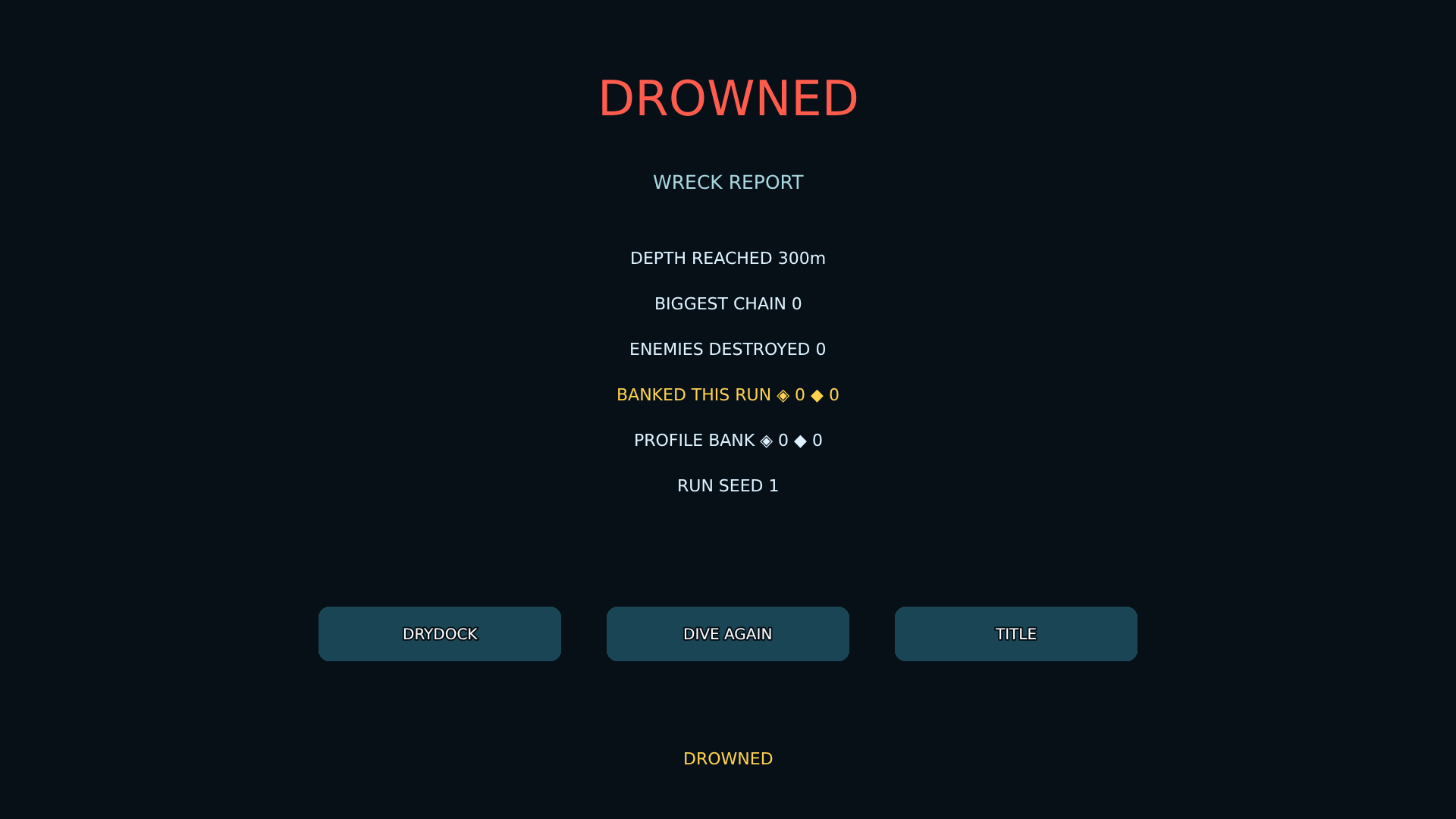}{100}{Hull update missing}{RCInk}
\RCOxygenView{11.68}{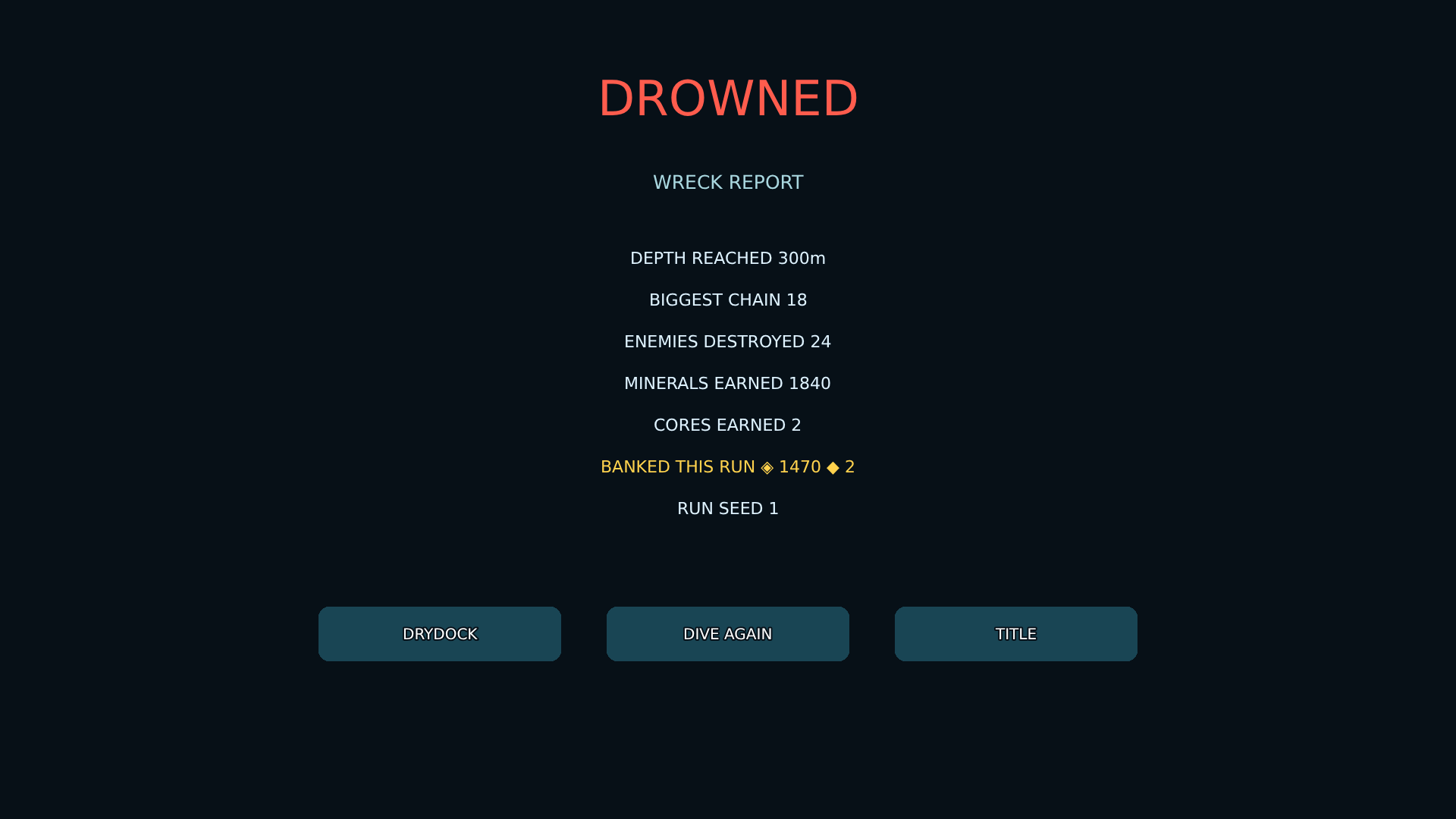}{0}{Hull update applied}{RCAccent}
\end{scope}
\end{tikzpicture}%
}
\end{subcaptiongroup}

\caption{\textbf{Requirement-level repairs beyond self-revision.} Each panel pairs a GDD requirement with the initial build, the \emph{self-revision} baseline, and \emph{Source + Replay + Playtest}; both revision columns show second-round builds. The examples compare \subref{fig:revision_case_chameleon} body recoloring, \subref{fig:revision_case_fogfall} estimate-class icons and confidence, \subref{fig:revision_case_beat} a held fragment following the pointer, and \subref{fig:revision_case_abyssal} the Hull update required by oxygen depletion. In each illustrated case, the violation remains in the initial and \emph{self-revision} builds and is repaired by \emph{Source + Replay + Playtest}. Detail crops enlarge the highlighted regions in \subref{fig:revision_case_chameleon}--\subref{fig:revision_case_beat}; crosshairs in \subref{fig:revision_case_beat} mark the actual pointer position. In \subref{fig:revision_case_abyssal}, all three builds display a drowned-run screen, but only the feedback revision sets Hull to 0. The state cards report values recorded during execution. In \subref{fig:revision_case_fogfall}, $R$ is the reveal radius and $d$ is Manhattan distance from the ship.}
\label{fig:revision_cases}\label{fig:revision_qualitative}
\endgroup
\end{figure*}

Figure~\ref{fig:revision_cases} compares initial builds with the \emph{self-revision} baseline and \emph{Source + Replay + Playtest} after two revision rounds in four games, pairing each GDD requirement with the resulting behavior. In \texttt{Chameleon Slide}, pressing a color key at rest must immediately recolor the character's body. The initial and \emph{self-revision} builds retain a green body, whereas \emph{Source + Replay + Playtest} recolors it. In \texttt{Fogfall Delivery}, \emph{Source + Replay + Playtest} displays estimate-class icons and confidence in the highlighted cells left blank by the initial and \emph{self-revision} builds.

The remaining cases show how requirement-specific feedback repairs interaction and state-update errors that persist under \emph{self-revision}. In \texttt{Beat Reroute}, a captured fragment must follow the pointer with a 20\,px vertical offset. The initial and \emph{self-revision} builds leave it in its slot, while \emph{Source + Replay + Playtest} renders it at the required offset. In \texttt{Abyssal Chain}, oxygen depletion must immediately set Hull to 0. All three builds show a drowned-run screen, but Hull remains 100 in the initial and \emph{self-revision} builds; only \emph{Source + Replay + Playtest} performs the specified state update. These cases illustrate why revision feedback must address the behavior and state changes required by the GDD. Appendix~\ref{app:gdd_revision_cases} provides the matched probe details and an additional pointer-release comparison.

\subsection{Evaluation Reliability}
\label{sec:evaluation_reliability}

We examine the consistency of evaluation targets and agreement across evaluator configurations. In the requirement-extraction comparison on 100 GDDs, contract construction achieves 93\% citation overlap across five runs on \emph{Big}, compared with 57\% for a naive judge that identifies requirements while evaluating a build. In a separate experiment, three rule generations per GDD with fixed vocabulary and source-bound entries preserve all entries and source associations across 300 outputs; reachability agreement, dependency-edge Jaccard similarity, and role preservation are each at least 0.999. These results support using a common, frozen contract to keep evaluation targets aligned across builds. Appendix~\ref{app:requirement-representation} and Appendix~\ref{app:contract_stability} describe the two experiments and within-rule variation.

We also compare evaluator configurations on fixed inputs. Against GPT-5.6-Sol reference judgments, the default GPT-5.6-Luna high configuration reaches a rule-level Pearson correlation of 0.72 for source-code evaluation and an item-level correlation of 0.85 for scenario-based replay assessment, compared with 0.74 and approximately 0.86 for independent reference-model re-runs. The reference judgments for source-code evaluation and scenario-based replay assessment use extra-high reasoning effort, while the adaptive-playtest reference uses high effort. On fixed playtest traces, GPT-5.6-Luna's verdict agreement is 0.910 on \emph{Small} and 0.888 on \emph{Big}; only 6 of 222 and 3 of 242 rules, respectively, change whether they contribute to the playtest score. These comparisons assess agreement under shared contracts and evidence. Appendix~\ref{app:cost_evaluation} and Appendix~\ref{app:ablation-mllm} provide the configurations, run-to-run references, and costs.

The aggregate ranking is also robust to the relative weights of the evidence channels. Holding the measured axis scores and source dependency parameter $\alpha=0.5$ fixed, Claude-Fable-5.1, Claude-Opus-5, and GPT-6-Astra retain the first three positions under every nonnegative axis weighting that sums to one, separately on \emph{Small}, \emph{Big}, and All. For All, 30 of 36 agent-pair orderings are invariant over this full domain. For the All aggregate, when every axis retains at least half its default contribution, 34 of 36 orderings are invariant and each agent remains within one position of its equal-weight rank. Appendix~\ref{app:axis_weight_sensitivity} gives the continuous-domain analysis and representative reweightings.

\subsection{Component Analysis}
\label{sec:component_analysis}

\begin{table}[t]
    \centering
    \renewcommand{\thesubtable}{(\alph{subtable})}
    \captionsetup{subrefformat=simple}
    \captionsetup[subtable]{font={small,bf},labelformat=simple,labelsep=space,justification=centering,singlelinecheck=false,position=top,skip=2pt,hypcap=true}

    \caption[Component analysis]{\textbf{Component analysis.} \subref{tab:dependency_weighting} Mean and maximum absolute changes in source-code score from uniform weighting (Appendix~\ref{app:ablation-dependency-weight}). \subref{tab:adaptive_frame_selection} Evidence support for fixed-rate versus adaptive frame selection (Appendix~\ref{app:frame_selection}). \subref{tab:adaptive_playtesting} Judgment coverage (\%) after a second normal or adversarial pass under matched budgets. The first normal pass reaches 40.5\%; gains are percentage points relative to this shared first pass (Appendix~\ref{app:adversarial_playtesting}). Coverage includes both \emph{satisfied} and \emph{violated} requirements.}
    \label{tab:component-analysis}
    \vspace{-2pt}

    \scriptsize
    \setlength{\tabcolsep}{1.8pt}
    \renewcommand{\arraystretch}{1.08}

    \begin{subtable}[t]{0.28\linewidth}
        \vspace{0pt}
        \centering
        \caption{Dependency Weighting}
        \label{tab:dependency_weighting}
        \begin{tabularx}{\linewidth}{@{}>{\raggedright\arraybackslash}p{0.41\linewidth}>{\centering\arraybackslash}X>{\centering\arraybackslash}X@{}}
            \toprule
            \multirow[c]{2}{*}[-7.5pt]{\textbf{Setting} ($\alpha$)}
            & \multicolumn{2}{c}{\textbf{Score Change}} \\
            \cmidrule(lr){2-3}
            & \shortstack{\strut\textbf{Mean}\\[-3pt]\strut$|\Delta|$}
            & \shortstack{\strut\textbf{Max.}\\[-3pt]\strut$|\Delta|$} \\[-2pt]
            \midrule
            \strut \textbf{Default} ($0.5$) & 0.38 & 3.71 \\
            \strut Stronger ($0.1$) & 0.96 & 19.10 \\
            \bottomrule
        \end{tabularx}
    \end{subtable}\hfill%
    \begin{subtable}[t]{0.33\linewidth}
        \vspace{0pt}
        \centering
        \caption{Frame Selection}
        \label{tab:adaptive_frame_selection}
        \begin{tabularx}{\linewidth}{@{}>{\raggedright\arraybackslash}p{0.44\linewidth}>{\centering\arraybackslash}X>{\centering\arraybackslash}X@{}}
            \toprule
            \multirow[c]{2}{*}[-7.5pt]{\textbf{Method}}
            & \multicolumn{2}{c}{\textbf{Evidence Support}} \\
            \cmidrule(lr){2-3}
            & \shortstack{\strut\textbf{Item}\\[-3pt]\strut\textbf{Preference}}
            & \shortstack{\strut\textbf{Game}\\[-3pt]\strut\textbf{Wins}} \\[-2pt]
            \midrule
            \strut Fixed-Rate & 39.5 & 13 \\
            \strut Adaptive Selection & \textbf{60.5} & \textbf{35} \\
            \bottomrule
        \end{tabularx}
    \end{subtable}\hfill%
    \begin{subtable}[t]{0.35\linewidth}
        \vspace{0pt}
        \centering
        \caption{Adaptive Playtesting}
        \label{tab:adaptive_playtesting}
        \begin{tabularx}{\linewidth}{@{}>{\raggedright\arraybackslash}p{0.34\linewidth}>{\centering\arraybackslash}X>{\centering\arraybackslash}X@{}}
            \toprule
            \multirow[c]{2}{*}[-7.5pt]{\textbf{Second Pass}}
            & \multicolumn{2}{c}{\textbf{After Normal Play}} \\
            \cmidrule(lr){2-3}
            & \shortstack{\strut\textbf{Judgment}\\[-3pt]\strut\textbf{Coverage}}
            & \shortstack{\strut\textbf{Coverage}\\[-3pt]\strut\textbf{Gain}} \\[-2pt]
            \midrule
            \strut Normal & 41.0 & $+0.5$ \\
            \strut Adversarial & \textbf{48.5} & $\mathbf{+8.0}$ \\
            \bottomrule
        \end{tabularx}
    \end{subtable}

    \vspace{-5pt}
\end{table}

\paragraph{Dependency-Weighted Scoring.}
We examine how dependency weighting (Section~\ref{sec:source_eval}) changes source-code scores under fixed rule judgments. Relative to uniform weighting, the default $\alpha=0.5$ setting adjusts source-code scores by an average of 0.38 points in absolute terms, with a maximum adjustment of 3.71 points, as shown in Table~\ref{tab:dependency_weighting}. Applying stronger weighting at $\alpha=0.1$ increases these changes to 0.96 and 19.10 points, respectively, while model rankings remain unchanged across both splits. In a complementary source-code and playtest analysis, weighting lowers rule scores in 61.1\% of builds with confirmed high-reach failures, compared with 37.5\% of comparison builds (Appendix~\ref{app:ablation-dependency-weight}). Section~\ref{sec:integration_diagnosis} further examines how dependency context improves failure diagnosis while keeping enumerated source judgments fixed.

\paragraph{Adaptive Frame Selection.} We compare adaptive frame selection with fixed-rate sampling across all 50 \emph{Small} GDD games generated by GPT-5.6-Sol, utilizing identical recorded replays, visual rubrics, and frame counts per replay. A separate agent evaluates how effectively the cited frames support each requirement-level rationale. As shown in Table~\ref{tab:adaptive_frame_selection}, adaptive selection is preferred in 60.5\% of comparisons where either method is favored, and receives more preferences in 35 out of 50 games, compared to 13 for fixed-rate sampling, with 2 ties. These results indicate that selecting frames around relevant gameplay events provides stronger support for replay judgments than fixed-rate sampling. Appendix~\ref{app:frame_selection} presents a detailed analysis of these sampling methods.

\paragraph{Normal And Adversarial Playtests.}
We compare a second normal pass with an adversarial pass on 86 dependency-linked rules from 50 games, holding the builds, contracts, test policies, and budgets fixed. Both conditions retain the first normal pass's conclusive verdicts and re-test only its unverified targets. As shown in Table~\ref{tab:adaptive_playtesting}, the first pass establishes 40.5\% judgment coverage, which increases to 41.0\% with another normal pass and 48.5\% with an adversarial pass. The 7.5-percentage-point advantage under matched second-pass budgets supports prerequisite initialization as a way to extend verification beyond repeated normal play. Across all nine coding-agent configurations in the main benchmark, adversarial tests add both satisfaction and violation evidence. Appendix~\ref{app:adversarial_playtesting} provides the controlled protocol and the full decomposition of the main-table playtest results by agent and split.

\subsection{Extension to 3D Games}
\label{sec:3d_extension}

\begin{figure}[!htbp]
\centering
\begingroup
\providecommand{\AtoZFigFont}{\sffamily}
\colorlet{TDInk}{A2ZFigText}
\colorlet{TDMuted}{A2ZFigText!75!white}
\colorlet{TDBorder}{A2ZFigAxis}
\colorlet{TDFill}{A2ZFigRest}
\colorlet{TDAccent}{A2ZRed!78!black}
\renewcommand{\thesubfigure}{\alph{subfigure}}
\captionsetup{subrefformat=parens}
\begin{subcaptiongroup}
\resizebox{\linewidth}{!}{%
\begin{tikzpicture}[x=1cm,y=-1cm,text=TDInk,font=\AtoZFigFont,every node/.style={inner sep=0pt,outer sep=0pt},
  td title/.style={anchor=west,font=\AtoZFigFont\fontsize{11.2}{13.2}\selectfont\bfseries},
  td heading/.style={anchor=north west,text=TDAccent,font=\AtoZFigFont\fontsize{9.2}{11.2}\selectfont\bfseries},
  td body/.style={anchor=north west,text width=7.06cm,align=left,font=\AtoZFigFont\fontsize{9.4}{11.5}\selectfont},
  td note/.style={anchor=north west,text=TDMuted,font=\AtoZFigFont\fontsize{9.0}{11.0}\selectfont}]
\path[use as bounding box] (0,0) rectangle (15.44,7.18);
\newcommand{\TDPanel}[5]{%
  \node[td title] at (.08,.26) {\phantomsubcaption\label{#1}(\thesubfigure) #2};
  \draw[TDBorder,line width=.5pt] (0,.55)--(7.46,.55);
  \path[fill=TDFill!80!white,draw=TDBorder,line width=.4pt] (0,.76) rectangle (7.46,2.17);
  \node[td heading] at (.20,.91) {GDD requirements (summary)};
  \node[td body] at (.20,1.32) {#3};
  \node[anchor=north west] at (0,2.40) {\includegraphics[width=7.46cm]{assets/appendix/showcase/#4}};
  \draw[TDBorder,line width=.4pt] (0,2.40) rectangle (7.46,6.59625);
  \node[td note] at (.08,6.76) {#5};
}
\TDPanel{fig:3d_rocket_league}{\texttt{Rocket League}}{Drive, jump, and collect boost in a five-minute\\match against an opponent bot.}{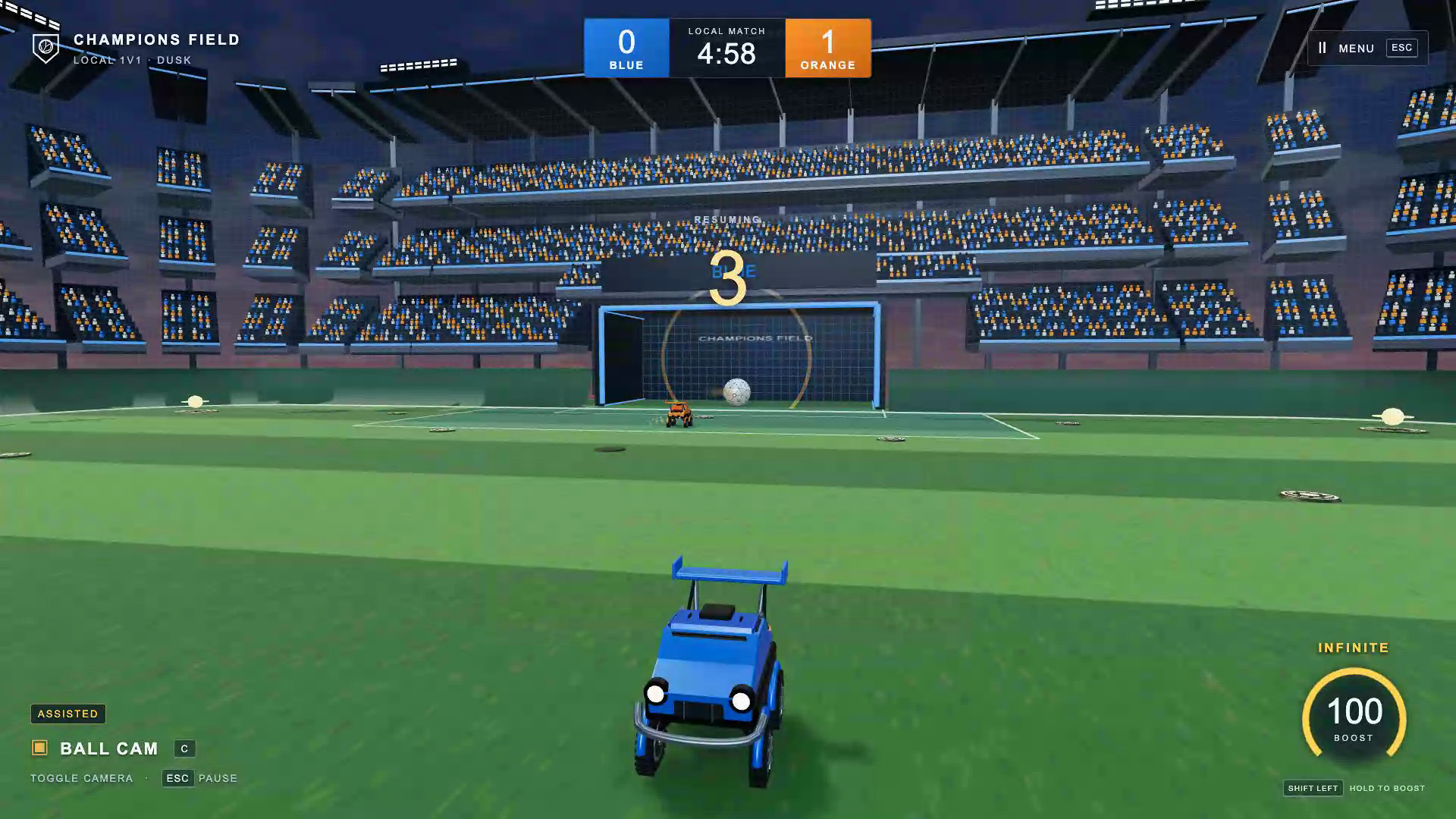}{Match resumption with a countdown.}
\begin{scope}[shift={(7.98,0)}]
\TDPanel{fig:3d_sonic}{\texttt{Sonic: Cascade Coast}}{Carry momentum through slopes, loops, and rails\\along a continuous coastal course.}{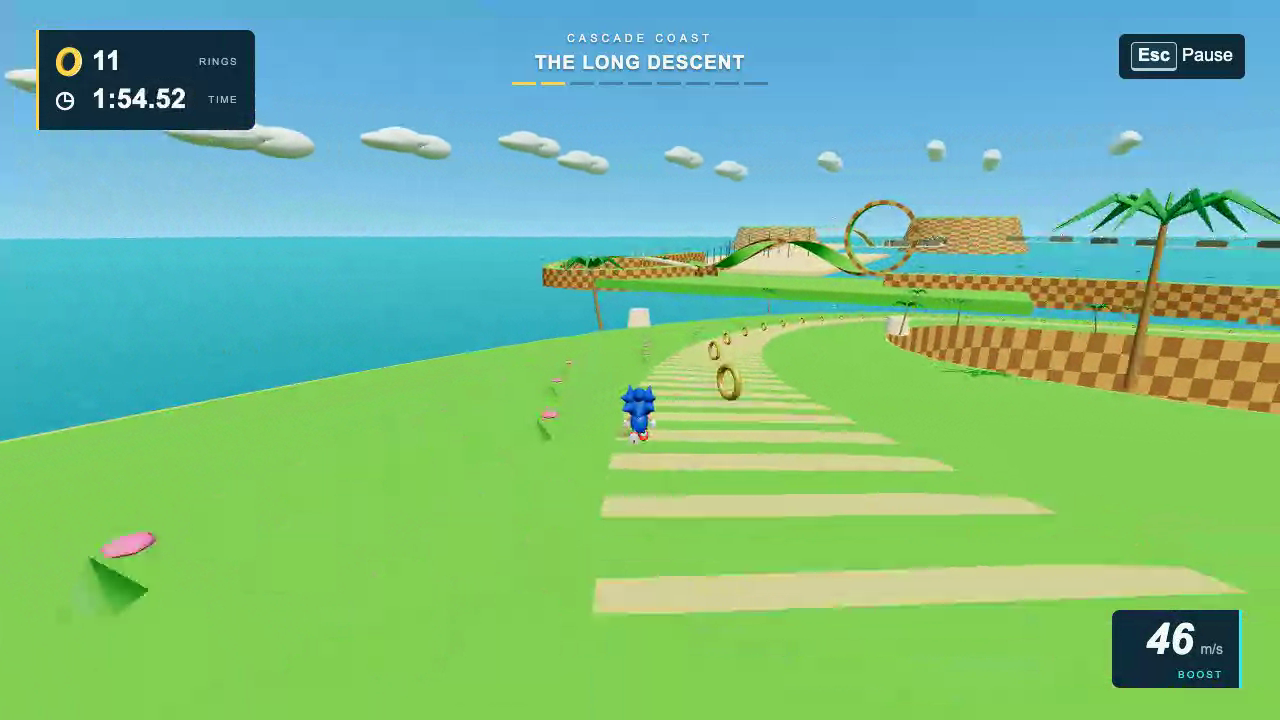}{A later sector: The Long Descent.}
\end{scope}
\end{tikzpicture}%
}
\end{subcaptiongroup}
\caption{\textbf{Extension of contract-based evaluation to 3D games.} GDD requirement summaries accompany original gameplay captures from two Three.js builds: \subref{fig:3d_rocket_league} vehicle-based ball play in \texttt{Rocket League} and \subref{fig:3d_sonic} momentum-based platforming in \texttt{Sonic: Cascade Coast}. Both builds are assessed through source-code inspection, scenario-based replay, and adaptive playtesting against their fixed contracts. Appendix~\ref{app:scalable_benchmark} reports the scores and requirement-level diagnoses.}
\label{fig:3d_extension}
\label{fig:scalibilty-showcase}
\endgroup
\end{figure}

The framework separates GDD-derived contracts from the engine-specific interfaces used to collect evidence, allowing the contract representation and three evaluation axes to be retained across runtimes. Extending the benchmark requires adapting game execution, player input, state observation, scenario initialization, and frame capture, while the brief-to-GDD workflow can specify designs for the target environment. We apply this structure to three additional 3D games built by GPT-6-Astra in Three.js: \texttt{Sonic: Cascade Coast}, \texttt{Diablo Cathedral}, and \texttt{Rocket League}. Each build is evaluated against its fixed GDD-derived contract through source-code inspection, scenario-based replay, and adaptive playtesting. Figure~\ref{fig:3d_extension} pairs GDD requirement summaries with gameplay examples from two of these builds. Appendix~\ref{app:scalable_benchmark} reports the three-axis scores and requirement-level diagnoses.

\newpage

\section{Conclusion}
\label{sec:conclusion}
We introduced \AtoZbench{}, a benchmark for evaluating coding agents' faithfulness to 100 long-form GDDs in end-to-end game development. Fixed~\textsc{Dependency-Aware Contracts} guide both source-code inspection and agent-generated~\textsc{Test Policies} for replay and adaptive playtesting, linking evidence to the same requirements. The results show that runnable outputs may not fully correspond to the intended design and that local source-code passes can coexist with failed prerequisites.

Dependency context expands the coverage of recorded execution failures, and the same evaluation structure produces requirement-level diagnoses in three additional 3D games. By connecting requirement-level evaluation with targeted revision, \AtoZbench{} enables both assessment and improvement of specification faithfulness in game development.

\section*{Acknowledgments}
We sincerely thank Kyungdo Park, Janghoon Ju, Jaeuk Kim, Inkyu Park, Myungseok Oh, Yujin Hong, and Inyoung Cho from the KRAFTON AtoZ team for helpful discussions and support throughout this project. We also thank Kangwook Lee and Junesig Sung from KRAFTON for their support, and Jiho Choi from KAIST for his valuable advice and feedback on this work.

\clearpage
\bibliography{reference}

\clearpage
\appendix
\label{sec:appendix}\clearpage

\setcounter{figure}{0}
\setcounter{table}{0}
\renewcommand{\thefigure}{S\arabic{figure}}
\renewcommand{\thetable}{S\arabic{table}}

\setcounter{topnumber}{3}
\setcounter{bottomnumber}{2}
\setcounter{totalnumber}{5}
\renewcommand{\floatpagefraction}{0.65}
\renewcommand{\dblfloatpagefraction}{0.65}
\makeatletter
\newcommand{\AtoZAppendixInput}[2][!htbp]{%
  \let\AtoZAppendixOriginalXFloat\@xfloat
  \def\@xfloat##1[##2]{\AtoZAppendixOriginalXFloat{##1}[#1]}%
  \input{#2}%
  \let\@xfloat\AtoZAppendixOriginalXFloat
}
\newcommand{\AtoZAppendixBarrier}{%
  \par
  \begingroup
  \let\@elt\relax
  \edef\AtoZAppendixPending{\@botlist\@deferlist\@dbldeferlist}%
  \ifx\AtoZAppendixPending\@empty\else
    \ifx\@fltovf\relax
      \clearpage
    \else
      \newpage
      \let\@fltovf\relax
      \AtoZAppendixBarrier
    \fi
  \fi
  \endgroup
  \suppressfloats[t]%
}
\makeatother

\pgfplotsset{a2zfig/.style={
  width=\linewidth, height=0.5\linewidth,
  axis background/.style={fill=white},
  axis line style={A2ZFigAxis,line width=0.5pt},
  tick style={A2ZFigAxis,line width=0.4pt},
  ymajorgrids, grid style={A2ZFigAxis!66!white,line width=0.35pt},
  every axis label/.append style={A2ZFigText,font=\rmfamily\footnotesize},
  every tick label/.append style={A2ZFigText,font=\rmfamily\scriptsize},
  title style={A2ZFigText,font=\rmfamily\footnotesize\bfseries,at={(0.5,1)},anchor=south,yshift=3pt},
  legend style={draw=none,fill=white,font=\rmfamily\scriptsize,text=A2ZFigText},
  nodes near coords style={A2ZFigText,font=\rmfamily\scriptsize},
  enlarge x limits=0.2,
}}

\makeatletter
\let\AtoZAppendixOriginalAddContentsLine\addcontentsline
\renewcommand{\addcontentsline}[3]{%
  \AtoZAppendixOriginalAddContentsLine{#1}{#2}{#3}%
  \def\AtoZAppendixExtension{#1}%
  \def\AtoZAppendixTOCExtension{toc}%
  \ifx\AtoZAppendixExtension\AtoZAppendixTOCExtension
    \ifnum\csname toclevel@#2\endcsname<3\relax
      \addtocontents{atoc}{%
        \protect\contentsline{#2}{#3}{\thepage}{\@currentHref}%
        \protected@file@percent}%
    \fi
  \fi
}

\section*{Appendix Contents}
\begingroup
\small
\setcounter{tocdepth}{2}
\setlength{\parskip}{0pt}
\hypersetup{linkcolor=black}
\renewcommand{\l@section}[2]{%
  \addpenalty{-\@highpenalty}%
  \vskip 4pt plus 1pt
  \@dottedtocline{1}{0em}{1.6em}{\bfseries #1}{\bfseries #2}%
}
\renewcommand{\l@subsection}{\@dottedtocline{2}{1.6em}{2.5em}}
\@starttoc{atoc}
\endgroup
\makeatother
\clearpage

\section*{Appendix}

\section{Game Design Document Dataset}
\label{app:gdd_dataset}

\AtoZbench{} comprises 100 long-form game design documents (GDDs), divided into \emph{Small} and \emph{Big} splits of 50 each. Because the benchmark measures how faithfully a game implements its GDD, the documents define the evaluation targets, and their quality affects the validity of the resulting scores. Sections~\ref{app:golden_gdd} and~\ref{app:golden_construction} define the admission criteria and the validation-and-repair procedure. Section~\ref{app:gdd_scope} describes the splits and dataset statistics, and Section~\ref{app:gdd_examples} illustrates the document structure with one example from each split. Separately, Section~\ref{app:causal_game_spec} examines causal descriptions in external game-design corpora, motivating the dependency-aware representation introduced in Section~\ref{sec:dependency_contract}.

\subsection{Brief-to-GDD Pipeline}
\label{app:gdd_pipeline}

Each GDD is derived from a short game brief in two stages, via a creative vision (CV), as shown in Figure~\ref{fig:brief_cv_gdd}. The brief fixes the genre, a one-line premise, the core mechanic, reference games, and the intended scope, under constraints shared by every game: PC, 2D, single player, and keyboard-and-mouse input.
A CV generator first expands the brief into a CV with a fixed set of sections: core experience, opening fiction, hook, one pass of the core loop, mechanic intent, and explicit non-goals, plus sections chosen from the game's genre tags. The CV establishes the intended experience and scope. The harness described in Section~\ref{app:gdd_harness} then turns the CV into a GDD that specifies the entities, rules, formulas, stages, screens, and assets required for implementation. This separation establishes the design intent before its detailed specification and provides a reference for evaluating the GDD's fidelity to that intent.

\AtoZAppendixInput{figures/appendix/brief_cv_gdd.tex}
\AtoZAppendixBarrier

\subsection{Harness Optimization}
\label{app:gdd_harness}

An agent performs the CV-to-GDD step inside a harness: the generator's instructions, the project files it can consult, and its authoring-and-review workflow. Rather than tuning the harness by hand, we refine it with an iterative search inspired by Meta-Harness~\citep{lee2026meta}.

\paragraph{Search procedure.} The search uses a fixed set of $N$ CVs. Starting from a seed harness, each iteration has a proposer agent read the search history, which contains earlier harnesses, their GDDs, the evaluator scores, and the evaluators' feedback. The
proposer then produces $K$ candidate harnesses that modify the instructions, the project files, or the workflow. Each candidate generates a GDD from every CV, and the GDDs are scored as described below. The CVs, the fidelity checklists derived from them, and the evaluator configuration stay fixed across candidates to support a controlled comparison of the harnesses.

\paragraph{Document-quality scores.}
\emph{CV fidelity} is the fraction of items in a fixed, CV-derived checklist that the GDD satisfies. The checklist covers the CV's mechanics, constraints, design intentions, and any stated quantitative details. The full checklist is always the denominator, so an item the judge omits counts as unsatisfied.
\emph{Buildability} is computed over ten disciplines: gameplay, systems, economy, AI, UX, level design, art, audio, narrative, and production. For each applicable discipline, the judge lists the required deliverables and computes the fraction that are specified concretely and consistently. A mechanic or asset that is only named does not count. The score is the unweighted mean over
applicable disciplines.
The two scores check each other. Fidelity alone would reward restating the CV, and buildability alone would reward detailed specifications that drift from it. With $F_j(h)$ and $B_j(h)$ the scores of harness $h$ on the $j$-th CV,

\begin{equation}
    R(h)=\frac{1}{N}\sum_{j=1}^{N}\bigl(F_j(h)+B_j(h)\bigr),
    \label{eq:app_harness_reward}
\end{equation}
which weights both dimensions and all CVs equally. A candidate is scored only if both dimensions are usable for every CV. These scores assess the document and are separate from \GDDmetric{}, which assesses the game.

\paragraph{Harness selection.}
Rewards guide the search, but a candidate replaces the current best harness only if it wins a head-to-head comparison. For each CV, fresh judge sessions compare the two GDDs on every fidelity-checklist item and every applicable buildability discipline. Each item-level comparison is repeated three times independently, and the item-level winner is determined by majority vote; a tie is recorded as undecided. For each dimension, we then determine the CV-level winner by majority vote over the decided item-level outcomes. A tie at this stage is also recorded as undecided. Thus, each CV casts at most one vote per dimension. A candidate must win a strict majority of the decided CV-level votes. If multiple candidates meet this criterion, we select the one with the highest win rate among decided votes.

\paragraph{Configuration.}
Table~\ref{tab:harness_config} lists the search settings and selected checkpoint. All 100 benchmark GDDs are generated with the harness selected at iteration~8.
Harness search improves the typical GDD but does not guarantee that any individual GDD is defect-free, so every GDD must still meet the criteria in Section~\ref{app:golden_gdd}.

\begin{table}[!htbp]
\centering\small
\begin{tabular}{@{}ll@{}}
\toprule
\textbf{Setting} & \textbf{Value} \\
\midrule
CVs used in the search ($N$)                & 13 \\
Candidate harnesses per iteration ($K$)     & 3 \\
Repeats per quality judgment / pairwise comparison & 3 / 3 \\
Proposer and generator                      & Claude-Opus-4.8 \\
Evaluator (scoring and pairwise judging)    & GPT-5.5 \\
Maximum iterations                          & 20 \\
Selected checkpoint                         & iteration 8 \\
\bottomrule
\end{tabular}
\caption{\textbf{Harness search configuration.}}
\label{tab:harness_config}
\end{table}

\subsection{Admission Criteria}
\label{app:golden_gdd}

A conformance score such as \GDDmetric{} is meaningful only when the GDD specifies requirements clearly enough to evaluate and consistently enough to admit a coherent implementation. Missing or undefined requirements cannot be checked reliably, while contradictory requirements can make full conformance impossible: any implementation may violate at least one. These specification defects can therefore make low scores difficult to attribute to the agent's implementation.
A rule that can never take effect penalizes faithful implementations. A formula open to more than one reading can make the agent and the evaluator compute different correct answers.

We call a document a \emph{Golden GDD} after it passes the validation-and-repair procedure in Section~\ref{app:golden_construction} with no detected violations of the four criteria in Table~\ref{tab:golden_criteria}. These criteria target the specification defects described above. G1 and G4 apply to individual statements. G2 and G3 apply to the document as a whole. G2 is static: the text agrees with itself. G3 is dynamic: the rules still hold together when the game runs. A document can pass G2 and still fail G3. For example, two consistent statements can together leave a mechanic that is impossible to reach.

\paragraph{Making the criteria checkable.}
GDDs use three conventions that support systematic checks of G2--G4.
(i) Every tunable value has a canonical definition with a stable identifier (\texttt{DBT-}$n$) in a \emph{design balance table}. Other sections refer to that identifier. Any repeated value must agree with its canonical definition. This makes references and repeated values traceable during G2 checks.
(ii) Discrete behavior is written as a \emph{rule table} (\texttt{RT-*}). Its rows are evaluated top to bottom. The first match applies, and a final \texttt{ELSE} row covers every remaining case, which makes the totality and determinism required by G3 visible.
(iii) Any value computed from several sources is written as a \emph{ledger} (\texttt{LED-*}). A ledger fixes the evaluation order and the bounds and gives a worked numeric example that a G4 check can recompute.
All 100 GDDs use balance tables, 83 use rule tables, and 35 use ledgers. Most ledgers appear in the \emph{Big} split, where values combine across systems.

\AtoZAppendixInput{tables/appendix/criteria_for_gdd}

These criteria concern the GDD's validity as a specification. A Golden GDD may describe a simple or an unambitious game. The CV fidelity and buildability scores used during harness search (Section~\ref{app:gdd_harness}) assess alignment with the CV and implementation readiness separately.

\subsection{Validation and Repair}
\label{app:golden_construction}

\paragraph{Candidate generation.}
The briefs span a diverse range of genres, core-mechanic families, and production scopes. Each brief expands into a CV and then into a GDD using the harness detailed in Section~\ref{app:gdd_harness}.

\paragraph{Validation.}
Validate-Agents check each candidate against G1--G4, with candidates distributed across validators running in parallel. In the initial validation, the system checks every applicable table, formula, and cross-reference. It does not sample within a document. In subsequent rounds, documents changed during repair are fully re-validated, while unchanged documents are spot-checked for likely regressions. The checks are applied wherever the GDD contains the relevant structure:
\begin{itemize}\itemsep0pt
  \item \textbf{G1}: no \texttt{[TBD]} or placeholder remains; every cited section and identifier exists.
  \item \textbf{G2}: recurring values agree across entity tables, systems, the stage catalog, and level descriptions. Different descriptions of the same stage agree on wave count, objectives, enemy composition, and learning goal.
  \item \textbf{G3}: difficulty arrays are non-decreasing; the first appearance declared for each entity or environment object corresponds to the stage catalog; timers and triggers are reachable given the interacting systems.
  \item \textbf{G4}: boundary tables are recomputed from their formulas and arrays; derived quantities such as $\lceil \mathit{HP}/\mathit{dmg}\rceil \times t_{\text{loop}}$ are re-derived; every exception to a formula is listed.
\end{itemize}

\paragraph{Repair and acceptance.}
Each violation is fixed at its canonical definition. When two sections disagree, the more specific one is kept, and the other is changed to match it, and every fix is logged with its criterion. Documents changed in a round are fully re-validated in the next round. Documents that did not change are spot-checked on the properties most likely to regress. A document is accepted
only after three consecutive rounds with no detected violations.
As an illustration, in one batch of 13 candidates, the logged rounds record 24 fixes, after which the batch passes three clean rounds in a row. Most fixes are numeric values that disagree with the document's own worked examples, mismatches between the stage catalog and the level descriptions, and first-appearance columns that do not match the catalog.

\paragraph{Implications for evaluation.}
The accepted documents form the 100 Golden GDDs in the benchmark. Passing G1--G4 reduces the risk that low conformance scores arise from detectable specification defects, although evaluator and implementation-interpretation errors can remain. Stable identifiers for constants, rules, and formulas provide fixed anchors for checking individual requirements and locating departures from the specification.

\subsection{Dataset Splits and Statistics}
\label{app:gdd_scope}

Accepted GDDs are organized into \emph{Small} and \emph{Big} splits of 50 each according to the scope of the specified game: the mechanics, content, and interacting systems to be implemented, rather than the length or quality of the document. Document lengths and extracted requirement counts are reported as dataset statistics and are not used as thresholds for assigning the splits.

\emph{Small} GDDs focus on a single core action or a compact set of mechanics, typically within a single screen and a short session. Each specifies clear success and failure conditions, together with input handling, state transitions, feedback, and boundary cases. Campaigns, progression systems, shops, and narrative content outside the core task are excluded. These restrictions limit the scope of the game without reducing specification detail: each GDD remains self-contained and includes the rules, numerical values, inputs, UI text, content, art, audio, and testing requirements needed for implementation. \emph{Small} GDDs contain approximately 54 outcome requirements on average, with a range of 9--151.

\emph{Big} GDDs cover broader content and interacting systems, including combinations of combat, progression, economy, exploration, narrative, and management. The split contains complete 2D single-player designs. Specifications that require 3D or 2.5D presentation, mandatory multiplayer, or further adaptation to the target setting are excluded, as are games that duplicate a \emph{Small} entry. \emph{Big} GDDs contain approximately 84 outcome requirements on average, with a range of 10--246.

To limit repetition within the \emph{Big} split, the default selection policy admits at most two entries from the same narrow genre and core-mechanic family. Within a broad genre, designs are treated as distinct families only when their primary inputs, repeated actions, success and failure criteria, and progression structure all differ. This allows several games from one broad genre while limiting repeated instances of the same core task.

The document lengths, requirement composition, and category coverage for the two sets are summarised in Table~\ref{tab:gdd_statistics}. Outcome requirements count individual promises listed in GDD rule-table rows, including presentation outcomes. A rule in the dependency graph specifies a source row's conditions, trigger, and state-changing or event-emitting outcomes. On average, the \emph{Big} GDDs contain about $1.87\times$ as many tokens and $1.55\times$ as many outcome requirements as the \emph{Small} GDDs. The 50 \emph{Small} GDDs cover eight task families (Table~\ref{tab:small-family-distribution}), and the 50 \emph{Big} GDDs cover 11 commercial game genres (Table~\ref{tab:big-genre-distribution}).

\AtoZAppendixInput{tables/appendix/gdd_dataset_stats}
\AtoZAppendixInput{tables/appendix/small_gdd_feats}
\AtoZAppendixInput{tables/appendix/big_gdd_feats}
\AtoZAppendixBarrier

\subsection{Examples of GDDs}
\label{app:gdd_examples}

The following examples show how the same document conventions apply to both splits.

\begin{figure*}[p]
  \centering
  \includegraphics[width=\textwidth,height=0.88\textheight,keepaspectratio]{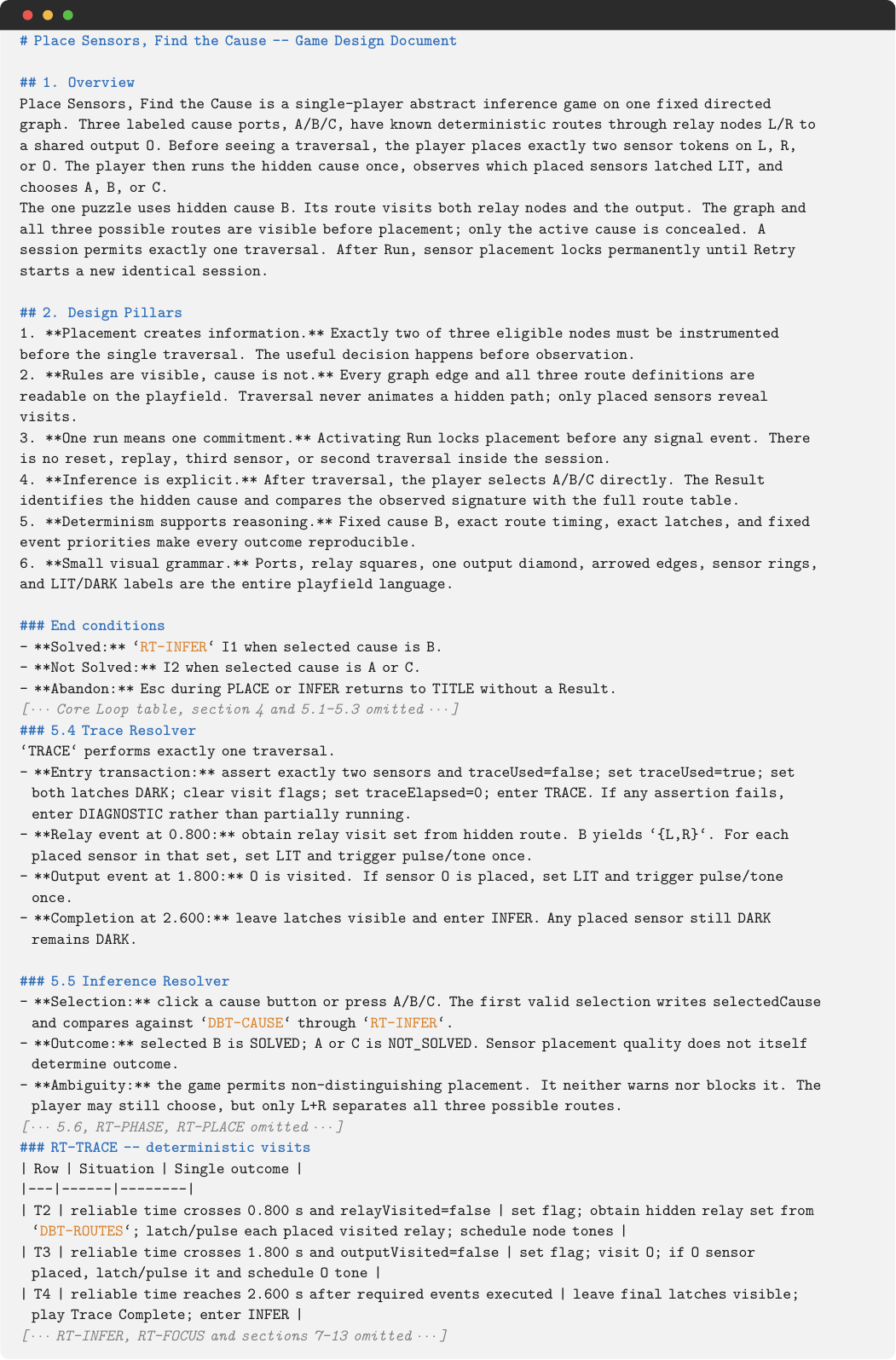}
  \caption{\textbf{A \emph{Small} golden GDD example.} Excerpt of \texttt{traces\_left}; omitted sections are marked.}
  \label{fig:gdd-window-small}
\end{figure*}

\begin{figure*}[p]
  \centering
  \includegraphics[width=\textwidth,height=0.88\textheight,keepaspectratio]{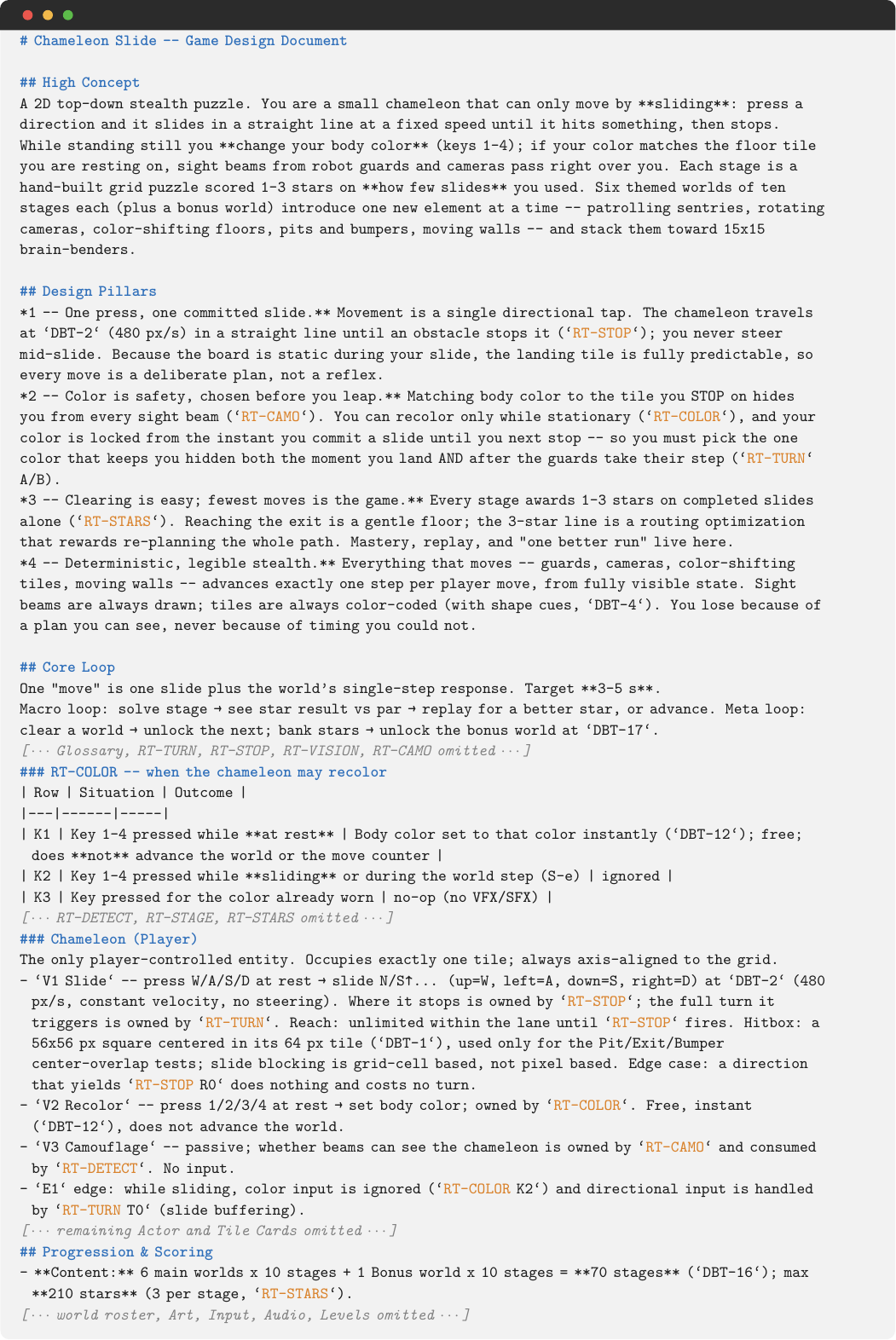}
  \caption{\textbf{A \emph{Big} golden GDD example.} Excerpt of \texttt{chameleon\_slide} GDD; omitted sections are marked.}
  \label{fig:gdd-window-big}
\end{figure*}

Figure~\ref{fig:gdd-window-small} shows an excerpt of the \emph{Small}-split GDD \texttt{traces\_left}. Its prose describes the game and its systems, while rule-table rows state situations and their required outcomes. Constants such as the 0.800\,s and 1.800\,s deadlines have canonical definitions that other sections reference. Each labeled row provides a requirement that can be checked and cited: for example, a build that leaves the phase unchanged when \texttt{T4}'s stated conditions and trigger hold fails that row's required outcome. Selected rules from these rows appear in the contract graph in Figure~\ref{fig:contract-graph}.

Figure~\ref{fig:gdd-window-big} shows a \emph{Big}-split GDD, \texttt{chameleon\_slide}, under the same conventions. The form doesn't change, but only the reach of each statement does. The pillars and the player card describe one action by citing several rule tables at once --- a slide is stopped by \texttt{RT-STOP}, hides the player by \texttt{RT-CAMO}, and advances the world by \texttt{RT-TURN} --- and the rule shown, \texttt{RT-COLOR} \texttt{K1}, is one that a build must honor in the middle of that chain.

\clearpage

\section{Details on Contract Generation}
\label{app:gdd-contract-construction}

\subsection{Dependency Structure in Game Specifications}
\label{app:causal_game_spec}
As summarized in Figure~\ref{fig:intro_analysis_causal}, we measure how often sentences in external game-design documents are classified as causal, on three corpora separate from the benchmark GDDs: 10 publicly available studio design documents ranging from short pitch documents to long-form production specifications (Table~\ref{tab:studio_gdd_causality}; 11,741 sentences, 37.9\% causal), 246 specifications from GameCraft-Bench~\citep{luo2026gamecraft} and GameGen-Verifier~\citep{jia2026gamegen} (141 and 105 documents), and 97 English-language primary rulebooks of BoardGameGeek's top-100 titles covering setup, play, and end conditions, without player aids, FAQs, or fan summaries; in total 353 documents and 73,700 sentences.

We normalize the extracted text, remove formatting artifacts and non-prose content, segment the prose into sentences, and label each sentence causal or non-causal with the published CiRA classifier~\citep{fischbach2021cira}; a corpus's causal rate is the share of retained sentences labeled causal. As shown in Table~\ref{tab:causality_corpora}, the rate ranges from 32.0\% to 45.3\% across the three corpora, with a corpus macro average of 38.4\% (the value in Figure~\ref{fig:intro_analysis_causal}) and a pooled sentence-level rate of 42.0\%, against roughly 28\% reported for general requirements documents~\citep{frattini2023causality}; causal statements thus form a substantial portion of game-design specifications across sources.

\begin{table}[!htbp]
    \centering
    \caption{\textbf{Studio game-design documents used in the causality analysis.} Sentence counts are computed after text normalization and non-prose filtering; causal rates are CiRA predictions over the retained sentences.}
    \label{tab:studio_gdd_causality}
    \small
    \setlength{\tabcolsep}{6pt}
    \renewcommand{\arraystretch}{1.06}
    \begin{tabular}{@{}lrr@{}}
        \toprule
        \textbf{Document} & \textbf{Sentences} & \textbf{Causal (\%)} \\
        \midrule
        \texttt{The Sky Above, the Sky Below} & 2,069 & 38.3 \\
        \texttt{Frontier Pharmacist} & 1,996 & 42.5 \\
        \texttt{Leisure Suit Larry 5} & 1,701 & 38.3 \\
        \texttt{Love for Sail} & 1,551 & 37.5 \\
        \texttt{Shape Up or Slip Out} & 1,444 & 44.6 \\
        \texttt{Claw} & 1,428 & 29.6 \\
        \texttt{Leisure Suit Larry's Casino} & 797 & 31.2 \\
        \texttt{Planescape: Torment} & 483 & 34.8 \\
        \texttt{Diablo} (pitch) & 143 & 35.0 \\
        \texttt{Grand Theft Auto} & 129 & 29.5 \\
        \midrule
        \textbf{Total} & \textbf{11,741} & \textbf{37.9} \\
        \bottomrule
    \end{tabular}
\end{table}

\AtoZAppendixInput{tables/appendix/gdd_causal_dataset}
\AtoZAppendixBarrier

\subsection{Rule Generation}

\begin{figure*}[!tbp]
    \centering
    \includegraphics[width=\textwidth]{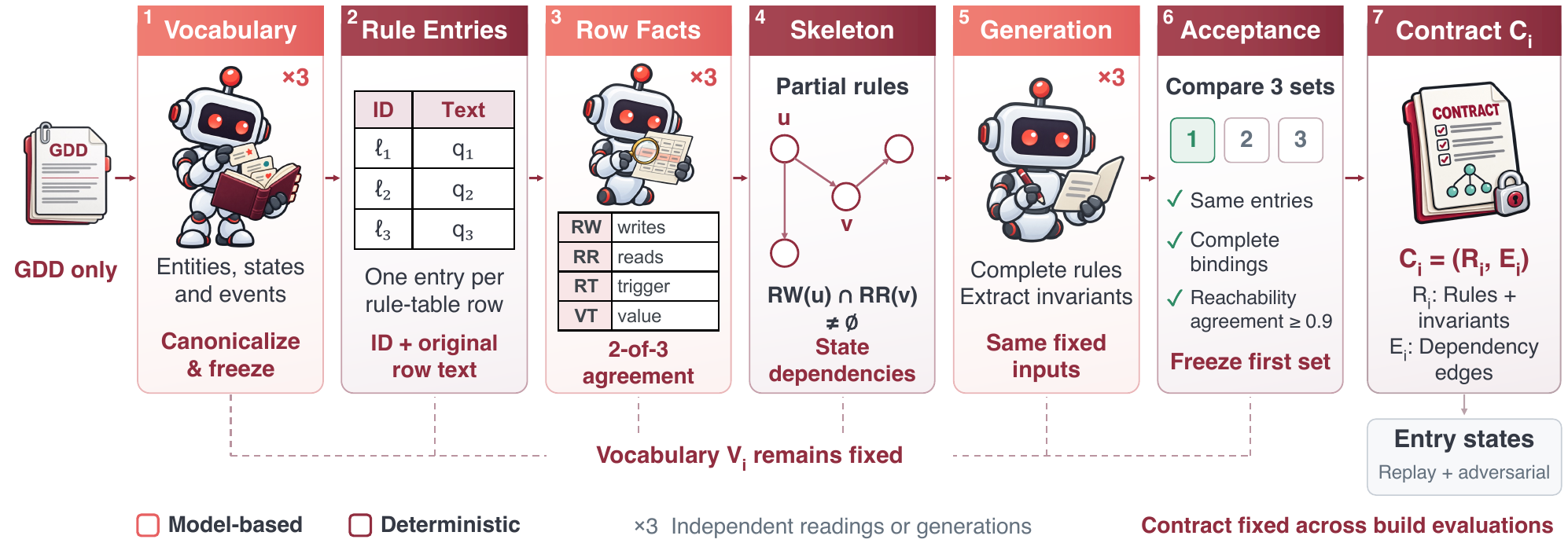}
    \vspace{-10pt}
    \caption{\textbf{Contract construction from a GDD.} The pipeline fixes vocabulary and rule entries before completing rule definitions and validating the contract. Coral boxes denote model-based stages, while dark-red boxes denote deterministic stages without model calls. The \texttt{$\times 3$} labels indicate independent repetitions: row facts are consolidated by two-out-of-three voting, whereas generated contracts are compared for structural consistency before one is frozen for evaluation. The dashed line indicates that the vocabulary $V_i$ remains fixed throughout subsequent stages.}
    \label{fig:contract-construction}
    \vspace{-5pt}
\end{figure*}

We construct the contract $\mathcal{C}_i=(\mathcal{R}_i,\mathcal{E}_i)$ from the GDD $d_i$ before inspecting any generated build. The procedure separates identifying the requirements from formalizing their conditions and effects: it first fixes the terminology and source entries, then uses them to constrain rule generation. The six stages in Figure~\ref{fig:contract-construction} combine model-based interpretation with deterministic extraction, assembly, and validation.

A rule $r=(t_r,\phi_r,F_r)$ specifies an event $t_r$, the preconditions $\phi_r$ under which it applies, and the expected effects $F_r$, expressed as state updates or event emissions. An invariant $r=(\varphi_r,\sigma_r)$ instead specifies a condition $\varphi_r$ that must hold throughout its scope $\sigma_r$. Dependencies in $\mathcal{E}_i$ connect a rule's effects to the conditions or triggering events of other rules. These links are derived from the rule fields rather than generated as a separate list of relationships.

\paragraph{Stage 1: Vocabulary (Model).} We establish a shared vocabulary $V_i$ for the entities, state attributes, and events described in the GDD. Three independent readings collect candidate terms, after which synonymous terms are grouped and assigned a common name. The resulting vocabulary is fixed for subsequent stages. This lets requirements that refer to the same game state use the same identifier, even when the GDD describes that state in different ways.

\paragraph{Stage 2: Rule Entries (Code).} Code extracts candidate rule entries $L_i$ from the GDD's labeled requirement tables. Each entry $\ell=(\mathrm{id}_{\ell},q_{\ell})$ retains its identifier and original row text. Generated rules remain linked to these entries through the provenance map $\pi:\mathcal{R}_i^{\mathrm{rule}}\to L_i$. The source entries therefore determine which requirements are to be formalized, rather than leaving each generation to choose a new set of evaluation targets.

\paragraph{Stage 3: Row Facts (Model).} For each entry, the model identifies four facts using the fixed vocabulary: the state attributes it changes, $\mathrm{RW}(\ell)$; the attributes its conditions inspect, $\mathrm{RR}(\ell)$; its triggering event, $\mathrm{RT}(\ell)$; and the operation and value associated with each update, $\mathrm{VT}(\ell)$. A condition-only rule has no triggering event. Each extraction is repeated three times independently. Read and write sets retain elements supported by at least two readings, while triggers and update values require two matching answers. A trigger or value without a majority remains unresolved and is completed from the source text during generation.

\paragraph{Stage 4: Skeleton (Code).} Given the fixed row facts, code assembles partial rule definitions and a state-dependency graph. Two entries are linked when one updates an attribute inspected by the other, that is, $\mathrm{RW}(\ell_u)\cap\mathrm{RR}(\ell_v)\neq\varnothing$. Through the source-entry associations, these links determine the state dependencies used to compute the downstream counts in Equation~\ref{eq:dependency_weights}. Each partial rule also records its known trigger, the attributes used by its preconditions, and the attributes changed by its effects. The next stage completes these fields rather than reconstructing the rule set. Event dependencies are derived from emitted events and matching triggers once the full rule fields are available.

\paragraph{Stage 5: Generation (Model).} The model completes each partial rule using its source text $q_{\ell}$, writing precondition predicates and effects consistent with the fixed read/write facts and update specifications. Any unresolved trigger or value is determined from the same source text. When a value cannot be expressed as a literal or supported symbolic reference, its document description is retained as an \emph{opaque} value instead of inventing a constant. Invariants are extracted separately from the GDD's state and constraint specifications, with supporting quotations. This stage produces three independent rule and invariant sets under the same fixed inputs.

\paragraph{Stage 6: Acceptance (Code).} Programmatic checks validate the generated fields and their source associations, and recompute the dependency graph from the rule definitions. The three generations are compared for source-entry consistency, provenance binding, and dependency reachability. The acceptance check requires the same source-entry set, complete rule-to-entry binding, and reachability agreement of at least 0.9. Once accepted, the first generation is frozen as $\mathcal{C}_i$ for all subsequent build evaluations. This fixes the detailed rule definitions as well as the evaluation targets and dependencies; Section~\ref{app:contract_stability} examines the variation that remains between independent generations.

\paragraph{Entry States.} Requirements sharing an entry condition are grouped into the canonical scenario set $\mathcal{S}_i$ used for replay evaluation. Each scenario specifies the state a build must establish before the replay begins. For targeted adversarial playtests, the harness similarly initializes the preconditions of a rule left unverified by normal play. Initialization supplies the starting conditions, not the expected outcome: the subsequent behavior must still be exercised and observed. These entry specifications support execution of the contract and are kept separate from the rules and dependency edges in $\mathcal{C}_i$.

\AtoZAppendixBarrier

\subsection{Properties of the Accepted Contract}

The contract accepted after Stage 6 provides a fixed representation for build evaluation. First, it separates triggers, preconditions, and effects, giving each judge explicit fields against which to organize its evidence.
Second, dependencies are derived from the rule fields using a fixed vocabulary.
State dependencies connect rules when one updates an attribute that another inspects, while event dependencies connect an emitted event to a matching trigger.
Reach, roots, and leaves are computed from the resulting graph, which identifies downstream requirements that may be affected by a failed rule.
Third, the evaluation targets are fixed before any build is inspected. Each accepted rule retains its source association with the GDD, and values that cannot be formalized are preserved as document descriptions rather than replaced with invented constants. Every build is evaluated against the same accepted contract.

\subsection{Contract Example}
\label{app:contract_example}

\AtoZAppendixInput{figures/appendix/appendix_contract_graph}
\AtoZAppendixInput{figures/appendix/appendix_contract_gdd_rows}

Figures~\ref{fig:contract-graph} and~\ref{fig:contract-gdd-rows} follow \texttt{traces\_left} from its GDD to a judged normal playtest. The GDD is shown in Figure~\ref{fig:gdd-window-small}. Each source row describes a situation and an outcome, which are formalized as the rule's trigger, preconditions, and effects. For example, \texttt{P2} adds a sensor when an eligible node is empty, the phase is \textsc{Place}, and fewer than two sensors are placed. It therefore both reads and updates \texttt{sensorNodes}.

Figure~\ref{fig:contract-graph} is an execution view linked to the fixed contract, rather than the complete static contract graph. Repeated nodes represent occurrences of the same rule during the recorded playtest. Three placements and one removal establish the sensor count required by \texttt{P4}. Committing the run sets the phase and initializes the clock used by subsequent timing rules; \texttt{T4} completes the trace, after which \texttt{I1} records the outcome. The focus-handling sequence is displayed separately for readability, although its rules also share \texttt{phase} with the main sequence.

The example illustrates how source-linked rules make execution evidence interpretable in its dependency context. A missing prerequisite can leave downstream behavior unverified, but does not automatically make every downstream rule violated. Section~\ref{app:contract_stability} examines consistency across contract generations, and Section~\ref{app:ablation-dependency-weight} examines how downstream reach affects source-code scoring.

\subsection{Stability of Contract Generation}
\label{app:contract_stability}

Contract construction is partially deterministic: model-based extraction supplies the vocabulary and row facts, code fixes the rule entries, binds them to their source GDD rows, and assembles the state-dependency skeleton from the extracted read/write sets, and rule generation then completes the preconditions and effects within that structure. In this fixed-input stability experiment, we independently generate rules three times for each of the 50 \emph{Small} and 50 \emph{Big} GDDs under the same fixed inputs and compare the 300 outputs.

As shown in Table~\ref{tab:contract-stability}, all three generations preserve the same rule entries and source associations, and the dependency graphs are nearly identical, with reachability agreement, dependency-edge Jaccard similarity, and role preservation all at or above 0.999; within each rule, the \texttt{trigger} name agrees in 0.94 of cases and the \texttt{predicate} structure, the entities, attributes, and operators used in preconditions and effects, in 0.70. Table~\ref{tab:predicate-examples} shows what the remaining variation looks like: the same row encoded with \texttt{SET} or \texttt{INCREMENT}, or a bound written as a comparison or as a set.

We can check whether this variation changes scores directly, because each generation was also used to judge the same build. 
Here, the unit is a rule in the dependency graph, which represents a source row's conditions, trigger, and state-changing or event-emitting outcomes. The 869 \emph{Small} and 1,860 \emph{Big} rules are therefore fewer than the individual outcome requirements counted in Table~\ref{tab:gdd_statistics}.
Pairing the three judgments for each rule yields three pairs per rule: 2,607 over the 869 \emph{Small} rules and 5,580 over the 1,860 \emph{Big} rules, with no rule excluded since the generations share the same entries. 
Re-judging a rule whose predicate structure is identical across two generations changes its score by at least 0.25 in 12.7\% of pairs on \emph{Small} and 8.8\% on \emph{Big}, a variation that comes with an agent judge and is of a similar magnitude to the status changes of the naive judge on its matched items; when the predicate structure differs, the rate is 17.7\% and 11.1\%, an increase of 5.0 and 2.3 percentage points (Table~\ref{tab:predicate-verdict}). 
Freezing the accepted contract keeps the evaluation targets, rule definitions, and dependencies fixed across builds and revision rounds. The generation analysis quantifies the variation that would otherwise arise from regenerating those definitions.

\begin{table}[!htbp]
    \centering\small
    \setlength{\tabcolsep}{6pt}
    \renewcommand{\arraystretch}{1.08}
    \caption{\textbf{Consistency of evaluation targets across repeated generations.} Three rule generations per GDD on 50 \emph{Small} and 50 \emph{Big} GDDs with fixed vocabulary and rule entries, averaged over the two splits; items are matched by their source entries.}
    \begin{tabular}{@{}lc@{}}
        \toprule
        \textbf{Agreement Measure} & \textbf{Contract} \\
        \midrule
        Matched Items & 100\% \\
        Dependency Graph & $\geq 0.999$ \\
        \texttt{Trigger} Name & 0.94 \\
        \texttt{Predicate} Structure & 0.70 \\
        \bottomrule
    \end{tabular}
    \label{tab:contract-stability}
\end{table}

\begin{table}[!htbp]
  \centering\scriptsize
  \setlength{\tabcolsep}{3pt}
  \renewcommand{\arraystretch}{1.15}
  \caption{\textbf{Examples of predicate-structure variation across contract generations.} Rules whose predicate structure differs between two contract generations, with the GDD row they encode and the source-code score $f_{i,r}^{\mathrm{src}}$ each generation received on the same build.}
  \begin{tabularx}{\textwidth}{@{}l>{\raggedright\arraybackslash}X>{\raggedright\arraybackslash\ttfamily}X>{\raggedright\arraybackslash\ttfamily}Xc@{}}
    \toprule
    \textbf{Game \& Rule} & \textbf{GDD statement} (abridged) & \normalfont \textbf{Generation $A$} & \normalfont \textbf{Generation $B$} & \textbf{Ratio of} $f_{i,r}^{\mathrm{src}}$ ($A / B$) \\
    \midrule
    \texttt{untangled\_cords} R8 (\emph{Small}) & All other frames: keep state, keep updating \texttt{t} in \textsc{Play\_Active} & timer INCREMENT $\langle$dt$\rangle$ & timer SET $\langle$advanced by dt$\rangle$ & 1.00 / 1.00 \\
    \texttt{magma\_arc} STAR2 (\emph{Big}) & Not STAR3, and hit events $\leq 3$ this attempt: 2 stars & event <= 3 \newline STAR SET 2 & event IN [1, 2, 3] \newline STAR SET 2 & 1.00 / 1.00 \\
    \bottomrule
  \end{tabularx}
  \label{tab:predicate-examples}
\end{table}

\begin{table}[!htbp]
  \centering\small
  \setlength{\tabcolsep}{5pt}
  \caption{\textbf{Effect of predicate-structure variation on the source-code score.} Rule pairs across the three generations, three per rule, judged on the same build; the last column is the share of pairs whose per-rule scores differ by at least 0.25.}
  \begin{tabular}{lrrlrc}
    \toprule
    \textbf{Split} & \textbf{games} & \textbf{rules} & \textbf{Predicate structure} & \textbf{pairs} & $|\Delta f_{i,r}^{\mathrm{src}}|\geq0.25$ \\
    \midrule
    \emph{Small} & 50 & 869   & identical & 1,773 & 12.7\% \\
                 &    &       & different & 834   & 17.7\% \\
    \emph{Big}   & 50 & 1,860 & identical & 4,045 & 8.8\% \\
                 &    &       & different & 1,535 & 11.1\% \\
    \bottomrule
  \end{tabular}
  \label{tab:predicate-verdict}
\end{table}
\AtoZAppendixBarrier

\subsection{Value Compatibility}

In construction, a state-dependency edge in $\mathcal{E}_i^{\mathrm{state}}$ records that one rule updates an attribute inspected by another rule's condition, without checking the written value against that condition, so two rules that write and read the same enumerated attribute with different values are linked as well. 
After removing state-dependency edges whose written values are provably incompatible with the reading conditions, rule-score rankings remain stable: the dependency-weighted rule score $F_i^{\mathrm{rule}}$ computed on the reduced graph agrees with the registered one at Spearman 0.99 on \emph{Small} and 1.00 on \emph{Big} with a maximum per-game difference of 0.07, the defining root of each root-defective game remains a root in 17 of 18 games, and the sign counts of Appendix~\ref{app:ablation-dependency-weight} stay at 11 of 18 root-defective games and move from 6 to 7 of 16 comparison games. We therefore keep the attribute-level definition, which is computable from the contract alone, and read its edges as potential rather than guaranteed prerequisites.

\subsection{Dependency-Adjusted Rule Pass Rate}
\label{app:dependency_analysis}

Using the source-code judgments and contract graphs of 100 GPT-5.6-Sol games, we compare local and dependency-adjusted rule pass rates. A rule passes locally when its aggregated source-code score is at least 0.8. The dependency-adjusted rate serves as a \emph{graph-based reachability proxy}: a rule counts only when it and all its direct and indirect predecessors pass. Both rates use the same dependency-linked rule set as the denominator within each game and are averaged across games. Figure~\ref{fig:intro_analysis_dependency} reports 72.4\% locally and 22.7\% after this check, a 49.7-point gap. Since edges encode potential prerequisites, exclusion indicates possible blocking rather than demonstrated runtime unreachability. This rule-level diagnostic does not change the benchmark's source-code score or propagate violation verdicts to downstream rules during playtesting.

\clearpage

\section{Experimental Details}
\label{app:experimental_details}

\paragraph{Game Development Environment.}
The 100 tasks in the main benchmark focus on 2D single-player browser games developed using Phaser. All coding agents start with the same project and runtime-interface specification, using Phaser~4.1, TypeScript, and Vite. Given a GDD, the agent creates the source project along with a runnable web build within this environment. The fixed evaluation contract is kept apart from the game-building instructions. This common development environment enables a controlled comparison across coding agents. The extension to a Three.js runtime is presented in Section~\ref{sec:3d_extension}, with additional results in Appendix~\ref{app:scalable_benchmark}.

\paragraph{Open-Weight Model Serving and Generation.}
We generate games with DeepSeek-V4-Pro-0813~\citep{deepseekai2026deepseekv4}, GLM-5.3~\citep{glm5team2026glm5}, and Kimi-K2.7-Code~\citep{moonshot2025kimik2}, each served with vLLM~0.28.0~\citep{kwon2023efficient} using four NVIDIA B300 GPUs per allocation on a shared \texttt{slurm} cluster. Each cluster node has eight NVIDIA B300 GPUs, 128 vCPUs, and about 4~TB RAM. GLM and Kimi use tensor parallelism across the four allocated GPUs, while DeepSeek uses data parallelism with expert parallelism. Generation follows each checkpoint's default sampling parameters: temperature~1.0 and top-$p$~0.95 for GLM and Kimi, and temperature~1.0 and top-$p$~1.0 for DeepSeek. For each model, we independently generate a game three times per GDD without best-of-$k$ selection. Serving allocations total approximately 1,100 GPU-hours.

\paragraph{Evaluation Execution and Resources.}
For runtime evaluation, we serve production builds locally and run them in Chromium via Playwright at each game's declared viewport ($1280\times720$ by default). Scenario-based replay applies recorded input sequences and captures rendered frames, while adaptive playtesting selects keyboard and pointer inputs based on observations of the running game. Generated games expose shared interfaces for state observation and controlled precondition initialization. Section~\ref{app:eval_details} details the three evaluation axes, including how these interfaces support normal and adversarial playtests. Evaluation workloads run on ten AWS \texttt{m7i.4xlarge} instances, each with 16 vCPUs, 64~GiB RAM, and Ubuntu Server~24.04 LTS (noble) on x86-64 hardware, plus one workstation with comparable specifications. The total measurement time is approximately 900 wall-clock hours (source-code evaluation takes 3.1\%, scenario-based replay assessment takes 40.1\%, and playtest takes 56.8\%). Games are rendered and replayed on headless Chromium 149.0.7827.55 (Chrome Headless Shell) via Playwright 1.61.0.

\clearpage

\section{Details on Evaluation Setup}
\label{app:eval_details}

\subsection{Source Code Evaluation}
\label{app:source_eval}

\subsubsection{Prompt}
\label{app:source_eval_prompt}
\AtoZAppendixInput{figures/appendix/prompt/fig_source_prompt}

Each source-code evaluation session uses one prompted model call per build. Its system prompt states the question, the continuous scale for rules with its calibration anchors, the two-valued outcome for invariants with the frame-boundary and injection-surface exclusions, and the evidence requirement; Figure~\ref{fig:source-prompt} reproduces the passages that shape the judgment. The task message is assembled by code from three parts: the rule set $\mathcal{R}_i^{\mathrm{rule}}$ in the order to be judged, the invariant set $\mathcal{R}_i^{\mathrm{inv}}$, and the path of the read-only source directory of $g_i$. The judge answers through a fixed-schema tool call with one entry per rule ($f_{i,r}^{\mathrm{src}}\in[0,1]$ and evidence) and one per invariant ($f_{i,r}^{\mathrm{src}}\in\{0,1\}$ and evidence), and a validator rejects a submission that omits an item, breaks the order, or leaves evidence empty.

\subsubsection{Implementation}
\label{app:source_eval_implementation}
\paragraph{Inputs and Session.}
The source-code judge receives the rule set $\mathcal{R}_i^{\mathrm{rule}}$ and the invariant set $\mathcal{R}_i^{\mathrm{inv}}$ for contract $\mathcal{C}_i$, along with read-only access to the \texttt{src/} directory of build $g_i$. For each item, the judge answers a static question: is this commitment implemented in the code? The judge does not execute the game. Each session evaluates a single build and must return exactly one entry per rule and per invariant, in the specified order, each accompanied by a score and supporting evidence. Submissions that omit items, disrupt the order, or lack evidence are rejected and must be resubmitted. Given the density of generated sources, the judge examines entire files rather than inferring absence from line counts, and does not reuse findings as evidence for multiple rules.

\paragraph{Scoring a Rule.}
Each rule is decomposed into preconditions, trigger, and effects, accompanied by its originating GDD sentence. The judge assigns a continuous score $f_{i,r}^{\mathrm{src}}\in[0,1]$ reflecting the extent to which the causal commitment is implemented, specifically whether the code gates the trigger on the stated conditions and produces the specified effects. The scoring scale is anchored: 1.00 indicates full implementation as specified, 0.85 denotes a minor deviation that does not alter behavior in the stated cases, 0.5 represents partial implementation (such as an unenforced condition or missing effect), 0.2 indicates only traces of implementation, and 0.00 signifies absence or contradiction. Intermediate values are permitted when they more accurately reflect the degree of implementation. Contract names are mapped to code identifiers by meaning. Rules are evaluated as specified, not as reasonable alternatives, and the assessment is existential; potential corruption by other code paths is addressed through invariant checks.

\paragraph{Checking an Invariant.}
An invariant is defined as a predicate over the game state that must hold at every frame boundary within its declared scope. The judge assigns $f_{i,r}^{\mathrm{src}}\in\{0,1\}$: a value of 1 if the code establishes the property and no reachable gameplay path can violate it within scope, and 0 if any path can break it or if the property is not established. Differences in encoding, such as a 0-based index representing 1-based numbering or alternative enum spellings, are not considered violations. The state-injection surface present in every build for the verifier (\texttt{setState}, scene mutators, scenario entries) is not treated as a violating path, as it is designed to permit arbitrary state modifications.

\paragraph{Evidence.}
Every score cites the file and the function or line region the judge read, with one sentence on how that code implements or fails the item; a rule the judge cannot locate scores 0, and its evidence records what is searched. A validator rejects submissions whose evidence strings largely copy one another or that score every rule zero without locating anything.

\paragraph{Aggregation.}
The rule component $F_i^{\mathrm{rule}}$ is calculated as the dependency-weighted mean of the continuous per-rule scores, using the weights $w_{i,r}$ defined in Equation~\ref{eq:dependency_weights}. Here, $D_{i,r}$ represents the number of rules reachable from $r$ via links where one rule updates a state that another rule's condition evaluates. All weights are set to 1 when $D_i^{\max}=0$. The invariant component $F_i^{\mathrm{inv}}$ is the fraction of invariants that hold, and is set to 1 if no invariants are declared for a contract. The overall source-code score is the product of these two components (Equation~\ref{eq:source_score}).

\paragraph{Runs.}
Each build is evaluated in three independent sessions, and the mean score is reported. The default source-code judge is GPT-5.6-Luna at high reasoning effort. Appendix~\ref{app:cost_evaluation} presents judge comparisons and cost analyses.

\AtoZAppendixBarrier

\subsection{Scenario-Based Replay Assessment}
\label{app:replay_eval}

\subsubsection{Prompt}
\label{app:replay_eval_prompt}

\AtoZAppendixInput{figures/appendix/prompt/fig_replay_prompts}

The replay judge is two prompted model calls per scenario replay. The frame-selection call receives the rubric $\rho_i^{\mathrm{vis}}$, the executed input trace $\xi_{i,q}$ with the engine's render timestamps, and the observation budget $M$, and returns a plan of replay positions per rubric item that a deterministic tool resolves to recorded frames; the scoring call receives only the resolved frames and the rubric and returns one score, one rationale, and the cited frame ids per item through a fixed-schema tool call. The scoring prompt is prefixed with the GameCraft-Bench scoring instruction, which defines the $0$--$1$ scale; Figure~\ref{fig:replay-prompts} reproduces the passages of both prompts that shape the evidence.

\subsubsection{Implementation}

The scenario-based replay assessment described in Section~\ref{sec:replay_eval} keeps the evaluation protocol of GameCraft-Bench~\citep{luo2026gamecraft} where the protocol is engine-neutral: the build gate, the four rubric categories with weights $0.15/0.35/0.15/0.35$, the per-item score in $[0,1]$, the per-item aggregation over replays by max or mean, and the score formula are reused unchanged, and only the Godot replayer is replaced by a Playwright replayer for browser builds, with the same trace format of timed input events. Three parts differ, because the contract rather than the submitter decides what is played, where the evidence is taken, and what is asked.

\paragraph{Scenario Replays.}
GameCraft-Bench scores the demonstration traces a submitter ships with the project, capped in number and judged on a fixed-length window of each. We replay one fixed policy $\xi_{i,q}$ per canonical scenario $s_{i,q}\in\mathcal{S}_i$. For each build, $A^{\mathrm{build}}$ declares the inputs and their timing, and a tool compiles the declaration into the trace format, rejecting scenarios outside $\mathcal{S}_i$. Every canonical scenario is replayed and scored. Thus, the scenario set is fixed across builds of the same game, while each build supplies its own fixed input policy for those scenarios.

\paragraph{Frame Selection.}
The reference judge sees frames sampled at a fixed interval, at most forty per replay and from a random window when the replay exceeds the cap, so the frames are independent of the inputs that produce the behavior being judged. Our replaying agent records the whole replay together with the executed inputs, the engine's render timestamps, and a context frame before and after the trace; the judge's selection call maps each item of $\rho_i^{\mathrm{vis}}$ to the first render after the relevant input or to an interval between inputs, a deterministic tool resolves these requests to recorded frames and removes duplicates within the budget $M$, and the judge must cite at least one retrieved frame for every item it scores, which the code verifies before accepting the verdict.

\paragraph{Rubric.}
GameCraft rubrics are hand-written per task. $\rho_i^{\mathrm{vis}}$ is generated from the contract under the same schema: each item states a condition and its expected visible outcome with the $0$, $0.5$, and $1$ anchors and stays linked to the rules or invariants it expresses, functional items and presentation items are kept in separate categories rather than sharing one criterion, the number of items per category follows what the contract supports (three to five), and the aggregation rule of each item is chosen by its meaning, max for a capability or a screen that one replay can prove and mean for a quality that must hold across all replays.

\paragraph{Reliability.}
The judge answers through a fixed-schema tool call with one score, one rationale, and the cited frames per item, so a response cannot omit an item or fail to parse; each replay is independently judged three times on the same frames and averaged, and the per-item scores are aggregated across scenarios by the item's rule and combined by the category weights into $F_i^{\mathrm{replay}}$ (judge settings in Section~\ref{app:cost_evaluation}).

\AtoZAppendixBarrier

\subsection{Adaptive Playtest}
\label{app:playtest_protocol}

\subsubsection{Prompt}

The adaptive-playtest pipeline uses three prompted roles: a normal-playtest author, an adversarial-playtest author, and a separate trace judge. The two authors instantiate $A^{\mathrm{play}}$ in Equation~(\ref{eq:playtest_bot}). Both receive the shared interface $\mathrm{API}$ as a fixed system-prompt prefix and the test objective $\omega$ as a JSON brief containing the target rules, dependency edges, and source paths. The normal objective $\omega_i^{\mathrm{pt}}$ asks the bot to complete $g_i$ and collect evidence for $\mathcal{R}_i^{\mathrm{rule}}$; the adversarial objective $\omega_{i,r}^{\mathrm{adv}}$ targets $r\in\mathcal{R}_i^{\mathrm{adv}}$ and its upstream dependencies. Each author returns the bot $b_{i,\omega}$ as a JavaScript module.

The playtest author is confined to player input from $z_i^0$ and must follow the snapshot contract the judge binds to: every action aimed at a rule is followed by a snapshot labeled with the rule id and its trigger, and a snapshot taken before the action carries the suffix \texttt{-before}. The adversarial author establishes $z_{i,r}\models\phi_r$ through a declared scenario entry or a minimal \texttt{ctx.setState} at the root precondition, records it as an \texttt{ASSISTED\_RULE} note, and is instructed to use only player input thereafter, without further state injection. Before-trigger snapshots establish the rule's situation; the required trigger and subsequent effects must be supported by execution evidence.

The trace judge restricts evidence to the snapshots, events, and notes of the trace $\tau_i(\omega,z)$, treats absence of evidence as unverified rather than violated, and answers through a fixed-schema call with one entry per requested rule; three such calls are combined by strict majority. Figure~\ref{fig:playtest-prompts} reproduces the passages of the three prompts that shape the evidence.

\AtoZAppendixInput{figures/appendix/prompt/fig_playtest_prompts}

\subsubsection{Implementation}

\paragraph{Shared Runtime Interface.}
The interface $\mathrm{API}$ is exposed to each JavaScript bot as a \texttt{ctx} object; Figure~\ref{fig:playtest_interface} shows excerpts of two recorded bots. \texttt{ctx.state()} reads \texttt{GameInspector.\allowbreak exportState()} and yields the observation $o_t$, while keyboard and pointer methods issue the player input $x_t$ to the running game. The bot uses observations in conditions and loops, waits for state changes, and records evidence with \texttt{ctx.snapshot()}. Snapshots contain labels, elapsed times, game states, and available events, and together with the input log they form the trace $\tau_i(\omega,z)$ of Equation~\ref{eq:playtest_trace}; \texttt{ctx.watch()} adds selected live-state observations. Adversarial tests additionally use declared scenario entries or \texttt{ctx.setState()} to establish $z_{i,r}$ before observing the specified behavior, whether it follows a player input or elapsed time.

\begingroup
\captionsetup{skip=5pt}
\AtoZAppendixInput{figures/appendix/playtest_interface}
\endgroup

\paragraph{Rule-Level Judgments.}
A separate evaluator checks each rule's conditions, trigger, and expected effects against the recorded trace, issuing a verdict only when the rule's required situation $\phi_r$ is established in the trace. The rule is \emph{satisfied} when the evidence supports its expected behavior and \emph{violated} when the observed outcome contradicts it. A promised value is interpreted by its meaning rather than its spelling, but the operation and number of fields must match: a \texttt{SET} with a different observed value, or an exposed field that contradicts a multi-field effect, constitutes a violation. Code re-checks every cited \texttt{SET} and overrides a satisfied judgment if it finds a contradiction.

When the trace does not establish the required situation or does not support a conclusive outcome, the rule is \emph{unverified}. This status does not count as a violation. Each verdict retains the rule identifier, supporting execution evidence, and whether the situation was reached through normal play ($\tau_i^{\mathrm{pt}}$) or assisted initialization ($\tau_{i,r}^{\mathrm{adv}}$). Source code may help interpret state fields, but does not serve as verdict evidence. Each rule receives three evaluator judgments; the strict majority determines its verdict, and a rule with no majority is treated as unverified.

\paragraph{Combining Results and Aggregation.}
Conclusive normal-playtest judgments form $\mathcal{R}_i^{\mathrm{pt}}$ and are retained, and adversarial tests address $\mathcal{R}_i^{\mathrm{adv}}\subseteq\mathcal{R}_i^{\mathrm{rule}}\setminus\mathcal{R}_i^{\mathrm{pt}}$. Across repeated adversarial attempts on the same rule, an observed violation takes precedence; otherwise, a supported satisfaction is retained. Each rule contributes once to the final result. Let $n_i=|\mathcal{R}_i^{\mathrm{rule}}|$ and let $n_i^{\mathrm{sat}}=\sum_{r\in\mathcal{R}_i^{\mathrm{rule}}} f_{i,r}^{\mathrm{adapt}}$ and $n_i^{\mathrm{vio}}$ count the rules finally judged satisfied and violated. The playtest score is $F_i^{\mathrm{adapt}}=n_i^{\mathrm{sat}}/n_i$, identical to Equation~\ref{eq:playtest_score}, and judgment coverage is $(n_i^{\mathrm{sat}}+n_i^{\mathrm{vio}})/n_i$. Both use the full rule set as the denominator. We report unweighted means of per-game values within each split.

\clearpage

\section{Additional Analysis on Benchmark Results}
\label{app:pairwise_elo}

\paragraph{Comparison Protocol.}
We compare six proprietary coding agents using per-GDD scores from the 100-task benchmark: Claude-Fable-5.1~\citep{anthropic2026claudecode}, Claude-Opus-5~\citep{anthropic2026claudecode}, Claude-Opus-4.8~\citep{anthropic2026claudecode}, GPT-6-Astra~\citep{openai2026gpt6astra}, GPT-5.6-Sol~\citep{openai2026gpt56}, and GPT-5.5~\citep{openai2026gpt55}. For each evaluation axis, a higher score on the same GDD contributes one win, a tie one half, and a lower score zero. Matrix entries average these outcomes over matched GDDs. We use scores before display rounding, including recorded zeros. Source uses graph-weighted fidelity with $\alpha=0.5$. Overall compares the arithmetic mean of the three axis scores for each game before aggregating model-pair outcomes.

\paragraph{Elo Ratings.}
We summarize outcomes against all opponents using Bradley--Terry ratings on an Elo scale~\citep{chiang2024chatbot}. For each split and axis, the expected comparison outcome for model $m$ against model $n$ is
\begin{equation}
    p_{mn}=\frac{1}{1+10^{(r_n-r_m)/400}}.
\end{equation}
Let $w_{mn}$ be wins plus half-credit ties and $n_{mn}$ the number of matched GDDs. We fit all model-pair outcomes jointly by maximizing
\begin{equation}
    \sum_{m<n}\left[
       \left(w_{mn}+\tfrac12\right)\log p_{mn}
       +\left(n_{mn}-w_{mn}+\tfrac12\right)\log(1-p_{mn})
    \right],\qquad
    \frac{1}{M}\sum_{m=1}^{M}r_m=1{,}000.
\end{equation}
Here $M=6$ is the number of compared models. Symmetric half-win and half-loss regularization keeps ratings finite under complete separation and does not alter the empirical heatmap. Each axis is fitted independently; Overall Elo is fitted from Overall comparisons rather than averaged from axis ratings. Pooled ratings combine \emph{Small} and \emph{Big} GDDs. Uncertainty is estimated with 2,000 GDD-level bootstrap resamples, preserving the \emph{Small} and \emph{Big} composition and keeping all comparisons from each sampled GDD together.

\begin{figure}[p]
\centering
\input{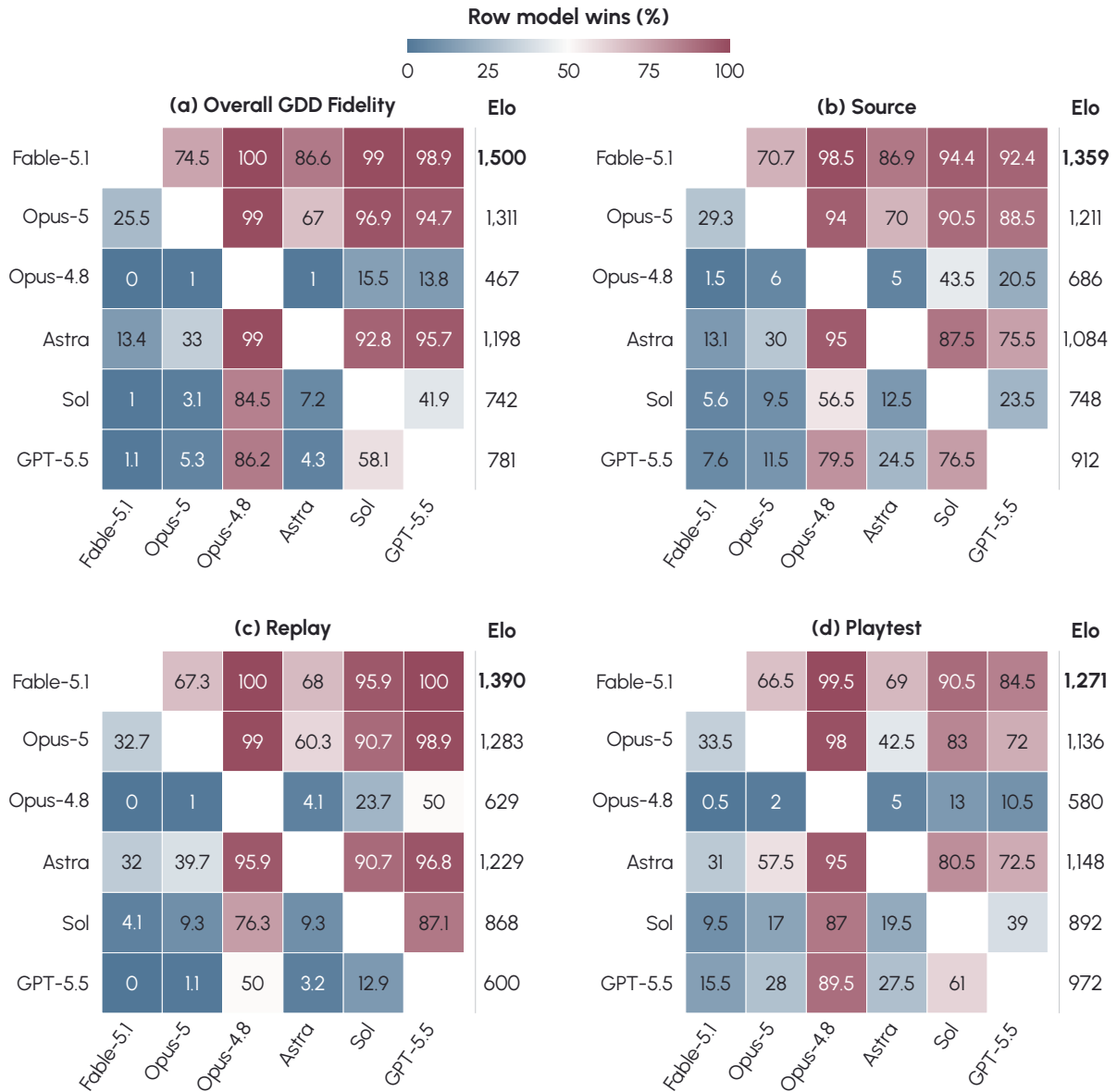}
\caption{\textbf{Pairwise model comparisons and Elo ratings: Pooled GDDs.} Matrix entries give the row model\textquotesingle s win rate against each column model, with ties counted as one half. The rightmost column reports the model\textquotesingle s Elo rating for that panel. Values above 50\% favor the row model; Elo ratings are relative and centered at 1,000 within each panel.}
\label{fig:pairwise_elo_all}
\end{figure}

\begin{figure}[p]
\centering
\input{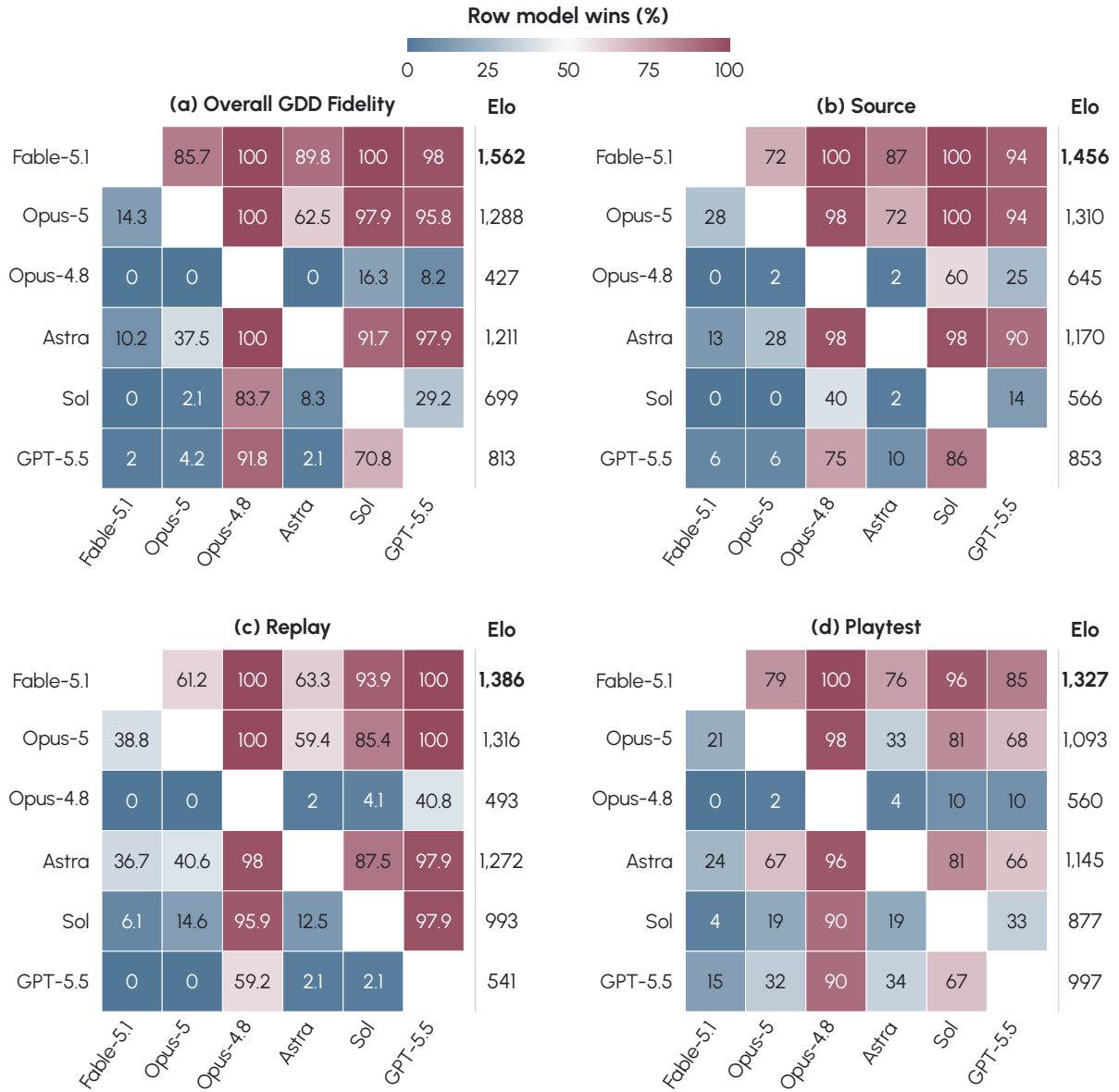}
\caption{\textbf{Pairwise model comparisons and Elo ratings on~\emph{Big} GDDs.} The comparison protocol, relative model order, and win-rate color scale follow Figure~\ref{fig:pairwise_elo_all}. Each panel includes an Elo column alongside the empirical pairwise matrix.}
\label{fig:pairwise_elo_big}
\end{figure}

\paragraph{Results.}
Figure~\ref{fig:pairwise_elo_all} combines the empirical win-rate matrices and model-level Elo summaries for the six proprietary agents. GPT-5.5~\citep{openai2026gpt55} has a win rate of 76.5\% against GPT-5.6-Sol~\citep{openai2026gpt56} in source-code evaluation and 61.0\% in Playtest, compared with 12.9\% in scenario-based replay assessment. Their relative Elo ordering follows the same directions, while Claude-Fable-5.1 has the highest pooled Elo on all four axes. Figures~\ref{fig:pairwise_elo_small} and~\ref{fig:pairwise_elo_big} provide split-specific comparisons. Mean fidelity measures the level of specification fulfillment. Specifically, win rates and Elo summarize comparative performance across GDDs, with Elo aggregating outcomes against all opponents.

\begin{figure}[p]
\centering
\input{figures/appendix/elo_score/pairwise_elo_small}
\caption{\textbf{Pairwise model comparisons and Elo ratings on~\emph{Small} GDDs.} The comparison protocol, relative model order, and win-rate color scale follow Figure~\ref{fig:pairwise_elo_all}. Each panel includes an Elo column alongside the empirical pairwise matrix.}
\label{fig:pairwise_elo_small}
\end{figure}

\clearpage

\section{Ablations on Evaluation}
\label{app:ablation_for_eval}

This section examines choices in the evaluation that the method does not force: whether a judge without a contract keeps a stable requirement set, what dependency weighting changes in the source-code score, how the replay judge's model and reasoning effort shift its scores, and what adversarial initialization adds over repeated normal play. Each ablation modifies a single factor while keeping other inputs constant, such as builds, scenario replays, rubrics, or traces, and results are read against the self-disagreement of the same judge.

\subsection{Source-Code Evaluation}
\label{app:ablation-source-code-eval}

We examine the source-code evaluation in Section~\ref{sec:source_eval} through repeated judgments, changes in judge configuration, and an ablation of dependency-weighted scoring.

\subsubsection{Requirement-Set Consistency}
The naive GDD-based judge identifies its own requirements from the GDD and
source project, while the contract-based judge scores a fixed set of rules
and invariants. Only the naive judge reconstructs its targets on each run.
We therefore test whether repeated judgments of an unchanged build use a
consistent set of requirements.

\begin{figure*}[!htbp]
    \centering
    \colorlet{A2ZReqSmall}{A2ZRed!60!white}
\colorlet{A2ZReqLarge}{A2ZRed!78!black}
\begin{tikzpicture}
\begin{axis}[a2zfig, enlarge x limits=0.01, height=0.5\linewidth, xmin=0, xmax=51, xtick=\empty,
  ymode=log, ymin=1.5, ymax=3000, ytick={2,5,10,20,50,100,200,500,1000}, yticklabels={2,5,10,20,50,100,200,500,1000}, minor ytick={},
  ylabel={\textbf{Number of Requirements}}, xlabel={\textbf{Games, Sorted by Rule Counts}},
  every tick label/.append style={font=\rmfamily\scriptsize},
  label style={font=\rmfamily\footnotesize},
  legend style={at={(0.01,0.99)}, anchor=north west, legend columns=2, font=\rmfamily\scriptsize,
                /tikz/every even column/.append style={column sep=6pt}}, legend cell align=left]
\addlegendimage{line width=1.4pt, A2ZReqSmall}\addlegendentry{Naive,~\emph{Small} (5 runs)}
\addlegendimage{line width=1.4pt, A2ZReqLarge}\addlegendentry{Naive,~\emph{Big} (5 runs)}
\addlegendimage{line width=1.3pt, A2ZReqSmall, opacity=0.6}\addlegendentry{Ours, \emph{Small} (5 runs)}
\addlegendimage{line width=1.3pt, A2ZReqLarge, opacity=0.6}\addlegendentry{Ours, \emph{Big} (5 runs)}
\addplot[A2ZReqSmall, forget plot, only marks, mark=*, mark size=0pt, error bars/.cd, y dir=both, y explicit, error bar style={line width=1.4pt, A2ZReqSmall}, error mark=none]
  table[x expr={\thisrow{idx}-0.2}, y=small_naive_lo, y error plus expr={\thisrow{small_naive_hi}-\thisrow{small_naive_lo}}, y error minus expr={0}] {data/appendix/req_count_naive_vs_contract5.dat};
\addplot[A2ZReqLarge, forget plot, only marks, mark=*, mark size=0pt, error bars/.cd, y dir=both, y explicit, error bar style={line width=1.4pt, A2ZReqLarge}, error mark=none]
  table[x expr={\thisrow{idx}+0.2}, y=large_naive_lo, y error plus expr={\thisrow{large_naive_hi}-\thisrow{large_naive_lo}}, y error minus expr={0}] {data/appendix/req_count_naive_vs_contract5.dat};
\addplot[forget plot, only marks, mark=*, mark size=0.9pt, A2ZFigText] table[x expr={\thisrow{idx}-0.2}, y=small_naive_mean] {data/appendix/req_count_naive_vs_contract5.dat};
\addplot[forget plot, only marks, mark=*, mark size=0.9pt, A2ZFigText] table[x expr={\thisrow{idx}+0.2}, y=large_naive_mean] {data/appendix/req_count_naive_vs_contract5.dat};
\addplot[forget plot, only marks, mark=*, mark size=0pt, error bars/.cd, y dir=both, y explicit, error bar style={line width=1.3pt, A2ZReqSmall, opacity=0.6}, error mark=none]
  table[x expr={\thisrow{idx}-0.2}, y=small_contract_lo, y error plus expr={\thisrow{small_contract_hi}-\thisrow{small_contract_lo}}, y error minus expr={0}] {data/appendix/req_count_naive_vs_contract5.dat};
\addplot[forget plot, only marks, mark=*, mark size=0pt, error bars/.cd, y dir=both, y explicit, error bar style={line width=1.3pt, A2ZReqLarge, opacity=0.6}, error mark=none]
  table[x expr={\thisrow{idx}+0.2}, y=large_contract_lo, y error plus expr={\thisrow{large_contract_hi}-\thisrow{large_contract_lo}}, y error minus expr={0}] {data/appendix/req_count_naive_vs_contract5.dat};
\addplot[forget plot, only marks, mark=diamond*, mark size=1.5pt, draw=A2ZReqSmall, fill=white] table[x expr={\thisrow{idx}-0.2}, y=small_contract_mean] {data/appendix/req_count_naive_vs_contract5.dat};
\addplot[forget plot, only marks, mark=diamond*, mark size=1.5pt, draw=A2ZReqLarge, fill=white] table[x expr={\thisrow{idx}+0.2}, y=large_contract_mean] {data/appendix/req_count_naive_vs_contract5.dat};
\end{axis}
\end{tikzpicture}\\[2pt]
    \makebox[\textwidth][c]{
    \centering
    \colorlet{A2ZReqSmall}{A2ZRed!60!white}
\colorlet{A2ZReqLarge}{A2ZRed!78!black}
\begin{tikzpicture}[trim axis left, trim axis right, /pgf/number format/assume math mode=true]
\begin{axis}[a2zfig, width=0.58\textwidth, height=0.3\textwidth, ybar, bar width=9pt, ymin=0, ymax=24, enlarge x limits=0.07,
  symbolic x coords={0--10,10--20,20--30,30--40,40--50,50--60,60--70,70--80}, xtick=data,
  every tick label/.append style={font=\rmfamily\scriptsize},
  ylabel style={font=\rmfamily\footnotesize},
  xlabel style={font=\rmfamily\bfseries\footnotesize, align=center},
  title style={font=\rmfamily\bfseries\footnotesize},
  nodes near coords style={font=\rmfamily\bfseries\scriptsize},
  ylabel={\textbf{Number of games}},
  xlabel={Requirements of First Run Found Again in Another\\Run of the Same GDD (\%, Jaccard $\ge 0.5$)},
  nodes near coords, nodes near coords align={vertical}, point meta=rawy,
  ymajorgrids, grid style={A2ZFigAxis!66!white,line width=0.35pt},
  axis lines*=left, axis line style={A2ZFigAxis,line width=0.5pt}, tick style={A2ZFigAxis,line width=0.4pt},
  legend style={at={(0.99,0.99)}, anchor=north east, font=\rmfamily\bfseries\scriptsize, draw=none, fill=none}, legend cell align=left,
  every node near coord/.append style={/tikz/shade=false,/tikz/fill=none,/tikz/draw=none},
  title={Cross-run requirement alignment (median:~\emph{Small} 38\%,~\emph{Big} 31\%)}]
\addplot[shade,bottom color=A2ZReqSmall!92!black,top color=A2ZReqSmall!72!white,draw=A2ZFigAxis!70!black,line width=0.22pt, rounded corners=1.2pt]
  coordinates {(0--10,0) (10--20,2) (20--30,9) (30--40,16) (40--50,9) (50--60,9) (60--70,3) (70--80,2) (0--10,0)};
\addplot[shade,bottom color=A2ZReqLarge!92!black,top color=A2ZReqLarge!72!white,draw=A2ZFigAxis!70!black,line width=0.22pt, rounded corners=1.2pt]
  coordinates {(0--10,0) (10--20,4) (20--30,17) (30--40,21) (40--50,6) (50--60,2) (60--70,0) (70--80,0)};
\legend{\emph{Small} GDDs, \emph{Big} GDDs}
\end{axis}
\end{tikzpicture}}
    \caption{\textbf{Requirement consistency across repeated source-code judgments.} The naive judge evaluates each of 100 unchanged builds five times. Top: the minimum, maximum, and mean number of extracted requirements per game, with rule counts shown for comparison. Bottom: the distribution of the fraction of one run's requirements matched in another run at a Jaccard threshold of 0.5.}
    \label{fig:direct-drift}
\end{figure*}
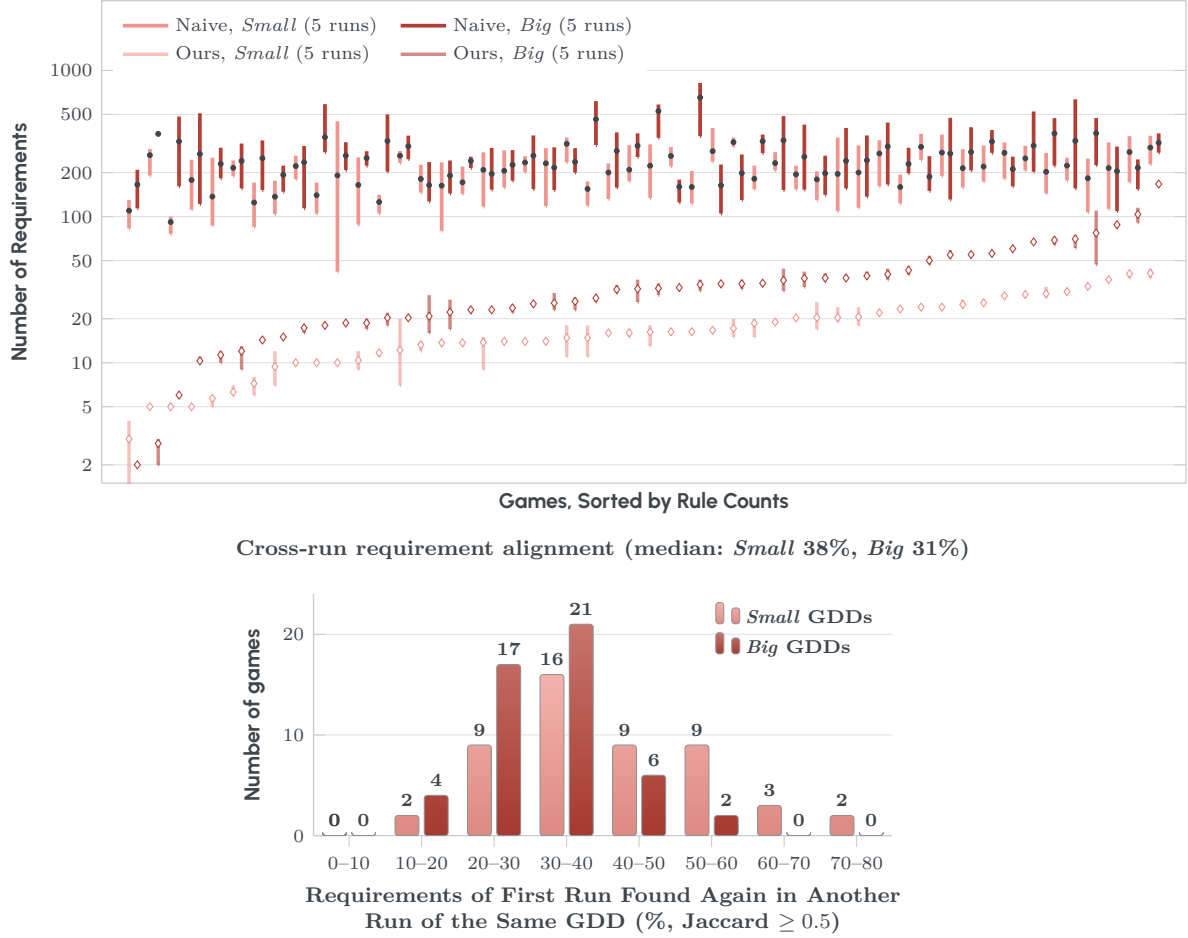

We run the naive judge five times per unchanged build for all 100 GDDs,
using GPT-5.6-Sol at high reasoning effort. We also repeat dependency-aware contract construction five times for each GDD. After the consistency checks pass, the first generated contract is frozen for benchmark evaluation across builds and revision rounds, as described in Section~\ref{app:gdd-contract-construction}. Scores vary little: the median
gap between a build's highest and lowest scores is 0.05. However, the extracted
requirements vary substantially: for the median GDD, item counts vary by
60\% of their mean, only 35\% of items match across runs at a Jaccard
threshold of 0.5, and 9\% appear in all five runs. This instability is
greater for \emph{Big} GDDs: the median cross-run match is 31\%, with 7\%
of items appearing in all five runs, compared with 38\% and 14\% for
\emph{Small} GDDs. Figure~\ref{fig:direct-drift} shows requirement-count
variation and cross-run item alignment. Thus, a stable score can mask shifting evaluation criteria, undermining requirement-level comparisons. Freezing the contract keeps targets fixed
across agents and revision rounds; Section~\ref{app:contract_stability}
examines how consistently the contract itself can be constructed.

On the same build $g_i$, the naive score correlates more strongly with the contract's rule component $F_i^{\mathrm{rule}}$ than with its invariant component $F_i^{\mathrm{inv}}$: Pearson correlations of 0.54 versus 0.16 for \emph{Small} GDDs, and 0.87 versus 0.23 for \emph{Big} GDDs. This limited correspondence with invariant pass rates motivates checking invariants explicitly alongside rules in the source-code measure, as in Equation~\ref{eq:source_score}.

\AtoZAppendixBarrier

\begin{figure}[t]
\centering
\input{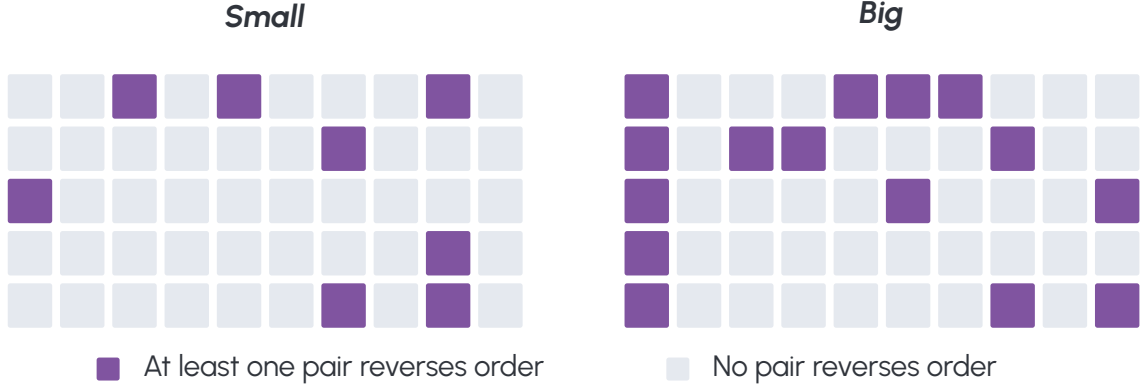}
\caption{\textbf{Dependency weighting changes game-level comparisons.} Each cell represents one GDD. Purple cells indicate a strict reversal in the source-code score ordering of at least one pair of the six proprietary agents between uniform weighting ($\alpha=1$) and downstream-reach weighting ($\alpha=0$), with rule judgments and invariant scores fixed. Such reversals occur on 23 of the 100 GDDs.}
\label{fig:alpha_game_comparisons}
\end{figure}

\subsubsection{Ablation On Dependency-Weighted Scoring}
\label{app:ablation-dependency-weight}

Table~\ref{tab:dependency_weighting} measures the sensitivity of the full source-code score $F_i^{\mathrm{src}}$ to dependency weighting. We hold the rule scores and invariant judgments fixed and compare $\alpha=0.5$ and $\alpha=0.1$ with uniform weighting ($\alpha=1$), reporting the mean and maximum absolute score changes. The analysis below separately examines changes in the rule component $F_i^{\mathrm{rule}}$ for builds with and without confirmed high-reach failures. Weighting rules by their downstream reach, as defined in Equation~\ref{eq:dependency_weights}, gives greater influence to mechanics on which other behaviors depend. We compare the dependency-weighted rule score $F_i^{\mathrm{rule}}$ at $\alpha=0.5$ against its unweighted counterpart at $\alpha=1$, and denote the difference by $\Delta_\alpha=F_i^{\mathrm{rule}}(\alpha)-F_i^{\mathrm{rule}}(1)$. A negative gap means that weighting lowers the score. This is expected when rules with greater downstream reach receive relatively low implementation scores; conversely, the score can increase when failures concentrate on lower-weight leaf rules.

\begin{figure}[!htbp]
    \centering
    \colorlet{A2ZReqSmall}{A2ZRed!60!white}
\colorlet{A2ZReqLarge}{A2ZRed!78!black}
\definecolor{A2ZCtrlSmall}{HTML}{B9BFC5}
\definecolor{A2ZCtrlLarge}{HTML}{7E858C}
\begin{tikzpicture}
\begin{axis}[a2zfig, width=\linewidth, height=0.36\linewidth, ybar=0pt, bar width=7pt, bar shift=0pt,
  xmin=0, xmax=37, enlarge x limits=false, ymin=-60, ymax=64,
  xtick={9.5,28.5},
  xticklabels={{root-defective: 11 of 18 below zero},{no root violation: 6 of 16 below zero}},
  xticklabel style={font=\rmfamily\bfseries\scriptsize},
  yticklabel style={font=\rmfamily\scriptsize},
  ylabel style={font=\rmfamily\footnotesize},
  ylabel={$\Delta_{\alpha}\times10^{-3}$},
  scaled y ticks=false,
  legend style={at={(0.01,0.99)}, anchor=north west, legend columns=2,
                font=\rmfamily\scriptsize, draw=none, fill=none,
                /tikz/every even column/.append style={column sep=5pt}},
  legend image code/.code={\draw[#1] (0cm,-0.045cm) rectangle (0.16cm,0.09cm);},
  legend cell align=left]
\addlegendimage{area legend, shade,bottom color=A2ZReqSmall!92!black,top color=A2ZReqSmall!72!white,draw=A2ZFigAxis!70!black,line width=0.22pt}\addlegendentry{\emph{Small}, Root-Defective}
\addlegendimage{area legend, shade,bottom color=A2ZReqLarge!92!black,top color=A2ZReqLarge!72!white,draw=A2ZFigAxis!70!black,line width=0.22pt}\addlegendentry{\emph{Big}, Root-Defective}
\addlegendimage{area legend, shade,bottom color=A2ZCtrlSmall!92!black,top color=A2ZCtrlSmall!72!white,draw=A2ZFigAxis!70!black,line width=0.22pt}\addlegendentry{\emph{Small}, No Root Violation}
\addlegendimage{area legend, shade,bottom color=A2ZCtrlLarge!92!black,top color=A2ZCtrlLarge!72!white,draw=A2ZFigAxis!70!black,line width=0.22pt}\addlegendentry{\emph{Big}, No Root Violation}
\addplot[forget plot, ybar, shade,bottom color=A2ZReqSmall!92!black,top color=A2ZReqSmall!72!white,draw=A2ZFigAxis!70!black,line width=0.22pt] table[x=x, y=delta_milli, restrict expr to domain={\thisrow{group}*\thisrow{small}}{1:1}] {data/appendix/weight_delta_typeA2.dat};
\addplot[forget plot, ybar, shade,bottom color=A2ZReqLarge!92!black,top color=A2ZReqLarge!72!white,draw=A2ZFigAxis!70!black,line width=0.22pt] table[x=x, y=delta_milli, restrict expr to domain={\thisrow{group}*(1-\thisrow{small})}{1:1}] {data/appendix/weight_delta_typeA2.dat};
\addplot[forget plot, ybar, shade,bottom color=A2ZCtrlSmall!92!black,top color=A2ZCtrlSmall!72!white,draw=A2ZFigAxis!70!black,line width=0.22pt] table[x=x, y=delta_milli, restrict expr to domain={(1-\thisrow{group})*\thisrow{small}}{1:1}] {data/appendix/weight_delta_typeA2.dat};
\addplot[forget plot, ybar, shade,bottom color=A2ZCtrlLarge!92!black,top color=A2ZCtrlLarge!72!white,draw=A2ZFigAxis!70!black,line width=0.22pt] table[x=x, y=delta_milli, restrict expr to domain={(1-\thisrow{group})*(1-\thisrow{small})}{1:1}] {data/appendix/weight_delta_typeA2.dat};
\addplot[forget plot, only marks, mark=none, draw=none, point meta=explicit,
  nodes near coords, nodes near coords align={anchor=center},
  nodes near coords style={font=\rmfamily\tiny, text=A2ZFigText, inner sep=0pt,
    /pgf/number format/fixed, /pgf/number format/precision=0, /pgf/number format/assume math mode=true}]
  table[x=x, y expr={\thisrow{delta_milli} + ((\thisrow{delta_milli}>=0) ? 3 : -3)}, meta=delta_milli]
  {data/appendix/weight_delta_typeA2.dat};
\draw[A2ZFigText] (axis cs:0,0) -- (axis cs:37,0);
\draw[A2ZFigAxis, dashed] (axis cs:19.5,-58) -- (axis cs:19.5,62);
\end{axis}
\end{tikzpicture}
    \caption{\textbf{Effect of dependency weighting on rule scores.} Each bar is $\Delta_\alpha=F_i^{\mathrm{rule}}(\alpha)-F_i^{\mathrm{rule}}(1)$ at $\alpha=0.5$. Left: root-defective games meeting the joint source-code and playtest criteria. Right: comparison games without an observed playtest violation among their assessed root rules. Negative values indicate lower weighted scores; the sign is unchanged at $\alpha=0$.}
    \label{fig:weight-delta}
\end{figure}
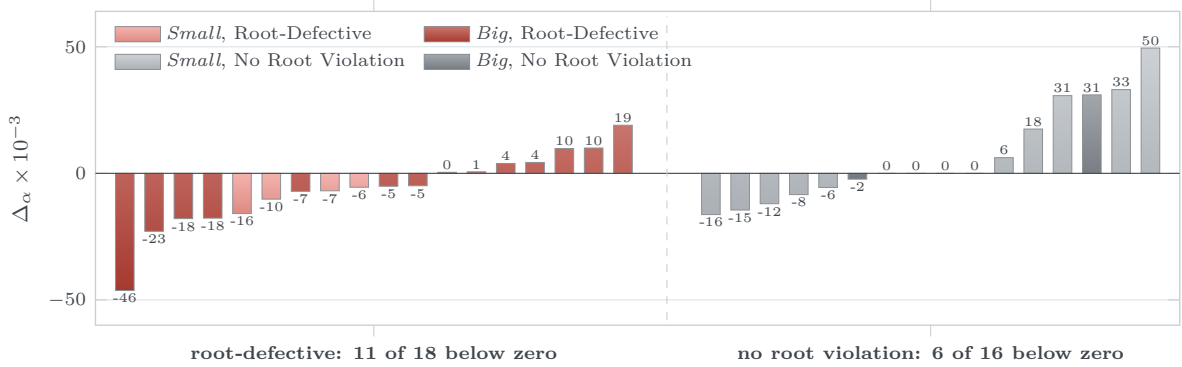

To examine this effect, we classify a game as \emph{root-defective} when a core root rule, whose downstream reach is at least half the maximum in the game's contract, receives a score of at most 0.5 under both source-code judge settings and is violated in the adaptive playtest. The comparison group contains games with no observed playtest violation among their assessed root rules. Both groups include \emph{Small} and \emph{Big} games. As shown in Figure~\ref{fig:weight-delta}, negative gaps occur in 11 of 18 root-defective games, compared with 6 of 16 comparison games. Thus, dependency weighting lowers the score more often in games with confirmed foundational failures. Positive gaps in the root-defective group arise primarily when failures are even more prevalent among leaf rules, which pull down the unweighted average more strongly. The sign consequently reflects the distribution of scores across the contract, rather than the presence of a root failure alone. The gaps are small relative to judge variability, so we interpret the group difference as a tendency in the observed scores. This tendency is consistent with the intended role of dependency weighting: giving greater influence to rules with broader downstream effects.

Furthermore, we examine how dependency weighting affects comparisons on individual GDDs. For the six proprietary agents in Appendix~\ref{app:pairwise_elo}, we hold the rule judgments and invariant scores fixed and compare source-code scores under uniform weighting ($\alpha=1$) and downstream-reach weighting ($\alpha=0$). As shown in Figure~\ref{fig:alpha_game_comparisons}, the ordering of at least one agent pair reverses on 23 of the 100 GDDs. The same rule-level judgments can therefore favor different agents when greater influence is assigned to rules on which other requirements depend.

\AtoZAppendixBarrier

\subsection {Scenario-Based Replay Assessment}
\label{app:ablation-mllm}

Table~\ref{tab:vlm-effort} evaluates GPT-5.6-Sol and GPT-5.6-Luna at four reasoning efforts on the same six builds, scenario replays, and rubrics, against the original sol extra-high judgments; item agreement is the Pearson correlation between the per-item scores of $\rho_i^{\mathrm{vis}}$ and those judgments, with a re-run agreement of about 0.86. At extra-high effort, substituting luna for sol changes the mean $F_i^{\mathrm{replay}}$ by $-0.005$ at an agreement of 0.85. Lowering the effort increases mean scores relative to the reference, though not equally: sol rises by $+0.03$ to $+0.04$ at each lower level with agreement between 0.88 and 0.89, whereas luna stays at $+0.006$ at high and rises to $+0.035$ at medium and $+0.081$ at low, with agreement falling to 0.79, mostly on the visual and art items. We therefore use luna at high effort: its bias is smaller than sol's at any effort below extra-high, its agreement is close to the re-run agreement, and a judgment takes 41 seconds at about \$0.007 against 78 seconds and \$0.048 for the reference, a 6.7$\times$ reduction that brings the replay score of a game (three judgments over 13--24 scenario replays) from about \$2--3.5 to \$0.3--0.5.

\begin{table}[!htbp]
    \centering
    \caption{\textbf{Agent-as-a-judge on six builds with identical scenario replays and rubric.} Item $r$ is the Pearson correlation of rubric-item scores with the original sol extra-high judgments of the same builds (re-run agreement $\approx0.86$); $\Delta$ is the mean reward difference from them. Cost is per judge call at list prices. sol = GPT-5.6-Sol, luna = GPT-5.6-Luna.}
    \label{tab:vlm-effort}
    \small
    \setlength{\tabcolsep}{7pt}
    \renewcommand{\arraystretch}{1.15}
    \begin{tabular}{@{}l c c c c c@{}}
        \toprule
        \textbf{Judge, effort} & \textbf{Item $r$} & \textbf{Mean reward} & \textbf{$\Delta$ vs.\ ref.} & \textbf{s / call} & \textbf{\$ / call} \\
        \midrule
        sol, xhigh (reference) & -- & 0.639 & -- & 78 & 0.048 \\
        sol, high & 0.88 & 0.667 & $+0.027$ & 71 & 0.034 \\
        sol, medium & 0.88 & 0.669 & $+0.030$ & 63 & 0.034 \\
        sol, low & 0.89 & 0.680 & $+0.041$ & 52 & 0.045 \\
        \midrule
        luna, xhigh & 0.85 & 0.634 & $-0.005$ & 63 & 0.0074 \\
        luna, high & 0.85 & 0.645 & $+0.006$ & 41 & 0.0072 \\
        luna, medium & 0.83 & 0.674 & $+0.035$ & 27 & 0.0070 \\
        luna, low & 0.79 & 0.720 & $+0.081$ & 21 & 0.0070 \\
        \bottomrule
    \end{tabular}
\end{table}
\AtoZAppendixBarrier

\subsection{Adaptive Playtest}

\label{app:adversarial_playtesting}

\paragraph{Evidence Across Coding Agents.}
Figure~\ref{fig:playtest_coverage_agents} and Table~\ref{tab:playtest_coverage_agents} decompose the playtest results reported in Table~\ref{tab:main_results}. For each coding agent and split, we use the same saved runs and score-reporting cohort as the main comparison, holding the builds and requirement sets fixed across the normal and adversarial phases. The analysis uses the same enumerated contracts and playtest pipeline across agents, with GPT-5.6-Luna as the bot and judge, one normal-play round, and up to two adversarial attempts per unresolved target.

\AtoZAppendixInput{figures/exp_playtest_coverage}

For build $i$, let $n_i=|\mathcal{R}_i^{\mathrm{rule}}|$, and let $S_i^{\mathrm{N}}$ and $V_i^{\mathrm{N}}$ count requirements satisfied and violated after normal play. We join phase records by requirement identity and count $\Delta S_i$ and $\Delta V_i$ only when a previously unverified requirement becomes satisfied or violated. All conclusive normal verdicts are preserved in the paired records, so each requirement contributes once. The score and coverage decompose as
\begin{equation}
    F_i^{\mathrm{adapt}}=\frac{S_i^{\mathrm{N}}+\Delta S_i}{n_i},\qquad
    C_i^{\mathrm{N}}=\frac{S_i^{\mathrm{N}}+V_i^{\mathrm{N}}}{n_i},\qquad
    C_i^{\mathrm{N+A}}=C_i^{\mathrm{N}}+\frac{\Delta S_i+\Delta V_i}{n_i}.
    \label{eq:coverage_decomposition}
\end{equation}
Every rate retains the full per-game requirement set in its denominator. We first average game-level fractions within each agent and split; All equally averages \emph{Small} and \emph{Big}, and aggregate means then equally weight the nine agents. Added satisfaction and violation are percentage-point contributions to the combined result. Judgment coverage measures the extent of conclusive execution evidence obtained under the fixed playtest protocol, including initialized adversarial scenarios.

Mean judgment coverage increases from 21.7\% after normal play to 74.4\% after adversarial testing, adding 41.1 percentage points of satisfaction and 11.6 points of violation evidence. Every agent gains both types of evidence, with coverage increases of 40.2--59.0 percentage points. The increase is larger on \emph{Big}, where coverage rises from 8.2\% to 68.3\%, compared with 35.3\% to 80.5\% on \emph{Small}. Additional violation evidence also increases from 7.3 points on \emph{Small} to 15.8 points on \emph{Big}. These results characterize the contribution of adversarial testing to the evidence collected for the main benchmark scores.

The recorded judgments can have different compositions at similar coverage. On \emph{Big}, GPT-6-Astra and GPT-5.6-Sol reach coverage of 79.0\% and 77.2\%, respectively, but their reported Playtest scores are 67.4 and 57.9. Adversarial testing adds more satisfied requirements for GPT-6-Astra (58.7 vs.\ 48.9 points), while identifying more violations for GPT-5.6-Sol (18.6 vs.\ 11.4 points). The decomposition separates confirmed satisfaction from observed violations within the collected execution evidence.

A similar distinction arises when normal coverage is close. On \emph{Big}, Claude-Fable-5.1 and GPT-5.5 begin at 12.7\% and 12.0\% coverage, but adversarial tests add 62.5 and 50.3 points of satisfaction and 10.5 and 16.9 points of violation, respectively. Their final Playtest scores are 74.7 and 61.7, matching Table~\ref{tab:main_results}. Reporting both added outcomes shows how targeted execution evidence separates downstream successes from failures beyond the behavior observed during normal play.

\begin{table}[!tbp]
\centering
\caption{\textbf{Normal and adversarial playtest decomposition across agents.} Each split uses the same score-reporting cohort as Table~\ref{tab:main_results}; All is the mean of \emph{Small} and \emph{Big}. Sat., Viol., and Cov. denote satisfied, violated, and conclusively judged requirements, respectively. Adversarial gain reports additional judgments for previously unverified requirements in percentage points (pp). Combined Score reproduces the \emph{Small} and \emph{Big} Playtest scores in Table~\ref{tab:main_results} using the same archived per-game scores; phase contributions and coverage use exact verdict counts. All values are reported to one decimal place.}
\label{tab:playtest_coverage_agents}
\begingroup
\AtoZFigFont\fontsize{9.2}{11}\selectfont
\setlength{\tabcolsep}{4pt}
\renewcommand{\arraystretch}{1.12}
\setlength{\aboverulesep}{3pt}
\setlength{\belowrulesep}{3pt}
\begin{tabularx}{\linewidth}{@{}l*{3}{>{\raggedleft\arraybackslash}X}@{\hspace{12pt}}*{2}{>{\raggedleft\arraybackslash}X}@{\hspace{12pt}}*{2}{>{\raggedleft\arraybackslash}X}@{}}
\toprule
\multirow{2}{*}{\textbf{Coding agent}} & \multicolumn{3}{c}{\textbf{Normal (\%)}} & \multicolumn{2}{c}{\textbf{Adversarial gain (pp)}} & \multicolumn{2}{c}{\textbf{Combined (\%)}} \\
\cmidrule(lr){2-4}\cmidrule(lr){5-6}\cmidrule(l){7-8}
& Sat. & Viol. & Cov. & Sat. & Viol. & \textbf{Score} & Cov. \\
\midrule
\rowcolor{A2ZRed!7}
\multicolumn{8}{l}{\strut\bfseries All} \\
\addlinespace[3pt]
\ModelWithIcon{claude}{Claude-Fable-5.1} & 28.6 & 0.4 & 29.0 & 51.4 & 7.6 & \textbf{80.0} & 88.0 \\
\ModelWithIcon{claude}{Claude-Opus-5} & 26.7 & 0.4 & 27.1 & 49.0 & 9.0 & \textbf{75.7} & 85.1 \\
\ModelWithIcon{claude}{Claude-Opus-4.8} & 19.2 & 0.6 & 19.7 & 38.2 & 13.6 & \textbf{57.4} & 71.5 \\
\ModelWithIcon{gpt}{GPT-6-Astra} & 25.8 & 0.5 & 26.3 & 48.7 & 8.5 & \textbf{74.5} & 83.4 \\
\ModelWithIcon{gpt}{GPT-5.6-Sol} & 21.1 & 0.6 & 21.7 & 45.8 & 12.5 & \textbf{67.0} & 80.1 \\
\ModelWithIcon{gpt}{GPT-5.5} & 25.3 & 0.6 & 25.9 & 44.3 & 11.5 & \textbf{69.6} & 81.7 \\
\ModelWithIcon{glm}{GLM-5.3} & 13.6 & 0.6 & 14.2 & 39.0 & 12.8 & \textbf{52.6} & 66.0 \\
\ModelWithIcon{deepseek}{DeepSeek-V4-Pro} & 17.5 & 1.0 & 18.5 & 30.0 & 12.3 & \textbf{47.5} & 60.8 \\
\ModelWithIcon{kimi}{Kimi-K2.7} & 12.0 & 1.1 & 13.1 & 23.9 & 16.2 & \textbf{36.0} & 53.3 \\
\addlinespace[5pt]
\rowcolor{A2ZRed!7}
\multicolumn{8}{l}{\strut\bfseries \emph{Small}} \\
\addlinespace[3pt]
\ModelWithIcon{claude}{Claude-Fable-5.1} & 44.9 & 0.4 & 45.4 & 40.3 & 4.8 & \textbf{85.2} & 90.4 \\
\ModelWithIcon{claude}{Claude-Opus-5} & 41.8 & 0.5 & 42.3 & 41.8 & 5.5 & \textbf{83.6} & 89.7 \\
\ModelWithIcon{claude}{Claude-Opus-4.8} & 32.0 & 0.8 & 32.8 & 36.7 & 8.3 & \textbf{68.7} & 77.9 \\
\ModelWithIcon{gpt}{GPT-6-Astra} & 42.9 & 0.8 & 43.7 & 38.6 & 5.5 & \textbf{81.6} & 87.9 \\
\ModelWithIcon{gpt}{GPT-5.6-Sol} & 33.2 & 0.5 & 33.7 & 42.8 & 6.4 & \textbf{76.0} & 82.9 \\
\ModelWithIcon{gpt}{GPT-5.5} & 39.2 & 0.6 & 39.8 & 38.3 & 6.1 & \textbf{77.5} & 84.2 \\
\ModelWithIcon{glm}{GLM-5.3} & 23.4 & 0.7 & 24.2 & 41.5 & 8.4 & \textbf{64.9} & 74.0 \\
\ModelWithIcon{deepseek}{DeepSeek-V4-Pro} & 31.1 & 1.5 & 32.6 & 36.9 & 8.5 & \textbf{68.0} & 78.0 \\
\ModelWithIcon{kimi}{Kimi-K2.7} & 21.4 & 1.8 & 23.1 & 24.4 & 12.5 & \textbf{45.7} & 60.0 \\
\addlinespace[5pt]
\rowcolor{A2ZRed!7}
\multicolumn{8}{l}{\strut\bfseries \emph{Big}} \\
\addlinespace[3pt]
\ModelWithIcon{claude}{Claude-Fable-5.1} & 12.3 & 0.4 & 12.7 & 62.5 & 10.5 & \textbf{74.7} & 85.6 \\
\ModelWithIcon{claude}{Claude-Opus-5} & 11.6 & 0.3 & 11.9 & 56.1 & 12.5 & \textbf{67.7} & 80.5 \\
\ModelWithIcon{claude}{Claude-Opus-4.8} & 6.3 & 0.3 & 6.6 & 39.6 & 18.9 & \textbf{46.0} & 65.1 \\
\ModelWithIcon{gpt}{GPT-6-Astra} & 8.7 & 0.2 & 8.9 & 58.7 & 11.4 & \textbf{67.4} & 79.0 \\
\ModelWithIcon{gpt}{GPT-5.6-Sol} & 9.0 & 0.7 & 9.7 & 48.9 & 18.6 & \textbf{57.9} & 77.2 \\
\ModelWithIcon{gpt}{GPT-5.5} & 11.5 & 0.5 & 12.0 & 50.3 & 16.9 & \textbf{61.7} & 79.2 \\
\ModelWithIcon{glm}{GLM-5.3} & 3.8 & 0.5 & 4.2 & 36.6 & 17.2 & \textbf{40.4} & 58.1 \\
\ModelWithIcon{deepseek}{DeepSeek-V4-Pro} & 3.8 & 0.5 & 4.3 & 23.2 & 16.1 & \textbf{27.0} & 43.6 \\
\ModelWithIcon{kimi}{Kimi-K2.7} & 2.7 & 0.5 & 3.1 & 23.5 & 19.9 & \textbf{26.2} & 46.6 \\
\bottomrule
\end{tabularx}
\endgroup
\end{table}

\paragraph{Controlled Comparison.}
To separate prerequisite initialization from merely running another test, we compare the modes on 86 dependency-linked rules from 25 \emph{Small} and 25 \emph{Big} games, with the build, contract, \textsc{Test Policy}, interface $\mathrm{API}$, and budgets held fixed and two seeds per target; initialization supplies $z_{i,r}\models\phi_r$, never target outcomes.

Table~\ref{tab:controlled_playtest_comparison} reports a first normal run and a second pass that re-tests only its unverified targets, once normal and once adversarial, with the adversarial mode alone as reference. Each episode takes the majority of three GPT-5.6-Luna judgments at high effort, and coverage is averaged over targets and seeds, then games and splits.
A second normal test increases coverage from 40.5\% to 41.0\%, whereas an adversarial second test increases it to 48.5\%. The combined setting also exceeds the 45.0\% coverage of adversarial testing alone, supporting their complementary use: normal play provides evidence along ordinary gameplay paths, while initialization enables additional downstream checks.

\begin{table}[!htbp]
    \centering
    \caption{\textbf{Controlled playtest comparison.} Mean judgment coverage (\%) on 86 target rules from 25 \emph{Small} and 25 \emph{Big} games. A second pass keeps the first pass's conclusive verdicts and re-tests only the targets it left unverified.}
    \label{tab:controlled_playtest_comparison}
    \small
    \setlength{\tabcolsep}{10pt}
    \renewcommand{\arraystretch}{1.08}
    \begin{tabular}{@{}llrr@{}}
        \toprule
        \textbf{First pass} & \textbf{Second pass} & \textbf{Coverage} & $\Delta$ \\
        \midrule
        Normal      & --          & 40.5 & -- \\
        Normal      & Normal      & 41.0 & $+0.5$ \\
        Normal      & Adversarial & 48.5 & $+8.0$ \\
        \midrule
        Adversarial & --          & 45.0 & -- \\
        \bottomrule
    \end{tabular}
\end{table}
\AtoZAppendixBarrier

\begin{table}[!t]
\centering
\small
\caption{Sensitivity to three-axis aggregation weights. Each cell reports the aggregate score (0--100) and rank among nine agents. The weight order is source, replay, and adaptive playtest. Source dependency weighting remains fixed at $\alpha=0.5$. All-game axis scores equally average the \emph{Small} and \emph{Big} means. Bold cells indicate a rank change from equal weighting.}
\label{tab:axis_weight_sensitivity}
\setlength{\tabcolsep}{4pt}
\begin{tabular}{lrrrr}
\toprule
\textbf{Agent} & \textbf{Equal} & \textbf{Source emphasis} & \textbf{Replay emphasis} & \textbf{Adaptive emphasis} \\
 & $(\frac13,\frac13,\frac13)$ & $(\frac12,\frac14,\frac14)$ & $(\frac14,\frac12,\frac14)$ & $(\frac14,\frac14,\frac12)$ \\
\midrule
Claude-Fable-5.1 & 77.0 (1) & 79.4 (1) & 73.8 (1) & 77.7 (1) \\
Claude-Opus-5 & 73.9 (2) & 76.2 (2) & 71.1 (2) & 74.3 (2) \\
GPT-6-Astra & 71.6 (3) & 73.5 (3) & 68.9 (3) & 72.3 (3) \\
GPT-5.5 & 63.2 (4) & 64.5 (4) & \textbf{60.3 (5)} & 64.8 (4) \\
GPT-5.6-Sol & 61.8 (5) & 62.0 (5) & \textbf{60.4 (4)} & 63.1 (5) \\
Claude-Opus-4.8 & 56.8 (6) & \textbf{57.8 (7)} & 55.6 (6) & 56.9 (6) \\
GLM-5.3 & 55.6 (7) & \textbf{58.8 (6)} & 53.2 (7) & 54.9 (7) \\
DeepSeek-V4-Pro & 50.4 (8) & 54.0 (8) & 47.6 (8) & 49.7 (8) \\
Kimi-K2.7 & 39.5 (9) & 41.9 (9) & 37.9 (9) & 38.6 (9) \\
\bottomrule
\end{tabular}
\end{table}

\subsection{Complementarity of Evaluation}
\label{app:complementarity}
\AtoZAppendixInput{figures/appendix/cards_all.tex}

\paragraph{Code Implementation and Runtime Gap.}
Source-code inspection and playtesting provide different evidence for the same rule. Figure~\ref{fig:source_runtime_gap} shows three rules that receive a source-code item score of 1.00 but are judged \emph{violated} during playtesting. In these cases, the recorded source-code judgments credit relevant implementation logic, while execution reveals problems in its integration with the rest of the game. In \texttt{Rainwright}, the preview turns red when \texttt{buildDenied} is set, but the flag is updated only in \texttt{tryBuild()} on input release, so the preview remains teal while the input is held. In \texttt{Strata Keepers}, the map handler deducts two workdays after excavation setup has already reduced the remaining days from 20 to 18, producing a four-day cost. In \texttt{Chroma Bastion}, the run-end path stages a score initialized to 2,760 without replacing it with the completed run's score. These examples expose integration failures missed by the recorded source-code judgments: a condition checked at the wrong time, a duplicated effect, and an outdated value. Linking both forms of evidence to the same requirements makes these discrepancies identifiable and provides concrete targets for revision.

\paragraph{Axis Omission.}
\label{app:axis_omission}

We compare the full three-axis scores with scores recomputed after omitting each axis, using all 100 initial GPT-5.6-Sol builds, comprising 50 \emph{Small} and 50 \emph{Big} games. The full score is the arithmetic mean of the three axes, and each omitted-axis score is the arithmetic mean of the remaining two axes. Reversed orderings are counted over all unordered pairs of games within the same split, giving $2\binom{50}{2}=2{,}450$ comparisons. A pair counts as reversed only when its score ordering strictly changes direction; ties in either ordering remain in the denominator but are not counted as reversals.

\subsection{Sensitivity to Evaluation-Axis Weights}
\label{app:axis_weight_sensitivity}

Equal weighting gives source inspection, replay, and adaptive playtesting the same contribution to the reported aggregate. To assess the effect of this choice, we hold the measured axis scores and source dependency parameter $\alpha=0.5$ fixed and recompute the aggregate as $F(\boldsymbol{\lambda})=\lambda_{\mathrm{src}}F^{\mathrm{src}}+\lambda_{\mathrm{replay}}F^{\mathrm{replay}}+\lambda_{\mathrm{adapt}}F^{\mathrm{adapt}}$, where $\lambda_a\geq0$ and $\sum_a\lambda_a=1$. The All score first averages the \emph{Small} and \emph{Big} means equally within each axis, following Table~\ref{tab:main_results}. These aggregation weights change the relative contribution of the three evidence channels and are distinct from the dependency parameter $\alpha$, which weights rules within source-code evaluation.

For each pair of agents, the score difference is linear in $\boldsymbol{\lambda}$. Its extrema therefore occur at the vertices of the weight simplex, allowing us to determine whether a comparison is invariant over the complete continuous weight domain. Claude-Fable-5.1, Claude-Opus-5, and GPT-6-Astra retain the first three positions in this order for every weighting, separately for \emph{Small}, \emph{Big}, and All. For the All aggregate, 30 of the 36 pairwise orderings are invariant. In this aggregate, if each axis receives at least half its default contribution, $\lambda_a\geq1/6$, 34 of 36 orderings are invariant, and every agent remains within one position of its equal-weight rank.

Table~\ref{tab:axis_weight_sensitivity} reports the default weighting and three alternatives that double one axis relative to either of the others. The changes follow the measured axis strengths. GLM-5.3 exceeds Claude-Opus-4.8 in source-code score (68.4 versus 61.0) but trails it in runtime fidelity (49.3 versus 54.6), so emphasizing source inspection reverses their aggregate order. GPT-5.6-Sol scores higher than GPT-5.5 in replay (56.1 versus 51.6), while GPT-5.5 scores higher in source-code evaluation and adaptive playtesting; emphasizing scenario-based replay changes their order. Reporting the three axes alongside the aggregate makes these differences explicit.

\AtoZAppendixBarrier

\clearpage

\section{Additional Analysis on Evaluation}

\subsection{Dependency-Aware Contract}
\label{app:requirement_analysis}

\subsubsection{Rule Coverage and Run-to-Run Consistency}
\label{app:requirement-representation}

We compare the requirement sets produced by naive judgment and dependency-aware contract construction on all 100 GDDs. The naive judge identifies requirements while evaluating a generated game, whereas the contract is constructed from the GDD alone and represents state constraints as invariants and conditional behaviors as rules, following Appendix~\ref{app:gdd-contract-construction}. Both methods use GPT-5.6-Sol and require a supporting GDD quotation for every extracted item.

This requirement-extraction comparison uses five runs per method and is separate from the three fixed-input rule generations in Appendix~\ref{app:contract_stability}. For each GDD, we construct a shared pooled reference set from the rules referenced across the five runs of each method, consolidating repeated references to the same rule. Rule coverage is the fraction of rules in this shared set that are referenced by a run's extracted evaluation items, with each reference rule counted at most once. We average coverage over the five runs for each GDD. Table~\ref{tab:gdd_rule_coverage} reports the split-level means: the contract increases mean coverage from 0.84 to 0.88 on \emph{Small} and from 0.62 to 0.80 on \emph{Big}. These values measure coverage relative to the rules recovered by repeated extraction.

Citation overlap measures the mean pairwise Jaccard similarity between the sets of referenced requirements across runs of the same GDD. On \emph{Big}, it is 57\% for the naive judge and 93\% for repeated contract extractions. Figure~\ref{fig:app_analysis_contract} summarizes rule coverage and citation overlap across all 100 GDDs. Among naive items that can be matched across runs, judgment status agrees in 91\% of pairs on \emph{Small} and 87\% on \emph{Big}, indicating that much of the instability concerns which requirements are selected for evaluation.

\begin{figure*}[!htbp]
  \centering
  \begin{minipage}[b]{0.46\textwidth}
    \centering
    \captionof{table}{
        \textbf{Coverage against a pooled reference set.}
        Fraction of a shared per-GDD reference set cited by each method's
        extracted items, reported as mean (median) across each split.
        The reference set pools rules referenced across five runs
        of both methods; per-GDD coverage is averaged over five runs.
        The contract is constructed from the GDD alone, whereas the naive
        judge sees both the GDD and the generated game.
    }
\small
\setlength{\tabcolsep}{5pt}
\renewcommand{\arraystretch}{1.08}
\begin{tabular}{lcc}
    \toprule
    \textbf{GDD Split} & \textbf{Naive Judge} & \makecell{\textbf{Dependency-Aware}\\\textbf{Contract (Ours)}} \\
    \midrule
    \emph{Small} & 0.84 (0.92) & \textbf{0.88} (0.90) \\
    \emph{Big}   & 0.62 (0.63) & \textbf{0.80} (0.92) \\
    \bottomrule
\end{tabular}
    \label{tab:gdd_rule_coverage}
  \end{minipage}\hfill
  \begin{minipage}[b]{0.50\textwidth}
    \centering
    {\def\AtoZPanelWidth{0.7\linewidth}\input{figures/intro_contract_stats}}
    \caption{\textbf{Requirement set stability.} Across 100 GDDs, the contract yields more stable requirement sets than naive GDD-based judgment and higher citation overlap. Boxes show interquartile ranges; diamonds and labels mark medians across GDDs.}
    \label{fig:app_analysis_contract}
  \end{minipage}
\end{figure*}
\AtoZAppendixBarrier

\subsubsection{Comparison With Enumerated Evaluation}
\label{app:dependency_source_playtest}

We compare contract-based source evaluation with an \emph{enumerated} baseline, which scores individual GDD-derived requirements without explicitly representing their dependencies. This baseline evaluates a list of GDD-derived requirements, each recording a condition, an action, and an expected outcome, without explicit dependency edges between requirements. Supporting GDD quotations provide the correspondence between these requirements and contract rules. Using one initial GPT-5.6-Sol build for each of the 100 GDDs (50 \emph{Small}, 50 \emph{Big}), we analyze additional inspection targets, their coverage of recorded playtest violations, and implementation omissions identified by comparing against contract-based source evaluation.

\paragraph{Enumerated Source Judgments.} We hold the builds, enumerated source judgments, and rule-to-requirements mapping fixed. Averaging the mapped requirement scores provides source-code scores for 2,589 of the 2,729 frozen contract rules in Section~\ref{app:contract_stability}. We use the same attribute-level links from state updates to conditions as the source scorer (Section~\ref{sec:source_eval}). Full credit here means 1.0, whereas the dependency-adjusted analysis in Section~\ref{app:dependency_analysis} uses a confirmed-pass threshold of 0.8. A full-credit rule is eligible if it has at least one direct predecessor and all predecessor scores are available. We select it for further inspection if any predecessor scores are below 1.0.

This source-only analysis identifies 408 of 908 eligible targets with full enumerated credit (44.9\%) in 53 of the 100 games (Table~\ref{tab:source_dependency_context_100}). The share is 25.7\% in \emph{Small} and 58.6\% in \emph{Big}, with targets in 17 and 36 games, respectively. Of these, 365 targets across 47 games have a predecessor score of 0.5 or lower.

\begin{table}[!htbp]
    \centering\small
    \setlength{\tabcolsep}{6pt}
    \renewcommand{\arraystretch}{1.08}
    \caption{\textbf{Dependency context identifies additional source-code inspection targets.} A target has full enumerated source credit and a lower-scored direct predecessor. Percentages are pooled over eligible targets; the enumerated judgments remain fixed.}
    \label{tab:source_dependency_context_100}
    \begin{tabular}{@{}lrrr@{}}
        \toprule
        \textbf{Measure} & \emph{\textbf{Small}} & \emph{\textbf{Big}} & \textbf{All} \\
        \midrule
        Games & 50 & 50 & 100 \\
        Full-credit rules & 738 & 896 & 1,634 \\
        Eligible full-credit targets & 377 & 531 & 908 \\
        Additional inspection targets & 97 & 311 & 408 \\
        Games with additional targets & 17 & 36 & 53 \\
        Additional targets / eligible targets & 25.7\% & 58.6\% & 44.9\% \\
        Targets with a predecessor score $\leq 0.5$ & 60 & 305 & 365 \\
        \bottomrule
    \end{tabular}
\end{table}

Deduplicating targets with identical mapped requirement sets within each game leaves 361 distinct sets. Excluding targets that share any requirements with a predecessor still leaves 338 targets across 49 games: 80 in 16 \emph{Small} games and 258 in 33 \emph{Big} games.

\paragraph{Runtime Coverage.} We match enumerated source-code scores with conclusive adaptive-playtest verdicts for 2,185 targets, of which 560 are violated. A target is violated if any mapped requirement is violated, and satisfied if all are satisfied. The enumerated baseline inspects rules with mapped source-code scores below 1.0; adding dependency context includes the selected full-credit targets. Coverage is the fraction of recorded violated targets included in each inspection set.

Dependency context increases coverage from 398/560 (71.1\%) to 449/560 (80.2\%), adding 51 violated targets across 26 games to the inspection set (Table~\ref{tab:source_playtest_diagnostic_coverage}). These targets comprise 45 distinct game--requirement sets and account for 31.5\% of the 162 recorded violations outside the enumerated inspection set. The gain is 9.1 percentage points, with a 95\% paired game bootstrap interval of 6.1--13.0 points (20,000 resamples stratified by split). Dependency context thus directs inspection to execution failures that received full enumerated source credit.

\paragraph{Source-Judgment Comparison.} We compare archived enumerated (GPT-5.6-Luna) and contract-based (GPT-5.6-Luna) source judgments on identical source revisions, matching requirements through their cited GDD requirements. Among 2,446 rules with paired scores, 244 rules across 68 games receive full enumerated credit but less than full contract-based credit. The reverse occurs for 375 rules across 62 games, yielding 619 disagreements in full-credit status.

A retrospective assistant review checks all 619 disagreements against the original requirements, both judgment rationales, and the source code. It identifies implementation omissions in 17 rules across 13 \emph{Big} games that received full enumerated credit, corresponding to 15 distinct mapped requirement sets. Sixteen rules across 12 games concern conditions, state propagation, transitions, persistence, or ordering. The following examples illustrate how these connections reveal missing behavior.

In \texttt{Grand Atelier 2D}, rules \texttt{R-CL2} and \texttt{R-CL3} require a submitted collection with insufficient grades or inconsistent composition to fail with a reputation penalty of 12. The enumerated judgments credit the grade and coherence checks. However, \texttt{toggleGarment} enables submission only for passing collections, and \texttt{submitCollection} returns otherwise. These checks block the required failure transition, leaving both the failure branch and its penalty unimplemented (\texttt{scenes/\allowbreak{}AtelierScene.ts}, lines 412--415).

In \texttt{Last Announcement}, rule \texttt{RT-CAPTURE-1} requires chase success to take priority over capture, remove the threat, and update the retry checkpoint. The enumerated judgment credits the success-before-capture ordering. The \texttt{chaseSuccess} handler removes the threat and records completion but never updates \texttt{retryCheckpoint}, omitting the required link between encounter completion and retry state (\texttt{scenes/\allowbreak{}GameScene.ts}, lines 222--227 and 442--451).

In \texttt{Whispers of the Wild}, rule \texttt{L1} requires a qualifying landscape photo to receive a bonus and set the species' \texttt{landscapeBadge}. The code calculates the condition and bonus but updates the badge only when the photo also improves the species' best base score. A qualifying photo below that record therefore misses the required badge (\texttt{scenes/\allowbreak{}FieldScene.ts}, lines 451--455).

In \texttt{Rainwright}, rule \texttt{R5\#4} requires an overlapping pillar to move to the nearest free adjacent tile. The enumerated judgment credits a one-tile offset. The code shifts the pillar without checking whether the destination is free, so an occupied adjacent tile produces another overlap (\texttt{scenes/\allowbreak{}PlayScene.ts}, lines 844--853).

Tracing each condition through its required effects identifies the missing branch, state update, or constraint check and its code location. This produces concrete guidance for the contract-based revision setting in Section~\ref{sec:iterative_revision}.

\AtoZAppendixBarrier

\subsection{Cost Analysis}
\label{app:cost_evaluation}

\subsubsection{Source-Code Evaluation}
\AtoZAppendixInput{tables/appendix/source_code_cost}

We examine how judge model and reasoning effort affect source-code evaluation. We hold the contracts and builds fixed and compare seven configurations against fixed reference judgments produced by GPT-5.6-Sol at extra-high reasoning effort. An independent re-run of GPT-5.6-Sol at extra-high effort yields a rule-level Pearson correlation of 0.74 with those reference judgments. This re-run provides a baseline for interpreting agreement across configurations.

As shown in Table~\ref{tab:fid-effort}, the seven alternative configurations have rule-level correlations from 0.69 to 0.80 with the reference judgments, comparable in magnitude to the 0.74 of the independent re-run. Mean rule scores $f_{i,r}^{\mathrm{src}}$ range from 0.76 to 0.80 across all eight runs.

The benchmark's default configuration is GPT-5.6-Luna at high effort, as described in Section~\ref{sec:experimental_setup}. It has a rule-level correlation of 0.72 at an estimated cost of $0.15\times$ the independent sol re-run; at low effort, luna has a correlation of 0.71 at $0.12\times$. Across luna settings, invariant agreement ranges from 0.73 to 0.77, against 0.83 for the independent sol re-run. Luna's rule-level agreement is therefore comparable in magnitude to the re-run baseline at substantially lower estimated cost.

\AtoZAppendixBarrier

\subsubsection{Scenario-Based Replay Assessment}

\begin{figure*}[!htbp]
  \centering
  \begin{minipage}[t]{0.48\textwidth}\vspace{0pt}\begin{tikzpicture}
\begin{axis}[
  a2zfig,
  scale only axis,
  width=0.79\linewidth,
  height=0.32\linewidth,
  ybar,
  bar width=16pt,
  bar shift=0pt,
  ymin=0,
  ymax=5.6,
  symbolic x coords={480,640,1280,1920},
  xtick={480,640,1280,1920},
  xticklabels={
    \textbf{480px},
    \textbf{640px},
    \textbf{1280px},
    \textbf{Native}
  },
  xlabel={},
  /pgf/number format/assume math mode=true,
  every tick label/.append style={font=\rmfamily\scriptsize},
  every node near coord/.append style={
    font=\rmfamily\bfseries\scriptsize,
    inner sep=2pt,
    /tikz/shade=false,
    /tikz/fill=none,
    /tikz/draw=none
  },
  ylabel style={font=\rmfamily\footnotesize},
  title style={font=\rmfamily\bfseries\footnotesize},
  ylabel={K tokens / frame},
  ymajorgrids,
  grid style={A2ZFigAxis!66!white,line width=0.35pt},
  axis lines*=left,
  axis line style={A2ZFigAxis,line width=0.5pt},
  tick style={A2ZFigAxis,line width=0.4pt},
  title={Tokens per frame (fixed prompt excluded)}
]

\addplot[
  fill=A2ZGameCraft,
  draw=A2ZFigAxis!70!black,
  line width=0.22pt,
  point meta=y,
  nodes near coords={
    \pgfmathprintnumber[fixed, precision=1, assume math mode]{\pgfplotspointmeta}
  }
]
coordinates {(640,0.5) (1280,2.0) (1920,4.7)};

\addplot[
  fill=A2ZBGG,
  draw=A2ZFigAxis!70!black,
  line width=0.22pt,
  point meta=y,
  nodes near coords={
    \pgfmathprintnumber[fixed, precision=2, assume math mode]{\pgfplotspointmeta}
  }
]
coordinates {(480,0.36)};

\end{axis}
\end{tikzpicture}\end{minipage}\hfill
  \begin{minipage}[t]{0.48\textwidth}\vspace{0pt}\begin{tikzpicture}
\begin{axis}[a2zfig, scale only axis, width=0.79\linewidth, height=0.32\linewidth, ybar, bar width=30pt, bar shift=0pt, ymin=0.8, ymax=1.02, ytick={0.8,0.85,0.9,0.95,1.0},
  enlarge x limits=0.35,
  symbolic x coords={a,b}, xtick={a,b},
  xticklabels={480px vs 640px ,640px vs Native},
  xticklabel style={align=center, font=\rmfamily\bfseries\scriptsize},
  yticklabel style={font=\rmfamily\scriptsize},
  /pgf/number format/assume math mode=true,
  every node near coord/.append style={font=\rmfamily\bfseries\scriptsize, inner sep=2pt,
    /pgf/number format/.cd, fixed, precision=3},
  xlabel style={font=\rmfamily\footnotesize},
  ylabel style={font=\rmfamily\footnotesize},
  title style={font=\rmfamily\bfseries\footnotesize},
  ylabel={item-level Pearson $r$},
  nodes near coords, nodes near coords align={vertical}, point meta=rawy,
  ymajorgrids, grid style={A2ZFigAxis!66!white,line width=0.35pt},
  axis lines*=left, axis line style={A2ZFigAxis,line width=0.5pt}, tick style={A2ZFigAxis,line width=0.4pt},
  every node near coord/.append style={/tikz/shade=false,/tikz/fill=none,/tikz/draw=none},
  title={Agreement with the next-larger width}]
\addplot[
  fill=A2ZBGG,
  draw=A2ZFigAxis!70!black,
  line width=0.22pt
] coordinates {(a,0.921)};

\addplot[
  fill=A2ZGameCraft,
  draw=A2ZFigAxis!70!black,
  line width=0.22pt
] coordinates {(b,0.967)};
\end{axis}
\end{tikzpicture}\end{minipage}
  \caption{\textbf{Tokens vs. frame width}. Left: tokens per frame, fixed prompt excluded. Right: item-level agreement with the next-larger width.}
  \label{fig:vlm-res}
\end{figure*}
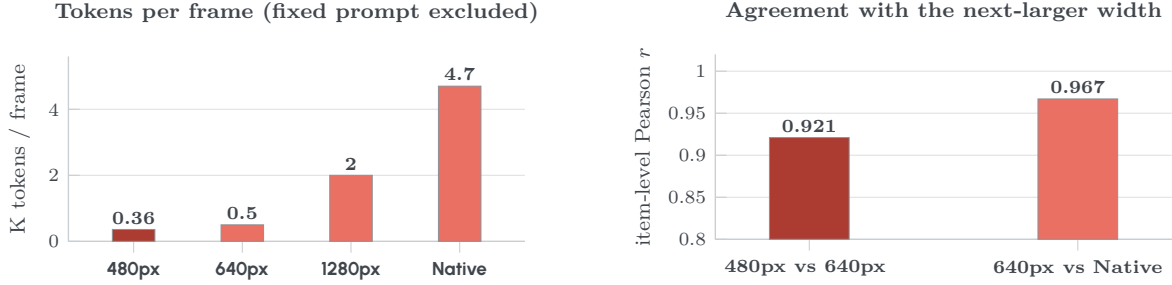

\begin{figure*}[!htbp]
  \centering
\begingroup
\setlength{\tabcolsep}{3pt}\renewcommand{\arraystretch}{0.9}
\rmfamily\footnotesize\color{A2ZFigText}
\begin{tabular}{@{}l@{\hspace{2pt}}ccc@{}}
 & \textbf{Native} & \textbf{640\,px} & \textcolor{A2ZBGG}{\textbf{480\,px}} \\
\rotatebox{90}{\scriptsize 1920$\times$1080} &
 \includegraphics[width=0.31\linewidth]{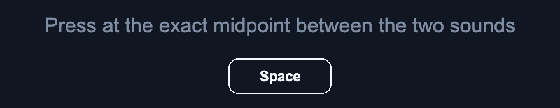} &
 \includegraphics[width=0.31\linewidth]{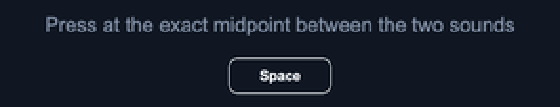} &
 \includegraphics[width=0.31\linewidth]{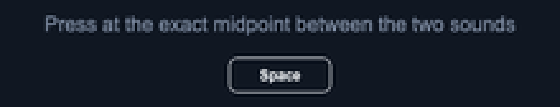} \\[2pt]
\rotatebox{90}{\scriptsize 1280$\times$720} &
 \includegraphics[width=0.31\linewidth]{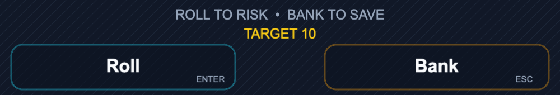} &
 \includegraphics[width=0.31\linewidth]{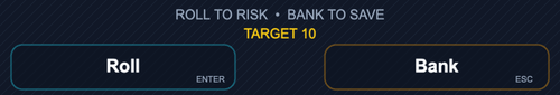} &
 \includegraphics[width=0.31\linewidth]{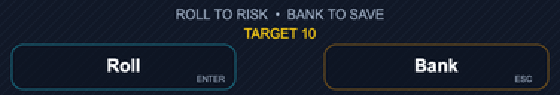} \\
\end{tabular}
\endgroup

  \caption{\textbf{A sampled frame at different resolutions.} The same frame region as the judge receives it at native resolution, 640\,px and 480\,px (Lanczos downscaling as in the pipeline). Top: 1920$\times$1080 game; bottom: 1280$\times$720 game.}
  \label{fig:vlm-res-example}
\end{figure*}

The number of image tokens increases with frame width, as shown on the left of Figure~\ref{fig:vlm-res}. A native-width frame uses about 4.7K image tokens, compared with 2.0K at 1280 pixels and 0.5K at 640 pixels. These values exclude the fixed prompt of roughly 39K tokens per call, one call per scenario replay $\xi_{i,q}$. Across six \emph{Small} and \emph{Big} games, evaluation at 640 pixels agrees closely with evaluation at native resolution, with a Pearson correlation of 0.967 between the per-item scores of $\rho_i^{\mathrm{vis}}$. Each call at 640 pixels uses approximately 45K tokens. Reducing the frame width from 640 to 480 pixels lowers image tokens per frame from 0.5K to 0.36K, a 28\% reduction. When both image dimensions scale proportionally, this is smaller than the 44\% reduction in pixel count. Total tokens per call decrease little because the fixed prompt accounts for most of the usage. At 480 pixels, the per-item correlation with the 640-pixel setting drops to 0.921. As shown in Figure~\ref{fig:vlm-res-example}, the word ``space'' becomes increasingly fragmented and difficult to read at lower resolutions.

\AtoZAppendixBarrier

\subsubsection{Adaptive Playtest}
\label{app:playtest-model}

Both the bot author $A^{\mathrm{play}}$ and the trace judge were developed with GPT-5.6-Sol; our default evaluation uses GPT-5.6-Luna at high reasoning effort.
On fixed traces (Table~\ref{tab:playtest-judge}), Luna achieves rule-level agreement of 0.910 on \emph{Small} and 0.888 on \emph{Big} against the Sol high reference, compared with 0.959 on both splits for an independent re-run of GPT-5.6-Sol at high reasoning effort.
Flips that change $f_{i,r}^{\mathrm{adapt}}$ occur on 6 of 222 \emph{Small} rules and 3 of 242 \emph{Big} rules, compared with 3 and 1, respectively, for the Sol re-run.
Token usage per vote decreases from 516K--688K to 51K--71K.

For the bot author, judgment coverage (Section~\ref{app:playtest_protocol}) differs by 0.15--0.17 between Sol and Luna runs, compared with 0.11--0.16 between repeated runs of the same model.
These ranges partially overlap, but do not establish equivalence between model substitution and re-running.
With three judgments per rule retained, the reported per-game playtest cost decreases from \$0.65--0.73 to \$0.08--0.11.

\begin{table}[!htbp]
    \centering
    \caption{\textbf{Playtest judge comparison on fixed traces.} Each cell reports verdict agreement with a GPT-5.6-Sol high reference, followed by the number of rules whose binary satisfaction score changes (\emph{Small}: 222 rules; \emph{Big}: 242 rules). Each rule receives three judgments. An independent Sol high re-run provides a reference for run-to-run variation.}
    \label{tab:playtest-judge}
    \small
    \setlength{\tabcolsep}{10pt}
    \renewcommand{\arraystretch}{1.15}
    \begin{tabular}{@{}l c c c@{}}
        \toprule
        \textbf{Judge, effort} & \textbf{\emph{Small}} & \textbf{\emph{Big}} & \textbf{Tokens / vote} \\
        \midrule
        Sol, xhigh & 0.959 / 2 & 0.930 / 4 & 485--753K \\
        Sol, high (re-run) & 0.959 / 3 & 0.959 / 1 & 516--688K \\
        Sol, low & 0.892 / 7 & 0.897 / 3 & 217--315K \\
        \midrule
        Luna, xhigh & 0.919 / 3 & 0.893 / 1 & 57--86K \\
        Luna, high & 0.910 / 6 & 0.888 / 3 & 51--71K \\
        Luna, low & 0.833 / 9 & 0.810 / 10 & 46--64K \\
        \bottomrule
    \end{tabular}
\end{table}

\AtoZAppendixBarrier

\subsection{Fixed-Rate Sampling vs. Adaptive Frame Selection}
\label{app:frame_selection}

\paragraph{Experimental Setup.} The comparison in Table~\ref{tab:adaptive_frame_selection} uses games built by GPT-5.6-Sol on all 50 \emph{Small} GDDs, comprising 244 scenario replays and 4,102 paired rubric-item assessments. Both conditions use the same recording and visual rubric, and the number of original-resolution frames matches for each replay (2,268 per condition in total). Two games, \texttt{Bin Bit} and \texttt{Same Or Shift}, are also used in preliminary diagnostics.

\paragraph{Selection and Judging.} Adaptive selection starts from the first and last frames and uses the input trace, scenario entry condition, and visual rubric to request additional timestamps. A deterministic tool retrieves the first captured frame at or after each requested time, allowing subsequent requests to use the observed frames. All inspected frames count toward the matched frame budget and are passed to the final judge. The selector and final judge use GPT-5.6-Sol with medium reasoning effort in separate sessions, and the final judge receives only the selected frames and rubric.

\paragraph{Audit and Metrics.} A separate GPT-5.6-Sol session with high reasoning effort compares how well the cited frames support the two rationales for each rubric item. It selects the better-supported rationale or assigns an \emph{equivalent} or \emph{insufficient-evidence} outcome. Method names and frame identifiers are masked, with each method appearing first in 122 scenarios, although image paths retain source directory names. This audit measures rationale support rather than accuracy against reference labels. \emph{Item preference} is each method's share of the 1,159 comparisons favoring either method, pooled across all rubric items and scenarios. Equivalent and insufficient-evidence outcomes are excluded from this denominator. \emph{Game wins} counts games in which a method receives more item-level preferences across its scenarios; equal totals are ties. Adaptive selection obtains 60.5\% item preference and 35 game wins, compared with 39.5\% and 13 for fixed-rate sampling, with two tied games. Table~\ref{tab:frame_selection_outcomes} includes the complete outcomes.

\begin{table}[!htbp]
    \centering
    \caption{\textbf{Rationale-Support Outcomes.} Counts and percentages over all 4,102 paired rubric-item assessments.}
    \label{tab:frame_selection_outcomes}
    \small
    \setlength{\tabcolsep}{8pt}
    \renewcommand{\arraystretch}{1.08}
    \begin{tabular}{@{}lrr@{}}
        \toprule
        \textbf{Audit Outcome} & \textbf{Assessments} & \textbf{Share (\%)} \\
        \midrule
        Fixed-rate preferred & 458 & 11.2 \\
        Adaptive preferred & 701 & 17.1 \\
        Equivalent & 2,848 & 69.4 \\
        Insufficient evidence & 95 & 2.3 \\
        \midrule
        \textbf{Total} & 4,102 & 100.0 \\
        \bottomrule
    \end{tabular}
\end{table}

\paragraph{Qualitative Examples.} Figure~\ref{fig:adaptive_frame_selection_qualitative} demonstrates that adaptive selection captures intermediate states that uniform sampling omits. In \texttt{Blackout Dispatch} (Figure~\ref{fig:afs_blackout}), the Confirm button is disabled during a drag, enabled after both tokens are assigned, and disabled following confirmation; uniform sampling fails to capture the enabled state. In \texttt{Chalk Escape} (Figure~\ref{fig:afs_chalk}), adaptive frames display a drawn path and its subsequent removal upon wall contact, while uniform frames only depict the pre-drawing state and the aftermath of the collision, omitting the path itself. These findings provide direct visual evidence for the specified transitions, rather than requiring inference from the final state alone.

\AtoZAppendixInput[p]{figures/appendix/frame_selection/qualitative}

\AtoZAppendixBarrier

\subsection{Gameplay Agent vs. Adaptive Playtest}
\label{app:bot_code_as_policy}

We compare the evaluation information provided by a gameplay agent from Orak~\citep{park2026orak} and our adaptive playtesting with~\textsc{Code-as-Policies} on ten \emph{Big} games generated by GPT-5.6-Sol. The online agent uses GPT-5.6-Luna with high reasoning effort, three seeds, and at most 200 action steps per trial. It receives textual state observations and clickable controls, and is scored by game-specific milestone attainment. Our playtests instead return judgments linked to individual GDD requirements and their execution evidence.

\texttt{Toybox Rider} illustrates the distinction. All three online trials receive credit for reaching \texttt{Art Desk}; the implemented milestone checks whether the area or its speed stage has been reached. In a separate normal-play trace of the same build, our evaluator reports that the game enters \texttt{Art Desk} at 180.23\,s but returns to Block Street at 182.98\,s, violating the prescribed area sequence and duration. It also reports a moving obstacle at 7.37\,s, despite the GDD excluding moving obstacles during the first 120\,s of the tutorial. Neither condition is checked by the implemented milestone predicates.

Adversarial playtesting extends this process to requirements left unverified during normal play (Table~\ref{tab:orak_playtest_coverage}). Across the same ten games, recorded judgment coverage---the fraction of requirements judged satisfied or violated, $(S+V)/N$---increases from 10.0\% with normal play to 73.4\% after adding adversarial results.

\begin{table}[!htbp]
    \centering
    \caption{\textbf{Recorded results for all ten games.} Gameplay agent progress is milestone attainment (mean $\pm$ sample SD over three seeds); Judgment coverage is $(S+V)/N$ over the archived GDD-derived requirements. The percentages use different target sets. Coverage is calculated from the saved judgments, with unverified requirements retained in the denominator.}
    \label{tab:orak_playtest_coverage}
    \small
    \setlength{\tabcolsep}{6pt}
    \renewcommand{\arraystretch}{1.1}
    \begin{tabular}{@{}l c c r r@{}}
        \toprule
        & \textbf{Gameplay Agent} & \textbf{Adaptive Playtest} & \multicolumn{2}{c}{\textbf{Judgment Coverage (\%)}} \\
        \cmidrule(lr){4-5}
        \textbf{Game} & \textbf{Progress (\%)} & \textbf{Requirements $N$} & \textbf{Normal} & \textbf{Normal + Adv.} \\
        \midrule
        \texttt{Abyssal Chain} & $6.7 \pm 0.0$ & 110 & 6.4 & 89.1 \\
        \texttt{Alias Alchemy Shop} & $13.3 \pm 0.0$ & 133 & 12.0 & 91.0 \\
        \texttt{Beat Reroute} & $26.7 \pm 0.0$ & 84 & 19.1 & 86.9 \\
        \texttt{Chameleon Slide} & $13.3 \pm 0.0$ & 87 & 0.0 & 67.8 \\
        \texttt{Fogfall Delivery} & $15.6 \pm 10.2$ & 119 & 5.0 & 88.2 \\
        \texttt{Night Shift Rx} & $23.1 \pm 0.0$ & 96 & 8.3 & 87.5 \\
        \texttt{Today's Best Spot} & $35.7 \pm 0.0$ & 103 & 5.8 & 62.1 \\
        \texttt{Toybox Rider} & $40.0 \pm 0.0$ & 112 & 20.5 & 88.4 \\
        \texttt{ToyFit} & $0.0 \pm 0.0$ & 100 & 1.00 & 40.0 \\
        \texttt{Whispers of the Wild} & $42.2 \pm 10.2$ & 117 & 21.4 & 32.5 \\
        \bottomrule
    \end{tabular}
\end{table}

Compared with the implemented gameplay-agent milestone scores, our benchmark provides a more specific account of implementation errors: it names the violated requirement and links it to execution evidence. That information gives a revision agent a concrete behavior to correct.

\clearpage

\section{Additional 3D Evaluation Results}
\label{app:scalable_benchmark}

\paragraph{Evaluation Setup.}
We evaluate the three GPT-6-Astra builds introduced in Section~\ref{sec:3d_extension}: \texttt{Sonic: Cascade Coast}, \texttt{Diablo Cathedral}, and \texttt{Rocket League}. These tasks cover momentum-based platforming, dungeon exploration and combat, and vehicle-based ball play. For each game, source-code inspection, scenario-based replay, and adaptive playtesting assess the reported build against its fixed GDD-derived contract. The Three.js runtime exposes the observation and interaction interfaces needed by the test policies, including precondition initialization for adversarial playtests. Figure~\ref{fig:3d_extension} shows requirement summaries and gameplay captures for two of the evaluated builds.

\AtoZAppendixInput{tables/exp_3d_transfer}

Table~\ref{tab:scalability_3d} shows that the framework produces complementary fidelity assessments in this 3D environment. \texttt{Sonic} receives a source-code score of 99.71, but its Replay and Playtest scores are 53.50 and 80.00, respectively. Thus, high implementation credit alone does not establish fidelity in rendered output or observed gameplay. These cases demonstrate the applicability of the contract-based evaluation structure beyond the main Phaser setting.

\paragraph{Requirement-Level Feedback.}
The evaluation also identifies concrete revision targets in these games. For example, \texttt{Sonic}'s GDD requires landing velocity to be projected onto the contacted tangent without creating tangential speed. The source evaluator identifies a rail-landing implementation that imposes a minimum speed of 12~m/s, which can increase speed when the projected component is smaller. This diagnosis points to a targeted revision of the velocity update and a subsequent test of low-speed rail landings. Linking such findings to the fixed contract allows a coding agent to revise the game and be reassessed against the same requirements, extending the framework's use to specification-driven development and refinement in additional environments.

\clearpage

\section{More Visual Results}

\subsection{GDD-Specific Revision Cases}
\label{app:gdd_revision_cases}

\begin{figure*}[!p]
\centering
\begingroup
\input{figures/revision_cases/appendix_helpers}
\begin{subcaptiongroup}
\resizebox{\linewidth}{!}{%
\begin{tikzpicture}[x=1cm,y=-1cm,text=RCInk,font=\AtoZFigFont,every node/.style={inner sep=0pt,outer sep=0pt}]
\path[use as bounding box] (0,0) rectangle (15.44,19.31);
\begin{scope}[shift={(0,0.00)}]
\RCArtRow{arms={3},
  label={fig:revision_art_toybox},
  game={\texttt{Toybox Rider}},
  rule={UI-02},
  condition={Restart selected\\from Pause.},
  effect={Ask for confirmation\\before discarding\\the current run.},
  effect y={2.47}}
\RCArtView{column={1},
  file={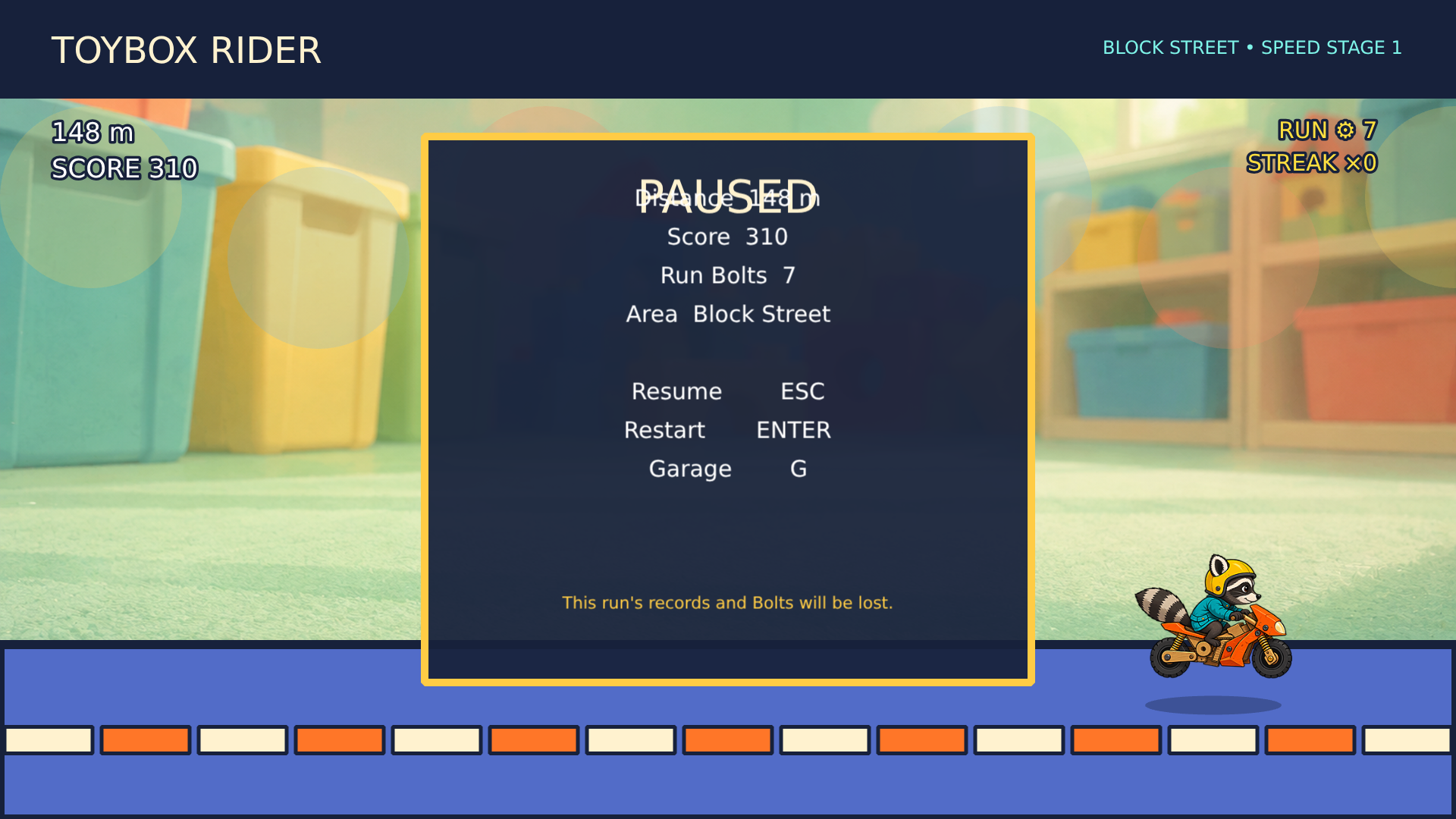},
  source width={1920},
  source height={1080},
  crop x={680},
  crop y={332},
  crop width={560},
  outcome={Still in Pause},
  color={RCInk},
  annotation={}}
\RCArtView{column={2},
  file={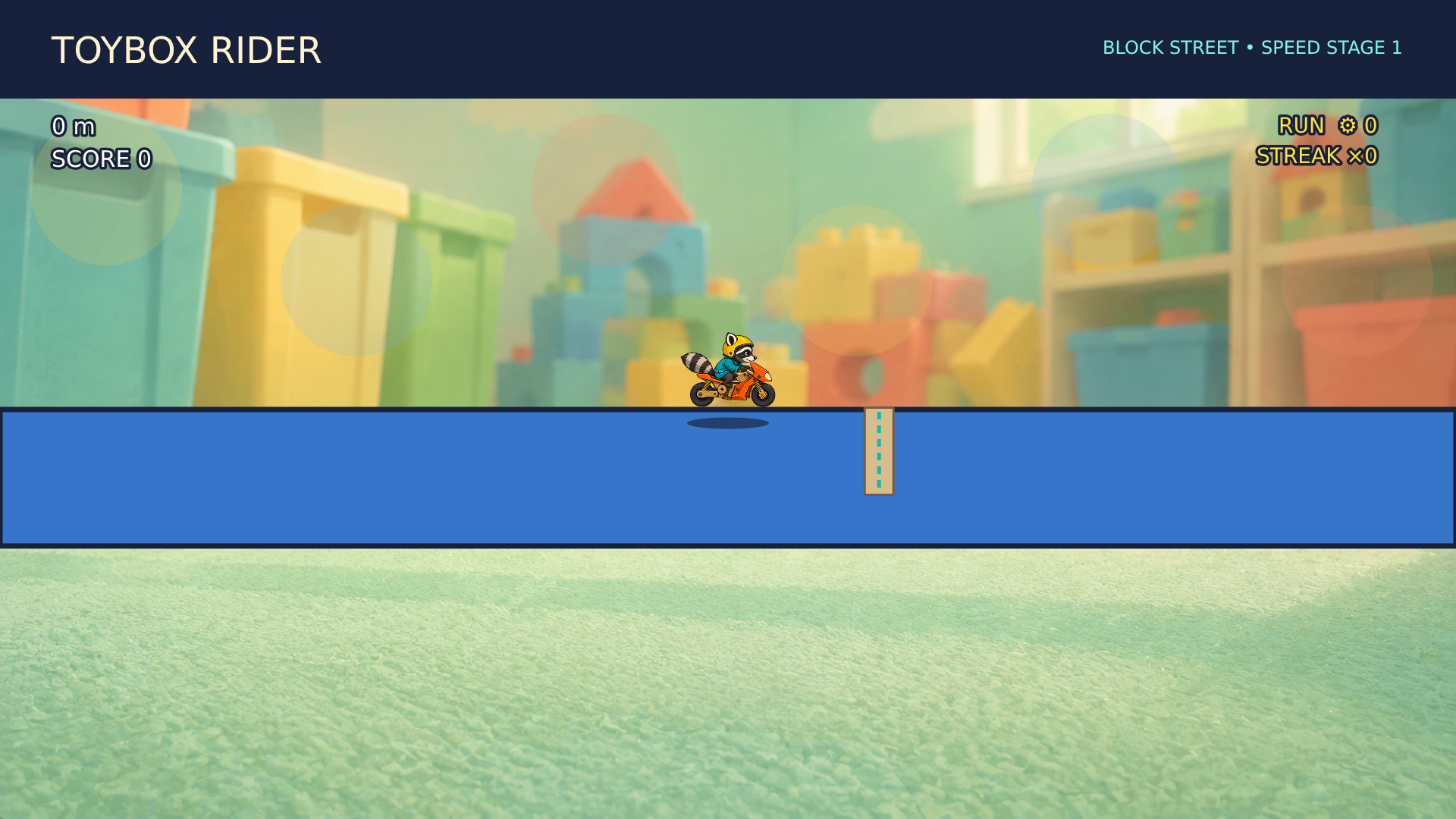},
  source width={1920},
  source height={1080},
  crop x={30},
  crop y={140},
  crop width={420},
  outcome={Restarts without prompt},
  color={RCInk},
  annotation={}}
\RCArtView{column={3},
  file={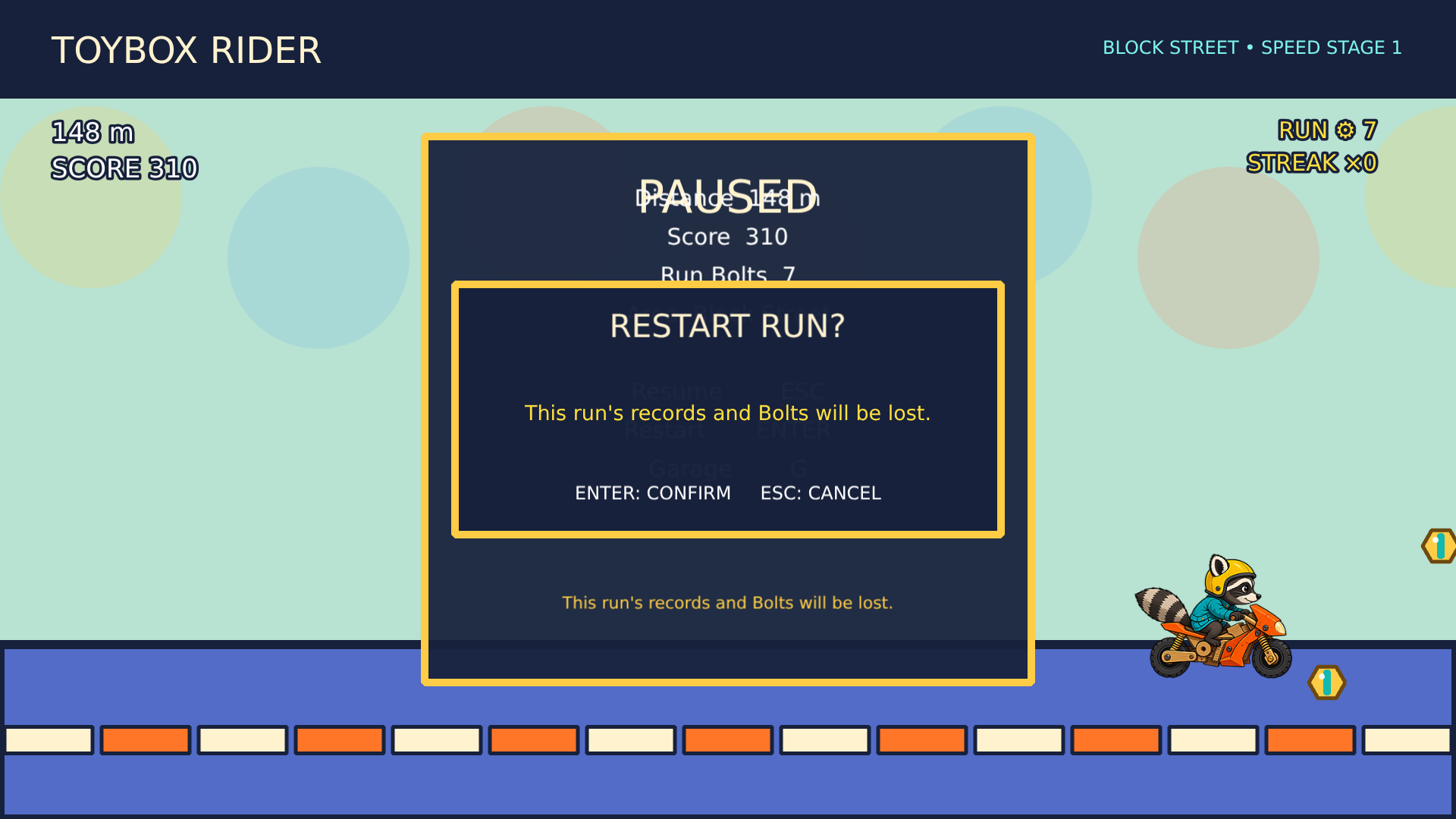},
  source width={1920},
  source height={1080},
  crop x={690},
  crop y={392},
  crop width={560},
  outcome={Confirmation opens},
  color={RCAccent},
  annotation={}}
\end{scope}
\hfill
\begin{scope}[shift={(0,4.87)}]
\RCArtRow{arms={3},
  label={fig:revision_art_nightshift_timeout},
  game={\texttt{Night Shift Rx}: Boss timeout},
  rule={\shortstack[l]{RT-STAGE R1\\VFX-CODEBLUE}},
  condition={Boss timer expires;\\patient uncured.},
  effect={Enter CODE BLUE;\\show a red-grey ring.},
  effect y={2.47}}
\RCArtView{column={1},
  file={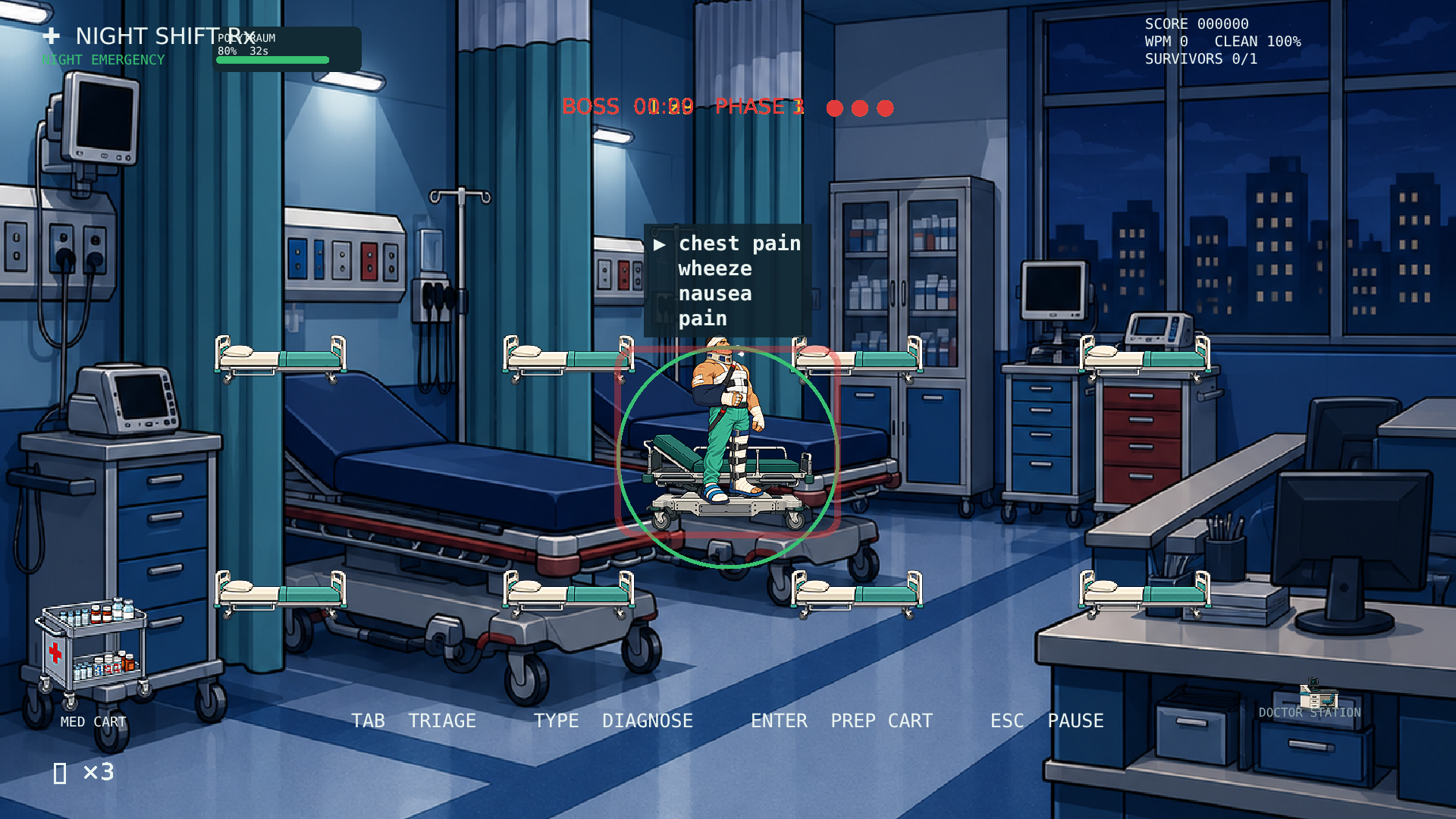},
  source width={1920},
  source height={1080},
  crop x={410},
  crop y={425},
  crop width={1100},
  outcome={Life 79.8; green},
  color={RCInk},
  annotation={}}
\RCArtView{column={2},
  file={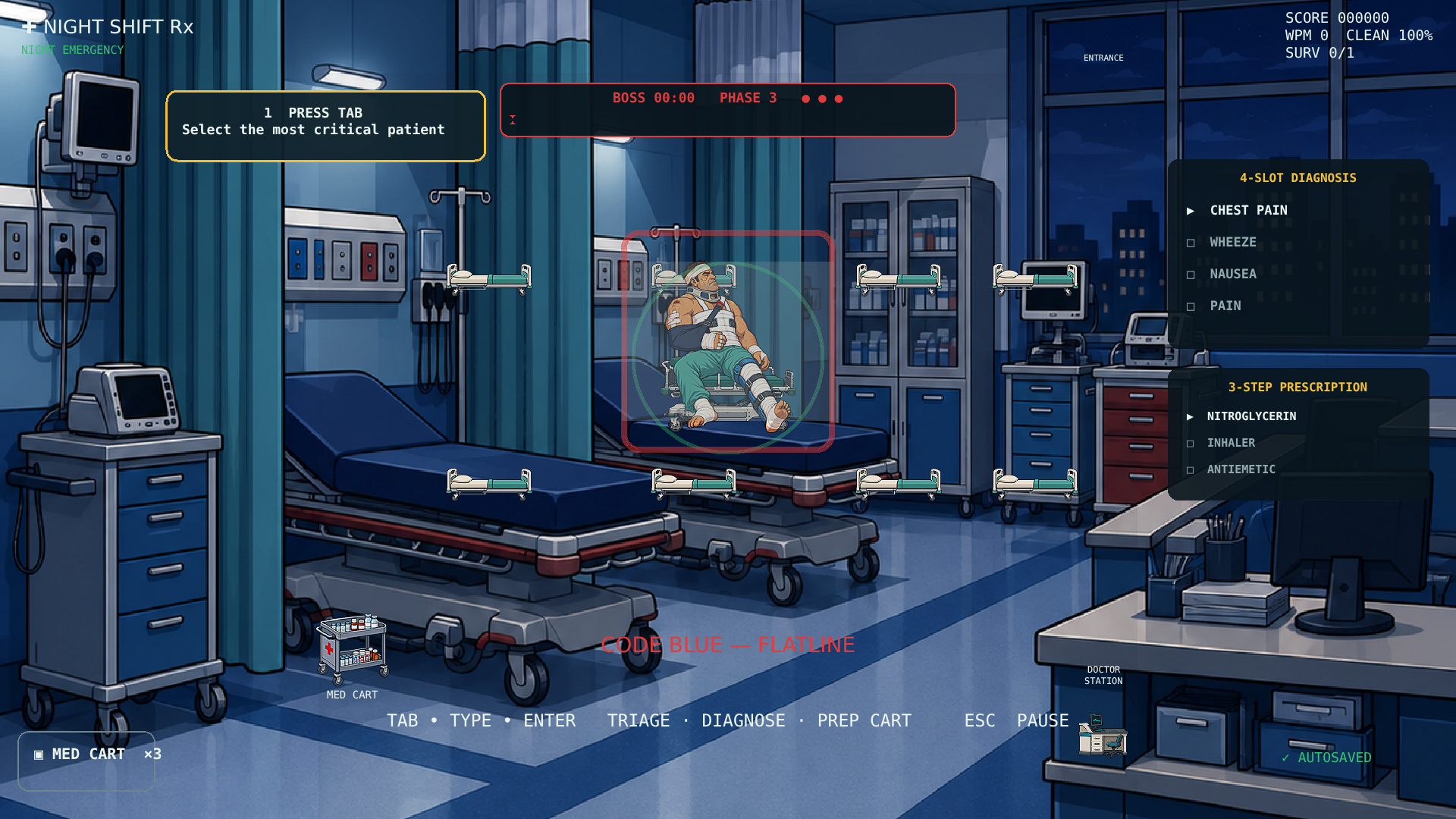},
  source width={1920},
  source height={1080},
  crop x={495},
  crop y={315},
  crop width={930},
  outcome={Life 79.8; green},
  color={RCInk},
  annotation={}}
\RCArtView{column={3},
  file={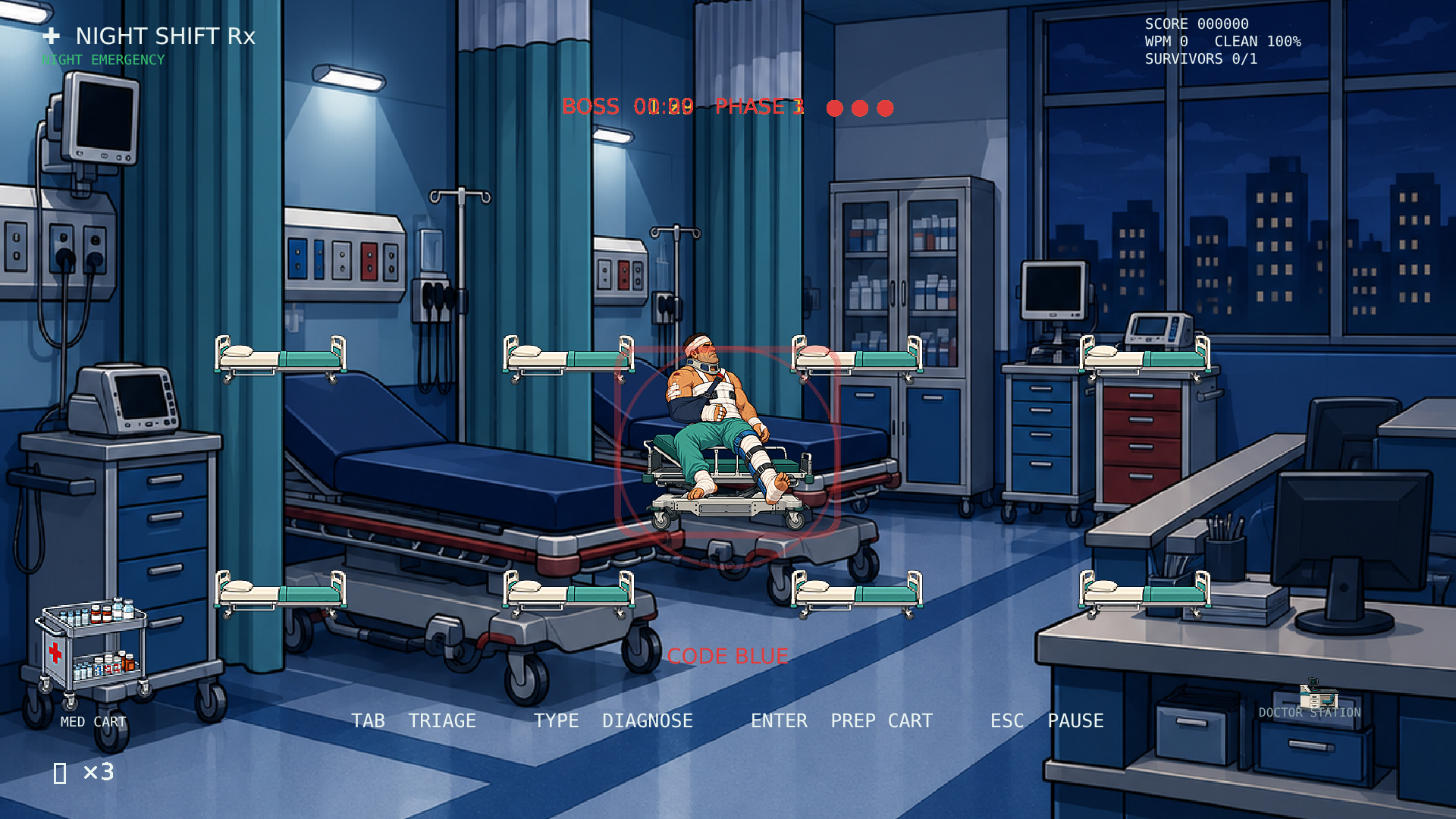},
  source width={1920},
  source height={1080},
  crop x={410},
  crop y={425},
  crop width={1100},
  outcome={Life 0; red-grey},
  color={RCAccent},
  annotation={}}
\end{scope}
\hfill
\begin{scope}[shift={(0,9.74)}]
\RCArtRow{arms={3},
  label={fig:self_revision_alias_selection},
  game={\texttt{Alias Alchemy Shop}: Opening the selected shop},
  rule={\texttt{AliasSelection}},
  condition={Click signboard 2;\\then \textsc{Open Shop}.},
  effect={Open the shop with\\the selected alias.},
  effect y={2.47}}
\newcommand{\RCAliasSequence}[7]{%
  \begingroup
  \pgfmathsetmacro{\AliasX}{4+(#1-1)*3.84}
  \node[font=\AtoZFigFont\fontsize{7.0}{8.8}\selectfont\bfseries,text=RCInk]
    at ({\AliasX+1.75},1.07) {1. #5};
  \RCAppendixCrop{\AliasX}{1.25}{3.50}{.62}{#3}{#4}{460}{alias_open_#2_selected.png}{1920}
  \node[font=\AtoZFigFont\fontsize{7.0}{8.8}\selectfont\bfseries,text=RCMuted]
    at ({\AliasX+1.75},2.07) {2. Click \textsc{Open Shop}};
  \node[anchor=north west] at (\AliasX,2.22)
    {\includegraphics[width=3.50cm]{assets/revision_cases/alias_open_#2_after_open.png}};
  \draw[RCBorder,line width=.4pt] (\AliasX,2.22) rectangle ({\AliasX+3.50},4.18875);
  \path[fill=#7!6!white,draw=RCBorder,line width=.4pt]
    (\AliasX,4.20) rectangle ({\AliasX+3.50},4.62);
  \node[ra art result,text=#7] at ({\AliasX+1.75},4.41) {#6};
  \endgroup
}
\RCAliasSequence{1}{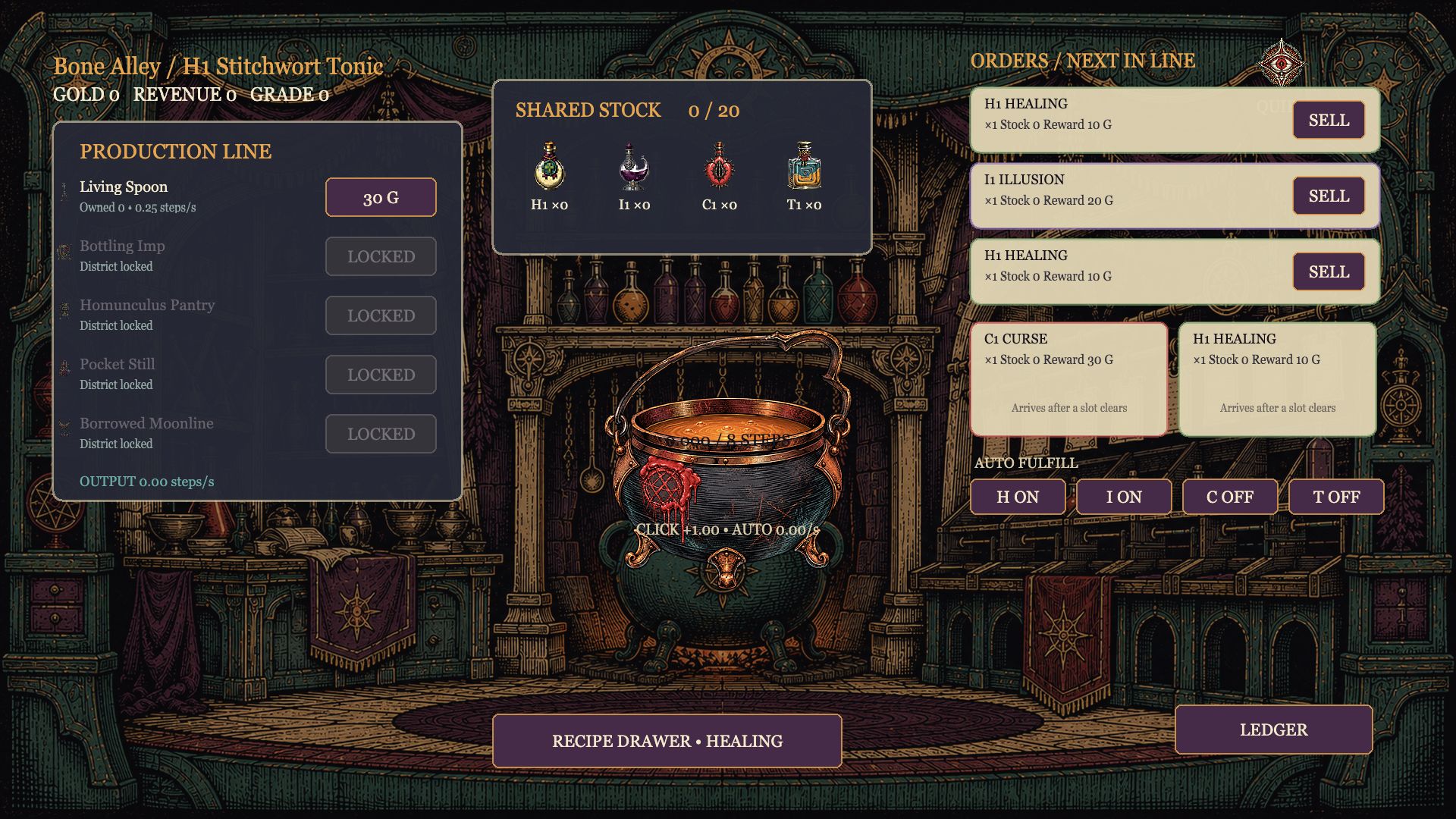}{250}{400}{First stays selected}{Selection ignored}{RCInk}
\RCAliasSequence{2}{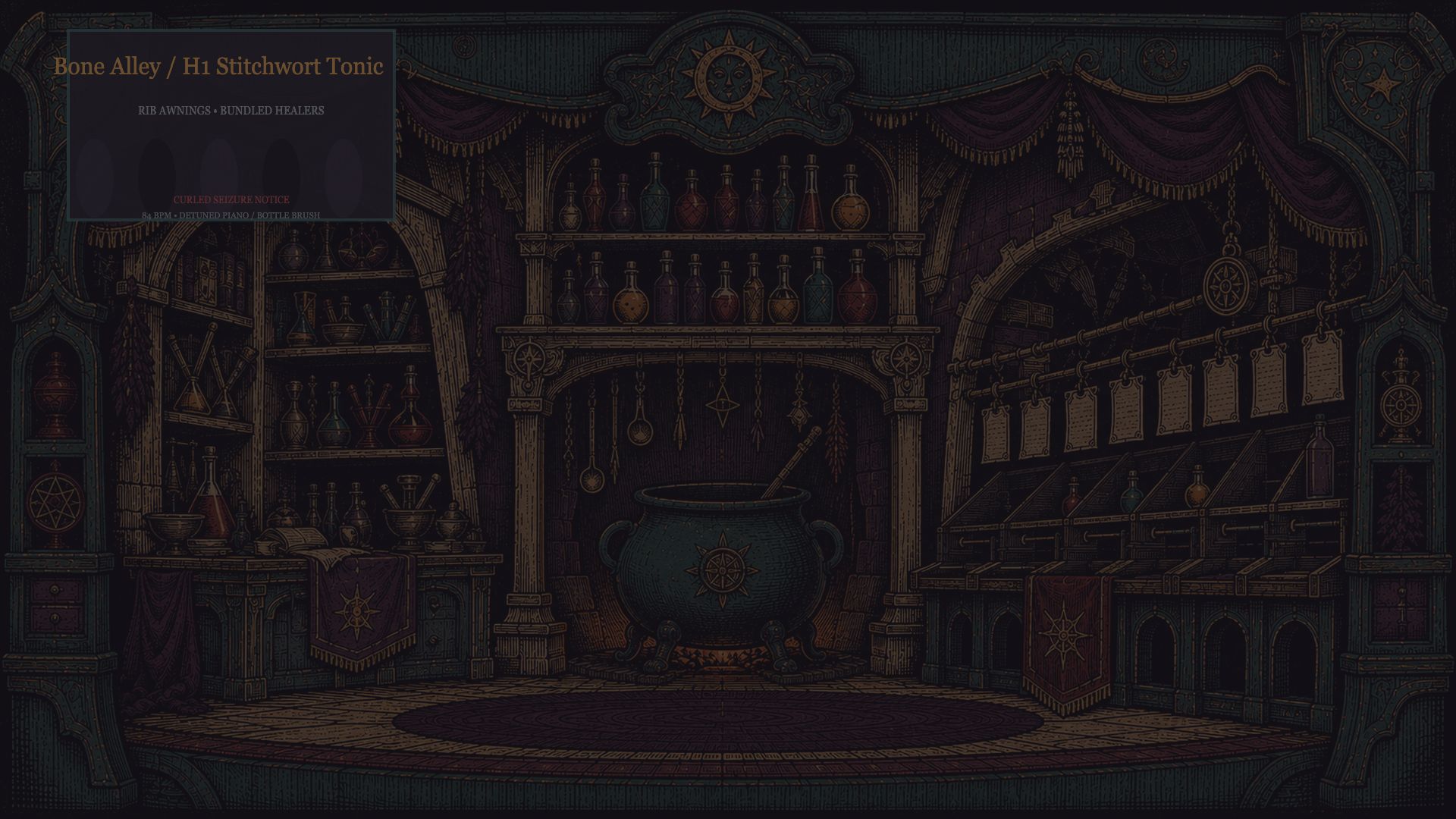}{730}{522}{Second selected}{Shop fails to open}{RCInk}
\RCAliasSequence{3}{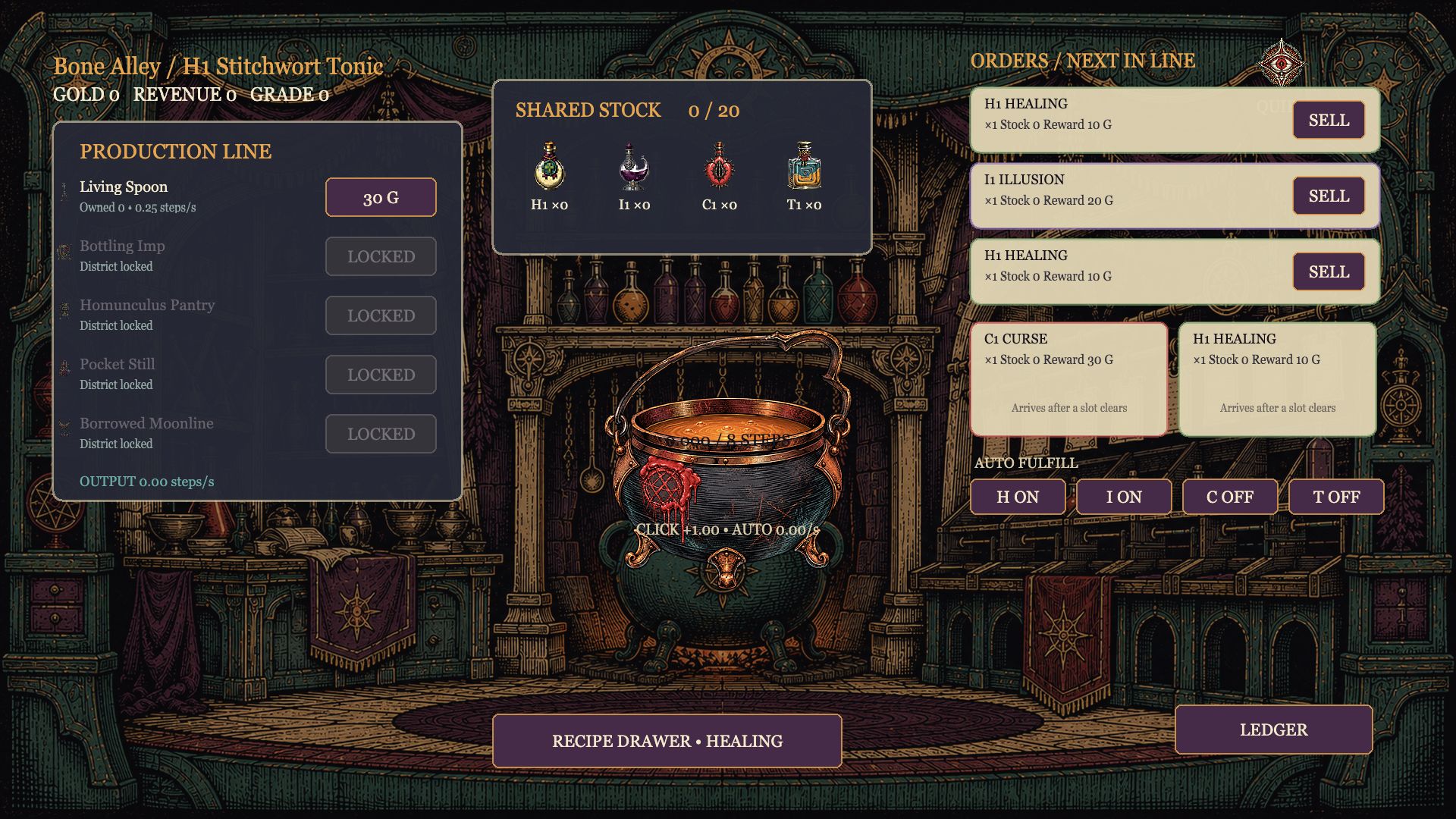}{730}{400}{Second selected}{Shop opens}{RCAccent}
\end{scope}
\hfill
\begin{scope}[shift={(0,14.61)}]
\RCArtRow{arms={3},
 label={fig:revision_self_today_hit},
 game={\texttt{Today's Best Spot}: Directional hit detection},
 rule={\shortstack[l]{CARD-PUNCH\\RT-PUNCH R3}},
 condition={Stage 3; aim $45^\circ$.\\Target vigor = 1.},
 effect={Rotate the hit area;\\reduce vigor to 0.},
 effect y={2.47}}
\newcommand{\RCMergedPunch}[4]{%
 \begingroup
 \pgfmathsetmacro{\PunchX}{4+(#1-1)*3.84}
 \node[font=\AtoZFigFont\fontsize{7.0}{8.8}\selectfont\bfseries] at ({\PunchX+1.75},1.08) {After one $45^\circ$ attack};
 \RCAppendixCrop{\PunchX}{1.25}{3.50}{2.91}{650}{640}{550}{punch_#2_after_native.png}{1920}
 \draw[black!75,line width=1.6pt] ({\PunchX+1.10},2.25)--({\PunchX+1.78},2.93);
 \draw[white,line width=.8pt,->] ({\PunchX+1.10},2.25)--({\PunchX+1.78},2.93);
 \path[fill=#4!6!white,draw=RCBorder,line width=.4pt] (\PunchX,4.20) rectangle ({\PunchX+3.50},4.62);
 \node[ra art result,text=#4] at ({\PunchX+1.75},4.41) {#3};
 \endgroup
}
\RCMergedPunch{1}{base}{Miss: vigor remains 1}{RCInk}
\RCMergedPunch{2}{naive_iter2}{Miss: vigor remains 1}{RCInk}
\RCMergedPunch{3}{v5_iter2}{Hit: vigor 0; departs}{RCAccent}
\end{scope}
\end{tikzpicture}%
}
\end{subcaptiongroup}
\endgroup
\caption{\textbf{Selected GDD-specific repairs beyond self-revision.} Initial builds are compared with second-round \emph{self-revision} and \emph{Source + Replay + Playtest} under matched inputs and preconditions. The cases test \subref{fig:revision_art_toybox} confirmation before discarding a run, \subref{fig:revision_art_nightshift_timeout} CODE BLUE feedback on boss timeout, \subref{fig:self_revision_alias_selection} opening the shop after alias selection, and \subref{fig:revision_self_today_hit} a diagonal attack against a target inside the specified hit area. Each displayed requirement remains unresolved after \emph{self-revision} and is satisfied by \emph{Source + Replay + Playtest} in the probe. Life and vigor values are recorded during execution; all targets in \subref{fig:revision_self_today_hit} start at vigor 1. Crops retain the archived builds' native art; arrows indicate attack aim. These selected outcomes do not establish complete GDD compliance.}
\label{fig:revision_gdd_cases}
\end{figure*}

We examine requirement-level behavior in the initial and second-round revision builds, complementing the aggregate results in Section~\ref{sec:iterative_revision}. Figure~\ref{fig:revision_gdd_cases} extends the main comparisons to four additional games from the \emph{Big} split. It compares the initial build, \emph{self-revision}, and \emph{Source + Replay + Playtest}. Each comparison uses matched inputs and preconditions, with the GDD requirement shown beside the resulting game behavior.

For the main-figure interaction case in \texttt{Beat Reroute}, SUR-V1 requires a captured fragment to follow the pointer with a 20\,px vertical offset. We establish the same held \texttt{LN-PER} fragment at slot 10 and move the pointer while holding its button, first to $(850,610)$ and then to $(1100,560)$. The initial and \emph{self-revision} builds leave the fragment at its timeline origin. \emph{Source + Replay + Playtest} renders its center at $(850,590)$ and $(1100,540)$, respectively (Figure~\ref{fig:revision_cases}\subref{fig:revision_case_beat}). In \texttt{Abyssal Chain}, RT-DAMAGE D5 requires Oxygen reaching 0 to set Hull to 0 immediately. The builds start with Oxygen $=0.5$\,s and Hull $=100$, with enemies and projectiles removed to isolate oxygen depletion. After 0.6\,s, all three display the drowned-run screen, but only \emph{Source + Replay + Playtest} sets Hull to 0. Figure~\ref{fig:revision_cases}\subref{fig:revision_case_abyssal} reports the values recorded during execution.

In \texttt{Toybox Rider}, UI-02 requires Restart in Pause to open a discard confirmation, with Cancel returning to Pause (Figure~\ref{fig:revision_gdd_cases}\subref{fig:revision_art_toybox}). The comparison starts from the same paused run at distance 148\,m, score 310, and 7 run Bolts. Pressing Enter leaves the initial build in Pause and immediately restarts \emph{self-revision} with zero run Bolts. \emph{Source + Replay + Playtest} opens the confirmation while preserving the run; Escape then returns to Pause with the same values. This comparison isolates the confirmation required before discarding a run.

In \texttt{Night Shift Rx}, boss timeout must trigger CODE BLUE, whose visual feedback includes a red-grey life ring (RT-STAGE R1 and VFX-CODEBLUE). We start with 0.05\,s remaining on the boss timer and Life $=80$, then advance the game by 0.1\,s. The initial build leaves the patient in its live state. \emph{Self-revision} enters CODE BLUE but retains Life $\approx79.8$ and the green ring. \emph{Source + Replay + Playtest} enters CODE BLUE, sets Life to 0, and displays the required red-grey ring (Figure~\ref{fig:revision_gdd_cases}\subref{fig:revision_art_nightshift_timeout}). The difference concerns the state and visual feedback accompanying timeout, beyond the transition to CODE BLUE itself.

In \texttt{Alias Alchemy Shop}, the \texttt{AliasSelection} controls require clicking a signboard and then \textsc{Open Shop}. We perform this two-step sequence in each build (Figure~\ref{fig:revision_gdd_cases}\subref{fig:self_revision_alias_selection}). The initial build ignores the second-signboard click and proceeds to the shop without applying that choice. \emph{Self-revision} selects the second candidate, but \textsc{Open Shop} raises a \texttt{TypeError} during shop rendering, leaving the shop interface incomplete. \emph{Source + Replay + Playtest} retains the selected alias and renders the shop interface; a subsequent cauldron click also advances production. Thus, candidate selection alone does not establish completion of the required interaction. The comparison concerns this selection-to-shop sequence. Candidate names may differ across builds.

In \texttt{Today's Best Spot}, the Paw Punch hit area must rotate to the cursor's nearest of eight directions (CARD-PUNCH); RT-PUNCH R3 specifies a $180\times90$\,px stage-3 hit area and one vigor pip of damage. We establish an airborne player at $(400,340)$ and a one-vigor territorial cat at $(490,430)$, then issue one left-click aimed diagonally down-right. The target lies about 127\,px along the aim and 0\,px across it, inside the specified area. After one execution frame, the initial and \emph{self-revision} builds leave the target at one vigor. \emph{Source + Replay + Playtest} reduces its vigor to zero and the target departs (Figure~\ref{fig:revision_gdd_cases}\subref{fig:revision_self_today_hit}). The recorded attack direction is $45^\circ$ in all three builds. This probe concerns directional target inclusion and damage, rather than exact boundary dimensions or attack timing.

The screenshots retain the archived game art. Controlled preparation establishes each requirement's preconditions; the original game logic produces the subsequent state changes and rendered outcomes.

\ifdefined\AtoZAppendixBarrier\AtoZAppendixBarrier\fi

\paragraph{Qualitative Gameplay Examples.}
Figures~\ref{fig:small_gameplay_01} and~\ref{fig:small_gameplay_02} show selected gameplay excerpts from eight generated \emph{Small} games. \texttt{Bloom or Weed} presents flowers and weeds that require clicking and withholding input, respectively, while \texttt{Odd Patch} asks the player to identify differences in color or shape. \texttt{Orbit} shows the player on two concentric rings followed by visible feedback at hazard contact, and \texttt{Echo Three} alternates between pattern presentation and player input. \texttt{Key Under Cups} depicts the reveal, shuffle, and selection feedback of a visual tracking task, while \texttt{Bin Bit} communicates shape categories through persistent labels and color-coded bins.
\texttt{Greedy Die} distinguishes temporary and banked points as the player rolls and secures the target total.
\texttt{Safe Dial} combines a rotating dial with a fixed pointer and indicators for confirmed and remaining combination entries. These examples illustrate how gameplay prompts and state changes appear in the generated interfaces.

\paragraph{Qualitative Examples From \emph{Big}-GDD Games.}
Figures~\ref{fig:big_gameplay_01} and~\ref{fig:big_gameplay_02} present selected gameplay excerpts from eight generated \emph{Big} games. \texttt{Cloud Cast} combines casting and reeling controls with catch rewards, while \texttt{Afterglow Network} presents traversal tools, regulator installation, and base defense. \texttt{Siege Deck 2D} exposes unit deployment, tactical order cards, and deck rewards, and \texttt{Wham Bam Logistics} provides route, fleet, and hub management within a shared city map. \texttt{ChromaShade} combines shadow controls with platform traversal, while \texttt{Ssitgim} presents soul encounters, a cleansing interface, and a boss encounter.
\texttt{Alias Alchemy Shop} exposes production, shared inventory, and the distinction between resources that reset and progress that persists at closure.
\texttt{Strata Keepers} provides separate interfaces for recording excavation context, rejoining artifact fragments, and assembling evidence for research hypotheses.
These examples illustrate how distinct mechanics are expressed through gameplay controls, state information, and feedback in the generated interfaces.

\AtoZAppendixInput[p]{figures/appendix/visual_result/small_gameplay_01}

\AtoZAppendixInput[p]{figures/appendix/visual_result/small_gameplay_02}

\AtoZAppendixInput[p]{figures/appendix/visual_result/big_gameplay_01}

\AtoZAppendixInput[p]{figures/appendix/visual_result/big_gameplay_02}

\clearpage

\section{Limitations}
\label{app:limitations}
\AtoZbench primarily evaluates 100 synthetic game design documents (GDDs) for two-dimensional single-player games developed in Phaser. Whether these results generalize to human-authored specifications, alternative game engines, and broader software tasks remains to be established. The three additional Three.js games in Appendix~\ref{app:scalable_benchmark} demonstrate applicability to 3D environments but do not establish performance across a broad range of engines or genres. Contract construction and evidence interpretation are based on generative models, so fixed contracts and repeated judgments cannot fully eliminate omissions or evaluator bias.
\clearpage

\end{document}